\documentclass[10pt]{article}

\usepackage{bm}
\usepackage{bbm}
\usepackage{fullpage}  % for narrow margins
\usepackage{url}
\usepackage{algpseudocode}
\usepackage{multirow} %//multirow for format of table
\usepackage{hhline}  % for \hhline in tables
\usepackage{verbatim}
\usepackage{graphicx}
\usepackage{amsmath,amssymb,amsfonts,graphicx,amsthm,mathtools,nicefrac}%},mathrsfs}
\usepackage{lscape}
\usepackage{color}
\usepackage{authblk}
\usepackage{subfigure}
\usepackage[ruled,vlined]{algorithm2e}

\usepackage{bbding}
\usepackage{indentfirst}
\usepackage{xurl}

\usepackage{graphicx}

\usepackage{multirow}
\allowdisplaybreaks
\newcommand\numberthis{\addtocounter{equation}{1}\tag{\theequation}}
\numberwithin{equation}{section}

\usepackage{bbm}

\usepackage{amsmath,mathrsfs,amsfonts,amssymb} 
\usepackage[dvipsnames]{xcolor}
\usepackage{subfigure}
\usepackage{parskip}

\usepackage{makecell}

\usepackage[utf8]{inputenc}
\usepackage[T1]{fontenc}
\usepackage{hyperref}
\usepackage{url}
\usepackage{booktabs}
\usepackage{amsfonts}
\usepackage{nicefrac}
\usepackage{microtype}
\usepackage{xcolor}
\usepackage{graphicx}
\usepackage{subcaption}
\usepackage{amsmath,amsthm,amssymb}
\usepackage{lipsum}
\usepackage[export]{adjustbox}
\usepackage{enumitem}
\usepackage{wrapfig,lipsum,booktabs}
\usepackage{soul} % 引入 soul 宏包
\usepackage{cleveref}
\usepackage[most]{tcolorbox}
\definecolor{lightgreen}{rgb}{0.56, 0.93, 0.56}
\definecolor{moonstoneblue}{rgb}{0.45, 0.66, 0.76}
\usepackage{authblk}

\usepackage{enumitem}    
\newcounter{ablation}

\makeatletter
\renewcommand\AB@affilsepx{, \protect\Affilfont}
\makeatother

\title{Skywork Open Reasoner 1 Technical Report}

\author[ ]{Jujie He$^{*,\dagger}$}
\author[ ]{Jiacai Liu$^*$}
\author[ ]{Chris Yuhao Liu}
\author[ ]{Rui Yan}
\author[ ]{Chaojie Wang}
\author[ ]{Peng Cheng}
\author[ ]{Xiaoyu Zhang}
\author[ ]{Fuxiang Zhang}
\author[ ]{Jiacheng Xu}
\author[ ]{Wei Shen}
\author[ ]{Siyuan Li}
\author[ ]{Liang Zeng}
\author[ ]{Tianwen Wei}
\author[ ]{Cheng Cheng}
\author[ ]{Bo An}
\author[ ]{Yang Liu}
\author[ ]{Yahui Zhou}
\affil[ ]{Skywork AI, Kunlun Inc}

\date{}

\begin{document}
\maketitle

\renewcommand{\thefootnote}{\fnsymbol{footnote}}
\footnotetext[1]{Equal contribution.}
\footnotetext[2]{Corresponding author: jujie.he@kunlun-inc.com}

\begin{center}
\end{center}
\vspace{-35pt}
\begin{center}
    \textbf{GitHub:} \url{https://github.com/SkyworkAI/Skywork-OR1} \\
    % \textbf{HuggingFace Datasets:} \url{https://huggingface.co/datasets/Skywork/Skywork-OR1-RL-Data} \\
    \vspace{3pt}
    \textbf{HuggingFace:} \url{https://huggingface.co/Skywork/Skywork-OR1-32B}
\end{center}

%%%%%%
\begin{abstract}
\noindent
The success of DeepSeek-R1 underscores the significant role of reinforcement learning (RL) in enhancing the reasoning capabilities of large language models (LLMs). In this work, we present Skywork-OR1, an effective and scalable RL implementation for long Chain-of-Thought (CoT) models. Building on the DeepSeek-R1-Distill model series, our RL approach achieves notable performance gains, increasing average accuracy across AIME24, AIME25, and LiveCodeBench from 57.8\% to 72.8\% (+15.0\%) for the 32B model and from 43.6\% to 57.5\% (+13.9\%) for the 7B model. Our Skywork-OR1-32B model surpasses both DeepSeek-R1 and Qwen3-32B on the AIME24 and AIME25 benchmarks, while achieving comparable results on LiveCodeBench. The Skywork-OR1-7B and Skywork-OR1-Math-7B models demonstrate competitive reasoning capabilities among models of similar size. We perform comprehensive ablation studies on the core components of our training pipeline to validate their effectiveness. Additionally, we thoroughly investigate the phenomenon of entropy collapse, identify key factors affecting entropy dynamics, and demonstrate that mitigating premature entropy collapse is critical for improved test performance. To support community research, we fully open-source our model weights, training code, and training datasets.

% \begin{figure}[h]
%     \centering
%     \includegraphics[width=0.9\linewidth]{figures/72b_bench.png}
%     \caption{The avg@32 performance of Skywork-OR1-32B on benchmarks}
%     \label{fig:72b_bench}
% \end{figure}
\begin{figure}[!htbp]
    \centering
    \includegraphics[width=0.95\linewidth]{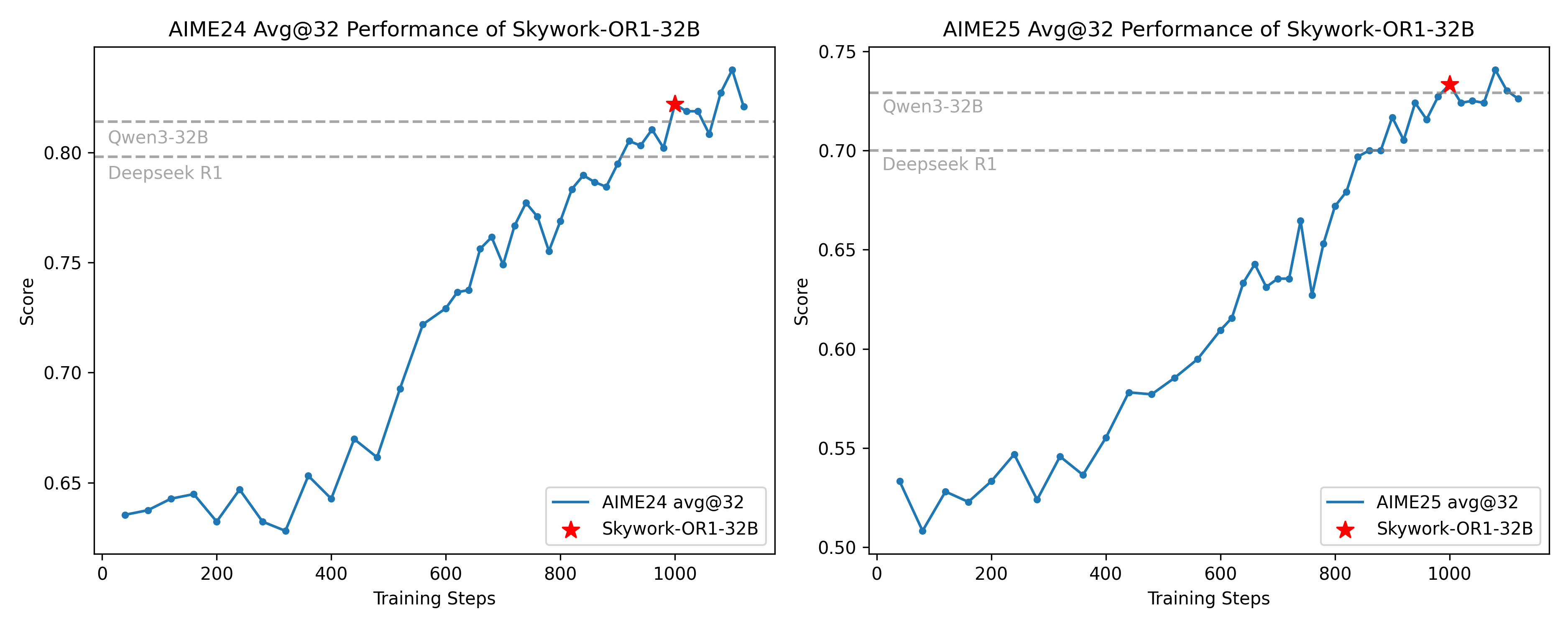}
    \vspace{-1em}
    \caption{The performance curve of Skywork-OR1-32B during RL training for AIME 2024 and AIME 2025. The red stars indicate the selected final checkpoints.}
    \label{fig:32b_perf}
\end{figure}

\end{abstract}

\newpage
\tableofcontents
\newpage

\section{Introduction}
In recent months, post-training techniques based on reinforcement learning (RL) have achieved groundbreaking success in enhancing the reasoning capabilities of large language models (LLMs). Representative models such as OpenAI-o1 \cite{jaech2024openai}, DeepSeek-R1 \cite{guo2025deepseek}, and Kimi-K1.5 \cite{team2025kimi} demonstrate RL's remarkable ability to significantly improve performance in mathematics and coding. While prior RL approaches have primarily relied on Monte Carlo Tree Search (MCTS) or Process Reward Models (PRMs) to improve reasoning over supervised fine-tuning (SFT) models, the success of DeepSeek-R1 demonstrates conclusively that online RL with a simple rule-based reward is sufficient to substantially enhance the reasoning capabilities of base models.

As model capabilities continue to advance, Chains-of-Thought (CoT) have grown progressively longer. For example, the DeepSeek-R1-Distill model series \cite{guo2025deepseek} generates CoT sequences averaging over 10K tokens on the AIME24 benchmark, significantly surpassing earlier popular SFT models such as the Qwen 2.5 model series \cite{yang2024qwen2} and the Llama 3.1 model series \cite{grattafiori2024llama}. Despite several reproduction efforts (e.g., Logic-RL \cite{xie2025logicrlunleashingllmreasoning}, Open-Reasoner-Zero \cite{hu2025openreasonerzeroopensourceapproach}, DAPO \cite{yu2025dapo}, VAPO \cite{yuan2025vapo}) following the success of DeepSeek-R1, most have focused on applying RL to base models rather than to long CoT models that have already undergone SFT. As a result, it remains unclear how to improve the reasoning abilities of long CoT models using RL in an efficient and scalable manner. While recent works such as DeepScaleR \cite{deepscaler2025}, Light-R1 \cite{wen2025light}, and DeepCoder \cite{deepcoder2025} have made preliminary progress toward efficient RL optimization for long CoT models, their analyses do not systematically disentangle the contributions of distinct algorithmic components during RL training.

In this work, we introduce Skywork Open Reasoner 1 (abbreviated as Skywork-OR1 throughout the report), an efficient and scalable RL recipe for long CoT models. Our experiments are based on the DeepSeek-R1-Distill model series and open-source datasets with rigorous preprocessing and filtering. As shown in Figure~\ref{fig:32b_perf} and Table~\ref{table:eval_results}, the Skywork-OR1 model series achieves significant performance improvements over base models, demonstrating the effectiveness of our RL implementation. Specifically, Skywork-OR1-32B achieves scores of 82.2 on AIME24, 73.3 on AIME25, and 63.0 on LiveCodeBench\cite{jain2024livecodebench} (2024-08 - 2025-02), outperforming DeepSeek-R1 and Qwen3-32B in the math domain. Skywork-OR1-7B achieves 70.2 on AIME24, 54.6 on AIME25, and 47.6 on LiveCodeBench, exhibiting competitive performance relative to similarly sized models in both math and coding tasks. Our previously released model, Skywork-OR1-Math-7B, also delivers strong performance among similarly sized models, scoring 69.8 on AIME24, 52.3 on AIME25, and 43.6 on LiveCodeBench. We conducted exhaustive ablation experiments to validate the effectiveness of the core components in the training pipeline.

Balancing exploration and exploitation is crucial in RL training \cite{sutton2018reinforcement}. We conducted a comprehensive study on premature entropy collapse, a phenomenon associated with excessive exploitation, and found that mitigating premature entropy collapse is essential for achieving better test performance. Through exhaustive ablation experiments, we identified key factors that influence entropy dynamics.

To ensure full reproducibility and support ongoing research within the LLM community, we release all of our training resources, including source code\footnote{\url{https://github.com/SkyworkAI/Skywork-OR1}}, the post-training dataset\footnote{\url{https://huggingface.co/datasets/Skywork/Skywork-OR1-RL-Data}}, and model weights\footnote{\url{https://huggingface.co/Skywork/Skywork-OR1-7B}} \footnote{\url{https://huggingface.co/Skywork/Skywork-OR1-32B}}. Furthermore, we conducted extensive ablation studies across both data and algorithmic dimensions to elucidate effective RL implementations for long CoT models. As a follow-up to our previously released Notion blog post \cite{skywork-or1-2025}, we present this more detailed technical report, with our key findings summarized as follows:

\begin{tcolorbox}[colback=SeaGreen!10!CornflowerBlue!10,colframe=RoyalPurple!55!Aquamarine!100!,title=\textbf{Data Collection}]
\begin{enumerate}[leftmargin=5pt,
rightmargin=5pt,itemsep=5pt]
    \item To ensure stable and effective training, it is crucial to incorporate problems from a diverse set of sources. We observe that, in the absence of consistent quality assessment and filtering procedures, previously successful datasets exhibit several failure modes with larger models (Section \ref{sec:dataset}).
    \item Rigorous filtering and quality control of training data significantly accelerate learning. Our proposed data mixture, constructed with stringent filtering criteria, outperforms a baseline mixture assembled with looser quality thresholds (Section \ref{sec:data_mixture}).
\end{enumerate}
\end{tcolorbox}

\begin{tcolorbox}[colback=SeaGreen!10!CornflowerBlue!10,colframe=RoyalPurple!55!Aquamarine!100!,title=\textbf{Training Strategy}]
\begin{enumerate}[leftmargin=5pt,
rightmargin=5pt,itemsep=5pt]
    \item Multi-stage training significantly improves training efficiency in the initial phase while preserving scalability for later stages (Section \ref{sec:multi_stage}).
    \item Addressing noisy training signals introduced by truncated trajectories in \emph{Stage I} does not lead to better scaling at large context lengths, e.g., 32\text{K} (Section \ref{sec:adv_mask}).
    \item High-temperature sampling results in lower test accuracy during the early training steps but ultimately yields greater performance improvements (Section \ref{sec:temperature}).
    \item On-policy training mitigates entropy collapse and leads to higher test performance (Section \ref{sec:entropy}).
\end{enumerate}
\end{tcolorbox}

\begin{tcolorbox}[colback=SeaGreen!10!CornflowerBlue!10,colframe=RoyalPurple!55!Aquamarine!100!,title=\textbf{Loss Function}]
\begin{enumerate}[leftmargin=5pt,
rightmargin=5pt,itemsep=5pt]
    \item Adaptive entropy control effectively keeps the model’s entropy lower-bounded by the target entropy throughout training, maintaining the model’s exploration ability and high learning plasticity, with test performance steadily improving (Section \ref{sec:adaptive_entropy}).
    \item The KL penalty hinders further improvements in test performance during multi-stage training. Therefore, we omit KL loss from our training pipeline (Section \ref{sec:kl}).
\end{enumerate}
\end{tcolorbox}

\begin{tcolorbox}[colback=SeaGreen!10!CornflowerBlue!10,colframe=RoyalPurple!55!Aquamarine!100!,title=\textbf{Empirical Results of Our Entropy Collapse Study}]
\begin{enumerate}[leftmargin=5pt,
rightmargin=5pt,itemsep=5pt]
    \item Faster entropy collapse generally correlates with poorer test performance (Section \ref{sec:entropy_motivation}). Appropriate entropy control that mitigates premature convergence can improve test outcomes (Section \ref{sec:entropy_control}). 
    \item Increasing rollout diversity by enlarging the batch and group sizes has only minor effects on entropy dynamics (Section \ref{sec:impact_of_rollout_diversity}), whereas using a higher sampling temperature significantly impacts initial entropy and learning dynamics (Section \ref{sec:temperature}).
    \item Off-policy training -- via increased mini-batches or data reuse -- accelerates entropy collapse and generally leads to degraded test performance compared to on-policy updates, due to the introduction of off-policy data (Section \ref{sec:sgd_steps}).
    \item The entropy loss exhibits high sensitivity to both the training data and the coefficient. By either adaptively adjusting the entropy loss coefficient or applying a clip-higher trick with an appropriate clip ratio, entropy dynamics become slower and more stable, leading to improved test performance. Nevertheless, entropy still converges faster than in on-policy training (Section \ref{sec:entropy_control}).
\end{enumerate}
\end{tcolorbox}

\paragraph{Organization} In Section \ref{sec:Preliminaries}, we introduce the preliminaries of several important policy optimization methods in RL. Section \ref{sec:MAGIC} elaborates on our training pipeline, including comprehensive ablation studies that validate the effectiveness of its core components. A systematic investigation of entropy collapse is presented in Section \ref{sec:entropy}, demonstrating that mitigating premature policy convergence is critical in RL training for enhancing exploration and achieving better test performance. We discuss training resource allocation in Section \ref{sec:training_time_scaling}. The implementation details of our training data preparation and rule-based reward are provided in Sections \ref{sec:dataset} and \ref{sec:verifiers}. Finally, Section \ref{sec:experiments} presents a comprehensive description of the training and evaluation details for our three released models: Skywork-OR1-Math-7B, Skywork-OR1-7B, and Skywork-OR1-32B.

\begin{figure}[!htbp]
    \centering
    \includegraphics[width=0.95\linewidth]{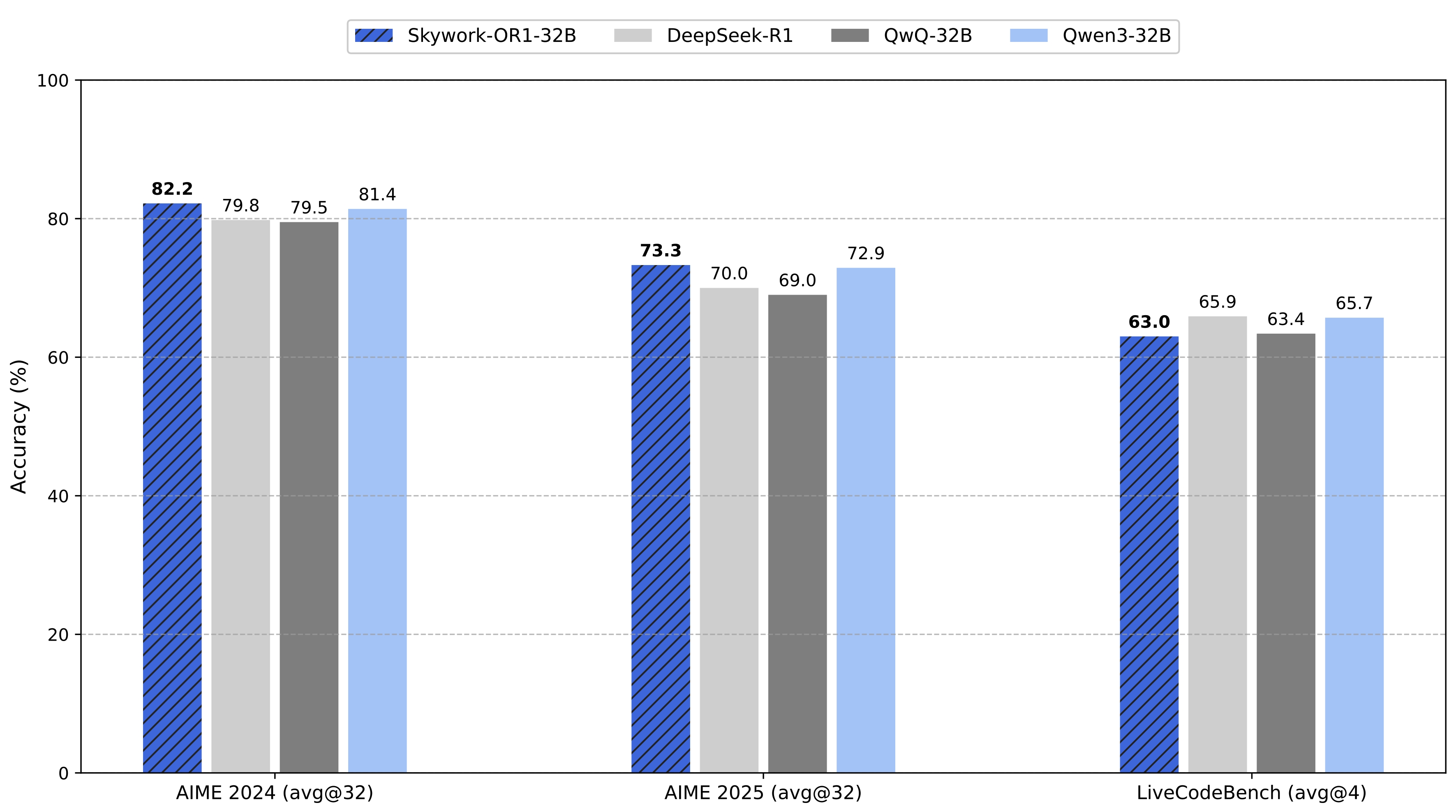}
    \caption{Performance of Skywork-OR1-32B on challenging mathematics and coding benchmarks.}
    \label{fig:32b_eval}
\end{figure}

\begin{figure}[!htbp]
    \centering
    \includegraphics[width=0.95\linewidth]{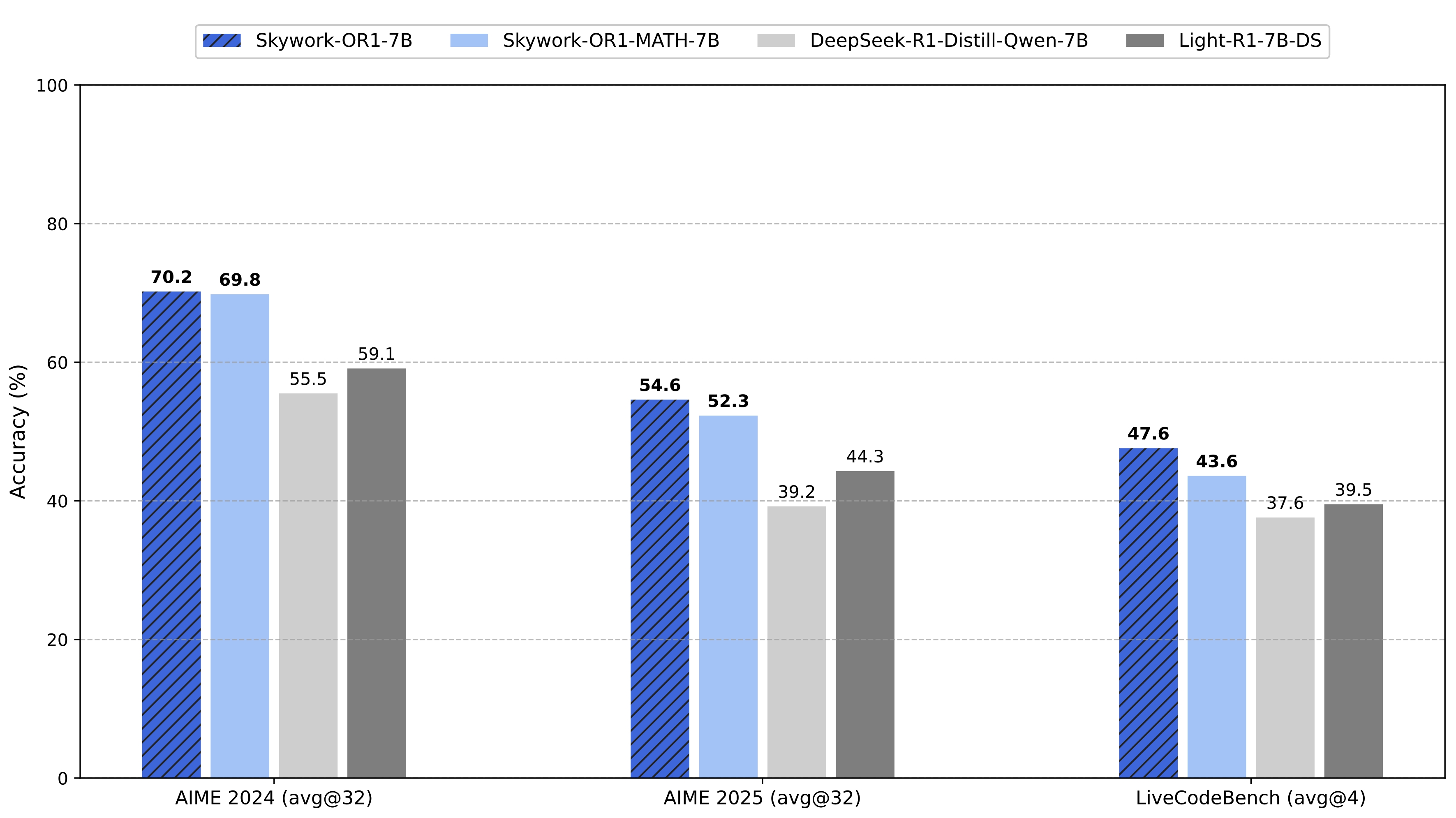}
    \caption{Performance of Skywork-OR1-7B on challenging mathematics and coding benchmarks.}
    \label{fig:7b_eval}
\end{figure}
\section{Preliminaries}
\label{sec:Preliminaries}
The success of Deepseek-R1 demonstrates that Policy Gradient (PG) methods \cite{sutton2018reinforcement}, especially Group Relative Policy Optimization(GRPO)  \cite{shao2024deepseekmath}, can effectively enhance the reasoning abilities of LLMs. Generally speaking, the RL objective is to find a policy $\pi$ that maximizes the reward, i.e.: 
\begin{align}
\underset{\pi}{\max}\,\,\left\{ \mathcal{J} \left( \pi \right) :=\mathbb{E} _{x\sim \mathcal{D}}\mathbb{E} _{y\sim \pi \left( \cdot |x \right)}\left[ r\left( x,y \right) \right] \right\},
\label{RL_obj}
\end{align}
where $x$ is the training prompt, $\mathcal{D}$ is the sampling distribution of $x$, $y$ is the response sampled by the policy $\pi$ for input prompt $x$, and $r$ denotes the reward function. 

In practice, we estimate a surrogate objective for $\mathcal{J}(\pi)$ at the batch level for tractable optimization. At each training step $k$, we sample a batch of $N$ prompts ${x_1, \ldots, x_N}$ from the data distribution $\mathcal{D}$, denoted as $\mathcal{T}_k$, and generate the corresponding responses ${y_1, \ldots, y_N}$ using the current policy $\pi$ with a context length $T$ and temperature $\tau$.
The batch-level surrogate objective at step $k$ can be formulated as:
\begin{align}
\underset{\pi}{\max}\,\,\left\{ \mathcal{J}_k \left( \pi \right) :=\mathbb{E} _{x_i\sim \mathcal{T}_k}\mathbb{E} _{y_i\sim \pi \left( \cdot |x_i \right)}\left[ r\left( x_i,y_i \right) \right] \right\},
\label{RL_batch_obj}
\end{align}
where $\pi_k$ is shorthand for the policy $\pi_{\theta_k}$ parameterized by $\theta_k$.

\paragraph{Vanilla Policy Gradient}
For a parameterized policy $\pi_\theta$, vanilla PG \cite{pg} uses gradient ascent to obtain the optimal parameter $\theta^*$, i.e.
\begin{align*}
\theta \gets \theta +\nabla_\theta \mathcal{J} \left( \pi _{\theta} \right).
\end{align*}
A valid first-order surrogate policy loss for vanilla PG at each iteration $k$ is given by: 
\begin{align}
\mathcal{L}^{\mathrm{PG}} _k\left( \theta \right) = - \mathbb{E} _{x_i\sim \mathcal{T}_k}\mathbb{E} _{y_i\sim \pi _k\left( \cdot |x_i \right)}\left[ \sum_{t=0}^{\left| y_i \right|-1}{\frac{\pi _{\theta}\left( a^t_i|s^t_i \right)}{\pi_k \left( a^t_i|s^t_i \right)}\cdot A^{\pi_{k}}\left( s^t_i,a^t_i \right)} \right],
\label{vanilla_policy_loss}
\end{align}
where the response $y_i=(a_i^0,...,a_i^{\left| y \right|-1})$ consists of $\left| y \right|$ tokens, $a_i^t$ is the $t$-th token in the sequence $y_i$, $s_i^t:=(x_i,a_i^0,...,a_i^{t-1})$ is the prefix context when generating $a_i^t$, and $A^{\pi_{k}}$ is the advantage function defined as 
\begin{align*}
A^{\pi _k}\left( s^t,a^t \right) :=\mathbb{E} _{y\sim \pi_k\left( \cdot |x \right)}\left[ r\left( x,y \right) |s^t, a^t \right] -\mathbb{E} _{y\sim \pi_k\left( \cdot |x \right)}\left[ r\left( x,y \right) |s^t \right].
\end{align*}
One can easily show that $\nabla _{\theta}\mathcal{L}^{\mathrm{PG}} _k\left( \theta_k \right) =-\nabla _{\theta}\mathcal{J}_k \left( \pi_k \right)$. 

\paragraph{Proximal Policy Optimization (PPO)} At each training step $k$, PPO \cite{schulman2017proximal} performs multiple gradient descent steps on the policy loss $\mathcal{L}_k$ with a clip trick to keep the new policy restricted within the trust region of $\pi_{k}$. The policy loss employed in PPO is formulated as: 
\begin{align*}
\mathcal{L} _{k}^{\mathrm{PPO}}\left( \theta \right) =-\mathbb{E} _{x_i \sim \mathcal{T}_k}\mathbb{E} _{y_i\sim \pi _{k}\left( \cdot |x_i \right)}\left[ \sum_{t=0}^{\left| y_i \right|-1}{\min \left( \rho_i^t\left( \theta \right) A^{\pi _k}\left( s_i^t,a_i^t \right) ,\mathrm{clip}\left( \rho_i^t\left( \theta \right) ,1-\varepsilon ,1+\varepsilon \right) \cdot A^{\pi _k}\left( s_i^t,a_i^t \right) \right)} \right],
\end{align*}
where $\rho_i^t(\theta):=\frac{\pi _{\theta}\left( a_i^t|s_i^t \right)}{\pi _{k}\left( a_i^t|s_i^t \right)}$, and $\varepsilon$ is the clip hyperparameter. In practice, PPO generally
uses GAE \cite{schulman2015high} to estimate the token-level advantage $A^{\pi_k}(s_i^t,a_i^t)$.

\paragraph{Group Relative Policy Optimization (GRPO)} 
Suppose $M$ i.i.d. responses $y_{i1},..,y_{iM}$ are sampled for each prompt $x_i$. GRPO \cite{shao2024deepseekmath} estimates the token-level advantage using the group-normalized rewards and introduces an additional length normalization term $\frac{1}{\left|y_{ij}\right|}$ for each response $y_{ij}$. The policy loss employed in GRPO is formulated as: 
\begin{align*}
\mathcal{L} _{k}^{\mathrm{GRPO}}\left( \theta \right) &=-\mathbb{E} _{x_i \sim \mathcal{T}_k}\mathbb{E} _{\left\{ y_{ij} \right\} _{j=1}^{M}\sim \pi _{k}\left( \cdot |x \right)}
\\
\,\,           &\left[ \frac{1}{M}\sum_{i=1}^M{\frac{1}{\left| y_{ij} \right|}\sum_{t=0}^{\left| y_{ij} \right|-1}{\min \left( \rho _{ij}^{t}\left( \theta \right) A_{ij}^{t},\mathrm{clip}\left( \rho _{ij}^{t}\left( \theta \right) ,1-\varepsilon ,1+\varepsilon \right) A_{ij}^{t} \right) -\beta D_{ij}^{t}\left( \theta \right)}} \right],
\numberthis
\label{grpo_policy_loss}
\end{align*}
where $y_{ij}=(a_{ij}^0,...,a_{ij}^{\left| y_{ij} \right|-1})$, $a_{ij}^t$ is the $t$-th token in the sequence $y_{ij}$, $s_{ij}^t:=(x_i,a_{ij}^0,...,a_{ij}^{t-1})$, $\rho_{ij}^{t}(\theta ):=\frac{\pi _{\theta}\left( a_{ij}^{t}|s_{ij}^{t} \right)}{\pi _{k}\left( a_{ij}^{t}|s_{ij}^{t} \right)}$, $\varepsilon$ is the clip hyperparameter, $D_{ij}^{t}$ is the token-level k3 loss \cite{shao2024deepseekmath} applied in $a_{ij}^t$ with coefficient $\beta$ to keep the policy $\pi_\theta$ stay in the trust region of reference policy $\pi_\text{ref}$, i.e.
\begin{align*}
D_{ij}^{t}\left( \theta \right) :=\frac{\pi _{\mathrm{ref}}\left( a_{ij}^{t}|s_{ij}^{t} \right)}{\pi _{\theta}\left( a_{ij}^{t}|s_{ij}^{t} \right)}-\log \frac{\pi _{\mathrm{ref}}\left( a_{ij}^{t}|s_{ij}^{t} \right)}{\pi _{\theta}\left( a_{ij}^{t}|s_{ij}^{t} \right)}-1,
\end{align*}
For each prompt-response pair $(x_i,y_{ij})$, a binary reward $r\left( x_i,y_{ij} \right) \in \left\{ 0,1 \right\}$ is given by a rule-based verifier.
The token-level advantage $A_{ij}^t$ is estimated by
\begin{align}
\forall t:A_{ij}^{t}=\frac{r\left( x_i,y_{ij} \right) -\mathrm{mean}\left( r\left( x_i,y_{i1} \right) ,...,r\left( x_i,y_{iM} \right) \right)}{\mathrm{std}\left( r\left( x_i,y_{i1} \right) ,...,r\left( x_i,y_{iM} \right) \right)}.
\label{eq:advantages}
\end{align}

% At each training step $k$, we sample $N$ prompts $x_1,…,x_N$ from the dataloader $\mathcal{D}$ and $M$ i.i.d. response $y_{i1},….,y_{iM}$ for each prompt $x_i$ with context windows $T$ and temperature $\tau$. For each prompt-response pair $(x_i,y_{ij})$, an binary reward $r\left( x_i,y_{ij} \right) \in \left\{ 0,1 \right\}$ is given by a rule-based verifier. We estimate the token-level advantage by

% The policy loss we employed takes the form of 
% \begin{align}
% \mathcal{L} \left( \theta \right) =-\frac{1}{T_k}\sum_{i\in \mathcal{T} _k}{\sum_{j=1}^M{\left\{ \sum_{t=0}^{\left| y_{ij} \right|-1}{\min \left\{ \rho _{t}^{ij}\left( \theta \right) A_{t}^{ij},\mathrm{clip}\left( \rho _{t}^{ij}\left( \theta \right) ,1-\varepsilon ,1+\varepsilon \right) A_{t}^{ij} \right\} +\alpha _k\mathbb{H} _{t}^{ij}}\left( \theta \right) \right\}}},
% \label{policy_loss}
% \end{align}
% where  $y_{ij}:=(a^{ij}_0,...,a^{ij}_{\left| y_{ij} \right|-1})$, $a^{ij}_t$ is the token in the sequence $y_{ij}$ of index $t$, $s^{ij}_t:=(x_i,a^{ij}_0,...,a^{ij}_{t-1})$ is the prefix context when generating $a_{t}^{ij}$, $\rho _{t}^{ij}(\theta ):=\frac{\pi _{\theta}\left( a_{t}^{ij}|s_{t}^{ij} \right)}{\pi _{\theta _k}\left( a_{t}^{ij}|s_{t}^{ij} \right)}$, $\mathbb{H}^{ij}_t(\theta):=\mathcal{H} \left( \pi _{\theta}\left( \cdot |s_{t}^{ij} \right) \right)$ is the entropy of the generation policy of token $a_{t}^{ij}$, $\alpha_k \ge 0$ is the coefficient of the entropy, $T_k:=\frac{1}{T_k}\sum_{i\in \mathcal{T} _k}{\sum_{j=1}^M{\left| y_{ij} \right|}}$ is the total number of tokens in the training batch. 

\section{MAGIC in Skywork-OR1}
\label{sec:MAGIC}

We employ a training pipeline built upon a modified version of GRPO \cite{shao2024deepseekmath}, referred to as \textbf{M}ulti-stage \textbf{A}daptive entropy scheduling for \textbf{G}RPO \textbf{I}n \textbf{C}onvergence (\textbf{MAGIC}). In the following sections, we first introduce the recipe of MAGIC and then analyze the effectiveness of each of its components.

\subsection{MAGIC}
\label{sec:policy_update}
%  At each training step $k$, we sample $N$ prompts $x_1,…,x_N$ from the dataloader $\mathcal{D}$ and $M$ i.i.d. response $y_{i1},….,y_{iM}$ for each prompt $x_i$ with context length $T$ and temperature $\tau$. For each prompt-response pair $(x_i,y_{ij})$, an binary reward $r\left( x_i,y_{ij} \right) \in \left\{ 0,1 \right\}$ is given by a rule-based verifier. We estimate the token-level advantage by
% \begin{align*}
% \forall t:A_{t}^{ij}=\frac{r\left( x_i,y_j \right) -\mathrm{mean}\left( r\left( x_i,y_1 \right) ,...,r\left( x_i,y_M \right) \right)}{\mathrm{std}\left( r\left( x_i,y_1 \right) ,...,r\left( x_i,y_M \right) \right)}.
% \end{align*}
\iffalse
\begin{figure}[h]
    \centering
    \includegraphics[width=0.85\linewidth]{figures/magic.png}
    \caption{Overview of MAGIC recipe for training Skywork-OR1 model series.}
    \label{fig:magic}
\end{figure}
\fi

In the following, we present the MAGIC framework by detailing its components in terms of Data Collection, Training Strategy, and Loss Function.

\paragraph{Data Collection} 
To ensure the quality of queries during post-training, we construct the initial dataset through stringent data preparation, as described in Section~\ref{sec:dataset}, and adopt more accurate verifiers to provide reward signals, as outlined in Section~\ref{sec:verifiers}. Additionally, we employ the following strategies to further improve sample efficiency:

\begin{enumerate} 
    \item \textbf{Offline and Online Filtering.} We apply data filtering both before and during training. Prior to training, we remove prompts with base model correctness rates of 1 (fully correct) or 0 (completely incorrect). During training, at the beginning of each stage, we also discard training prompts for which the actor model achieved correctness of 1 in the previous stage. This dynamic online filtering mechanism ensures that the actor model is consistently trained on challenging problems at each stage.
    \item \textbf{Rejection Sampling.} Responses in the zero-advantage group (as defined by Equation \eqref{eq:advantages}) do not contribute to the policy loss but may influence the KL loss or entropy loss, potentially leading to a more unstable training process due to the implicitly increased relative weight of these losses. To mitigate this issue, our training batches include only groups with non-zero advantages; specifically, the samples of prompt $x_i$ are filtered out if $i \notin \tilde{\mathcal{T}}_k$, where
\begin{align*}
\tilde{\mathcal{T}}_k := \left\{ i \in \left[ N \right] : \exists j \in \left[ M \right] \,\, \hat{A}_{ij} \ne 0 \right\}.
\end{align*}
\end{enumerate}

\paragraph{Training Strategy}
We made the following refinements to the training strategy of vanilla GRPO:
\begin{enumerate}
    \item \textbf{Multi-Stage Training.} Inspired by DeepScaleR \cite{deepscaler2025}, we progressively increase the context length $T$ and divide the training process into multiple stages. We found that multi-stage training significantly reduces computational costs while preserving scalability, as supported by the evidence presented in Section~\ref{sec:multi_stage}.
    \item \textbf{Advantage Mask for Truncated Responses.}
    To address potential noise in training signals when outcomes cannot be derived from truncated responses -- since assigning negative advantages in such cases may introduce bias -- we experimented with an advantage mask during the early stages of multi-stage training, when many responses are truncated. However, as shown in Section~\ref{sec:adv_mask}, penalizing truncated responses does not hinder later-stage improvements and enhances token efficiency. Based on these results, we do \textbf{not} employ any advantage mask strategy in our training pipeline.
    \item \textbf{High-Temperature Sampling.} We set the rollout temperature to $\tau=1$ to enhance the model's exploration capability and improve learning plasticity. This decision was motivated by our observation that the sampling policy either immediately enters (in the case of math data) or quickly transitions into (in the case of code data) a low-entropy state when using a smaller sampling temperature (e.g., $\tau=0.6$). See Section~\ref{sec:temperature} for further details.
    \item \textbf{On-Policy Training.} We adopted on-policy training for Skywork-OR1-7B and Skywork-OR1-32B, as we found that on-policy updates significantly slow entropy collapse and lead to higher test performance. See Section~\ref{sec:entropy} for our detailed findings on entropy collapse. In contrast, Skywork-OR1-Math-7B was trained with two gradient steps per training step (and was therefore not strictly on-policy). This setup preceded our complete understanding of the relationship between off-policy updates and premature entropy collapse. Nevertheless, adaptive entropy control (Section~\ref{sec:adaptive_entropy}) effectively mitigated collapse, allowing the model to achieve strong performance.
\end{enumerate}

\paragraph{Loss Function} 
To mitigate implicit length bias, we adopt a token-level policy loss by removing the length normalization term $1/\left| y_{ij} \right|$
from each response. The policy loss is averaged across all tokens in a training batch, formulated as follows:
\begin{align}
\mathcal{L}^{\text{MAGIC}}\left( \theta \right) =-\frac{1}{T_k}\sum_{i\in \tilde{\mathcal{T}} _k}{\sum_{j=1}^M{\left\{ \sum_{t=0}^{\left| y_{ij} \right|-1}{\min \left\{ \rho _{ij}^{t}\left( \theta \right) A_{ij}^{t},\mathrm{clip}\left( \rho _{ij}^{t}\left( \theta \right) ,1-\varepsilon ,1+\varepsilon \right) A_{ij}^{t} \right\} +\alpha _k\mathbb{H} _{ij}^{t}}\left( \theta \right) \right\}}},
\label{policy_loss}
\end{align}
where  $y_{ij}:=(a_{ij}^0,...,a_{ij}^{\left| y_{ij} \right|-1})$, $a_{ij}^t$ is the $t$-th token in the sequence $y_{ij}$, $s_{ij}^t:=(x_i,a_{ij}^0,...,a_{ij}^{t-1})$ is the prefix context when generating $a_{ij}^{t}$, $\rho_{ij}^{t}(\theta ):=\frac{\pi _{\theta}\left( a_{ij}^{t}|s_{ij}^{t} \right)}{\pi _{k}\left( a_{ij}^{t}|s_{ij}^{t} \right)}$, $\mathbb{H}_{ij}^t(\theta):=\mathcal{H} \left( \pi _{\theta}\left( \cdot |s_{ij}^{t} \right) \right)$ is the entropy of the generation policy of token $a_{ij}^{t}$, $\alpha_k \ge 0$ is the coefficient of the entropy, $T_k:=\sum_{i\in \tilde{\mathcal{T}}_k}{\sum_{j=1}^M{\left| y_{ij} \right|}}$ is the total number of tokens in the training batch. Meanwhile, we also introduce the following characteristics into the loss function:
\begin{enumerate}
    \item \textbf{Adaptive Entropy Control.} To preserve the model's exploration capability and maintain high learning plasticity, it is common to include an additional entropy loss to prevent entropy collapse. An appropriately weighted entropy loss can enhance generalization. However, our experiments show that selecting a suitable coefficient in advance is often challenging, as the entropy loss is highly sensitive to both the coefficient and the training data. To address this, we introduce an additional hyperparameter, \texttt{tgt-ent}, representing the target entropy. This hyperparameter dynamically adjusts the coefficient $\alpha_k$ based on the difference between the current entropy and the target entropy, ensuring that the current entropy remains lower-bounded by \texttt{tgt-ent}. See Section~\ref{sec:adaptive_entropy} for more details.
    \item \textbf{No KL Loss.} We found that including a KL loss term hinders performance gains, particularly in the later stages of multi-stage training. Therefore, we omit the KL loss from our training recipe. See Section~\ref{sec:kl} for further discussion.
\end{enumerate}

\subsection{Effectiveness of MAGIC Components}
In this section, we present results from extensive experiments conducted to examine how various components of our MAGIC recipe influence the performance improvement of reinforcement learning during post-training.

\subsubsection{Data Mixture}
\label{sec:data_mixture}
\begin{figure}[h]
    \centering
    \subfigure[]{\includegraphics[width=0.47\textwidth]{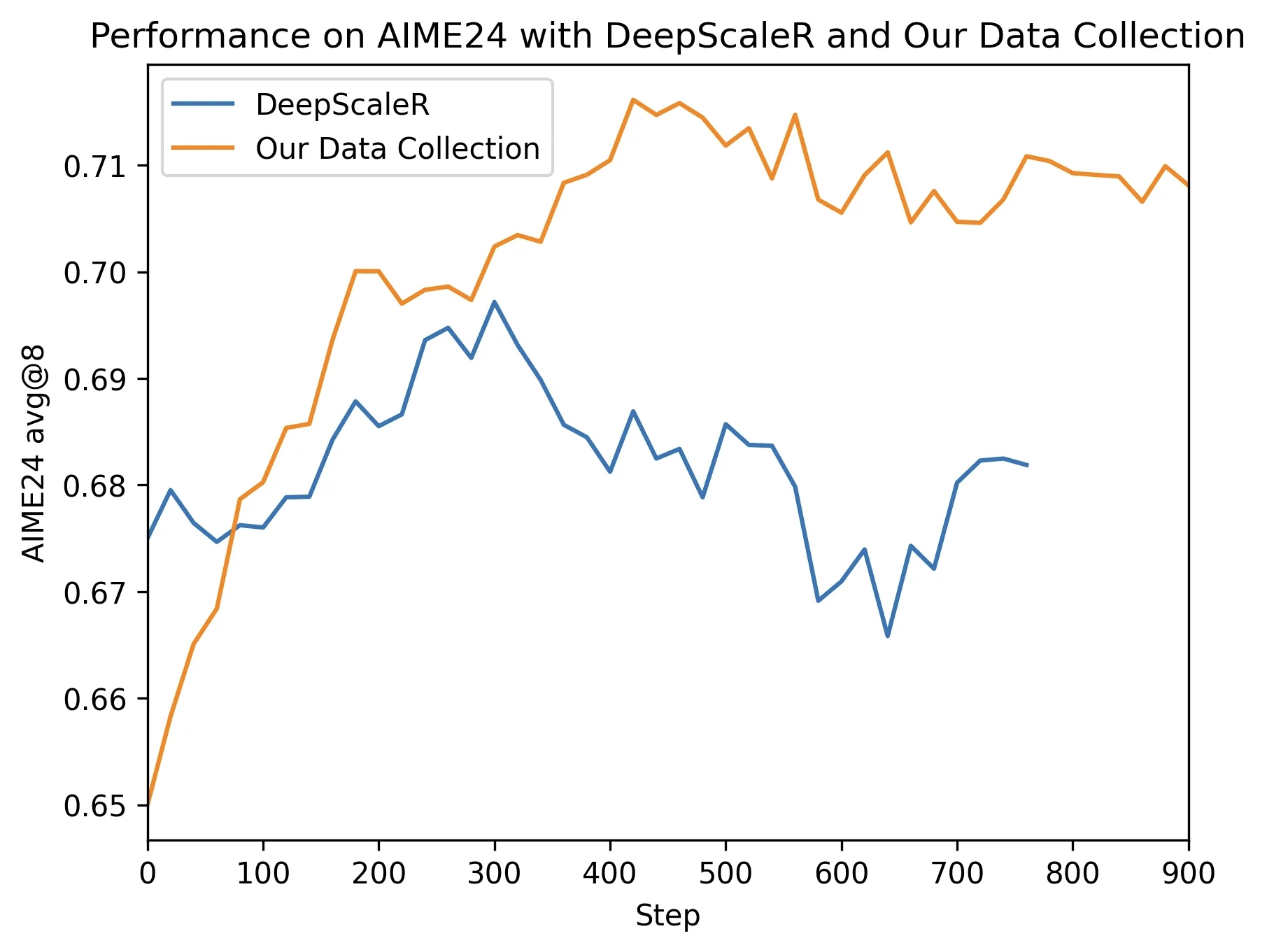}
    \label{fig:data_mixture_1}}
    \subfigure[]{\includegraphics[width=0.46\textwidth]{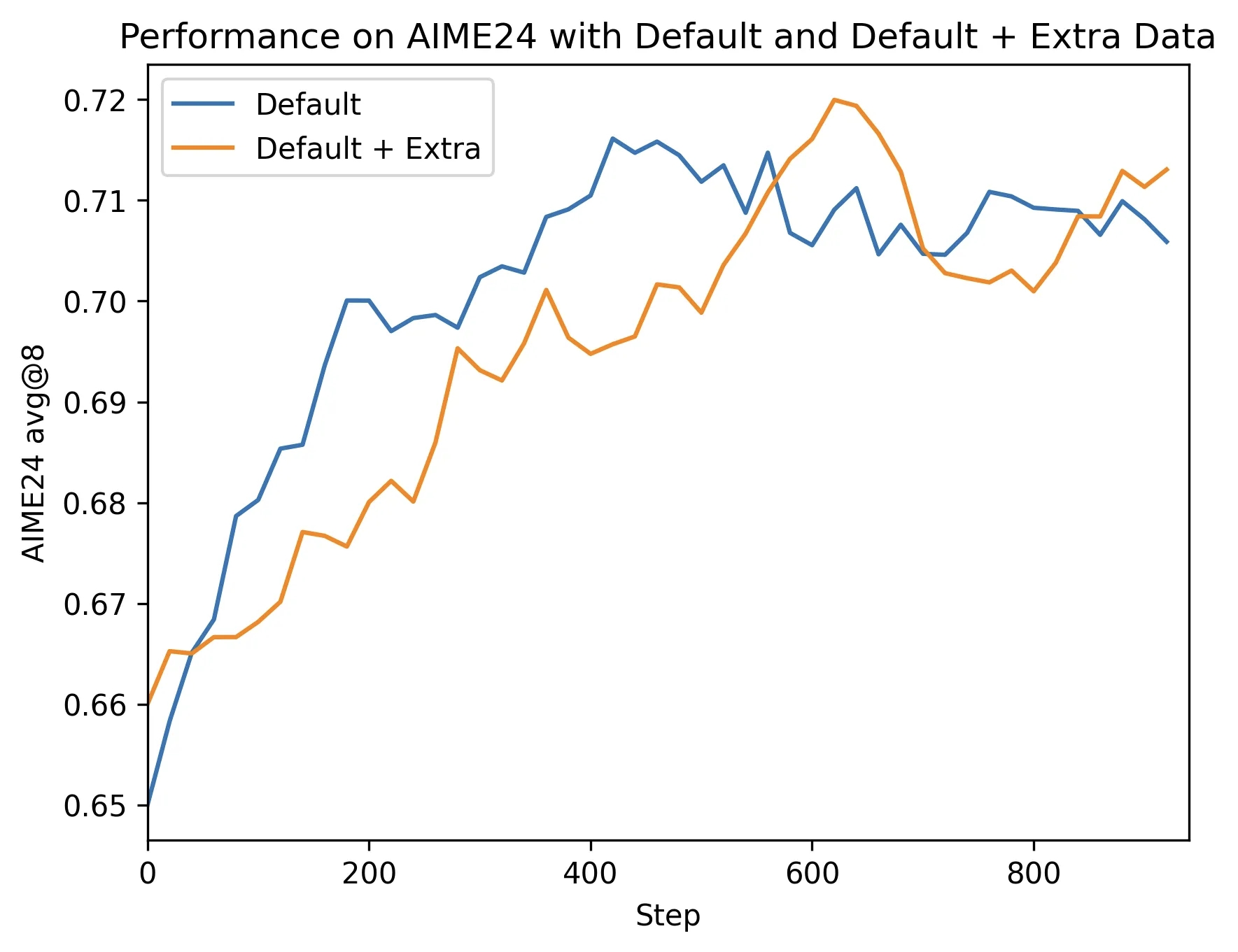}
    \label{fig:data_mixture_2}}
    \vspace{-1em}
    \caption{\textbf{Left:} Comparison of our data mixture with DeepScaleR's mixture. The experiment was conducted on an earlier version of the 32B variant, using only math data. \textbf{Right:} Comparison of AIME 24 performance between two mixtures: our official mixture (default) and a version with additional data selected using lower verification criteria (i.e., with potential errors in ground truth answers). Although the quality is lower, we observe only slower learning progress compared to the clean counterpart.}
    \label{fig:data_mixture}
\end{figure}
In our formal training recipe, we include additional hard problems filtered from NuminaMath-1.5 \cite{numina_math_datasets} to construct our final data mixture. We conduct the following ablation study to demonstrate the effectiveness of this design choice. We primarily compare against DeepScaleR's data mixture \cite{deepscaler2025}, as existing models trained on it have shown strong performance.

\stepcounter{ablation}
\newtcolorbox{mybox}[2][]{colbacktitle=red!10!white, colback=blue!10!white,coltitle=red!70!black, title={#2},fonttitle=\bfseries,#1}
\begin{mybox}{Ablation Experiments \arabic{ablation}: Existing Mixture vs. Our Data Mixture}
\begin{enumerate}[leftmargin=15pt,rightmargin=5pt,itemsep=5pt]
    \item DeepScaleR mixture \cite{deepscaler2025}: Comprises problems from previous years' AIME, AMC, Omni-MATH \cite{gao2024omnimathuniversalolympiadlevel}, and STILL \cite{Slow_Thinking_with_LLMs_3_Preview}.
    \item Skywork-OR1 mixture: Our custom mixture described in Section~\ref{sec:dataset}, incorporating problems from more diverse sources (e.g., NuminaMath-1.5) and selected via difficulty filtering and quality control.
\end{enumerate}
We use the same hyperparameters and approximately the same number of training steps across both experiments to control for the effect of data size. Results are shown in Figure~\ref{fig:data_mixture}.
\vspace{5pt}
\end{mybox} 

Although the DeepScaleR dataset performs well with smaller model variants, we observed a slight initial improvement on AIME24. However, performance degraded sharply after 300 training steps, eventually returning to the same accuracy as before training. Additionally, in Figure~\ref{fig:data_mixture_2}, we test our data mixture combined with an \textit{extra subset} obtained via a less stringent verification procedure. This extra subset contains hard problems from NuminaMath-1.5 that were previously excluded due to potential mismatches between extracted and provided solutions. We find that the performance difference between the two mixtures is negligible within the first 900 steps. The version including the extra subset exhibits slightly slower early progress, possibly due to noise in the provided answers. We hypothesize that RL training is robust to small amounts of ground truth noise, consistent with findings in \cite{zuo2025ttrl}. \textbf{Therefore, we adopt the default data composition described in Section~\ref{sec:dataset} for all subsequent exploration experiments.}

\subsubsection{Multi-Stage Training}
\label{sec:multi_stage}

\begin{figure}[t]
    \centering
    \subfigure[]{\includegraphics[width=0.45\textwidth]{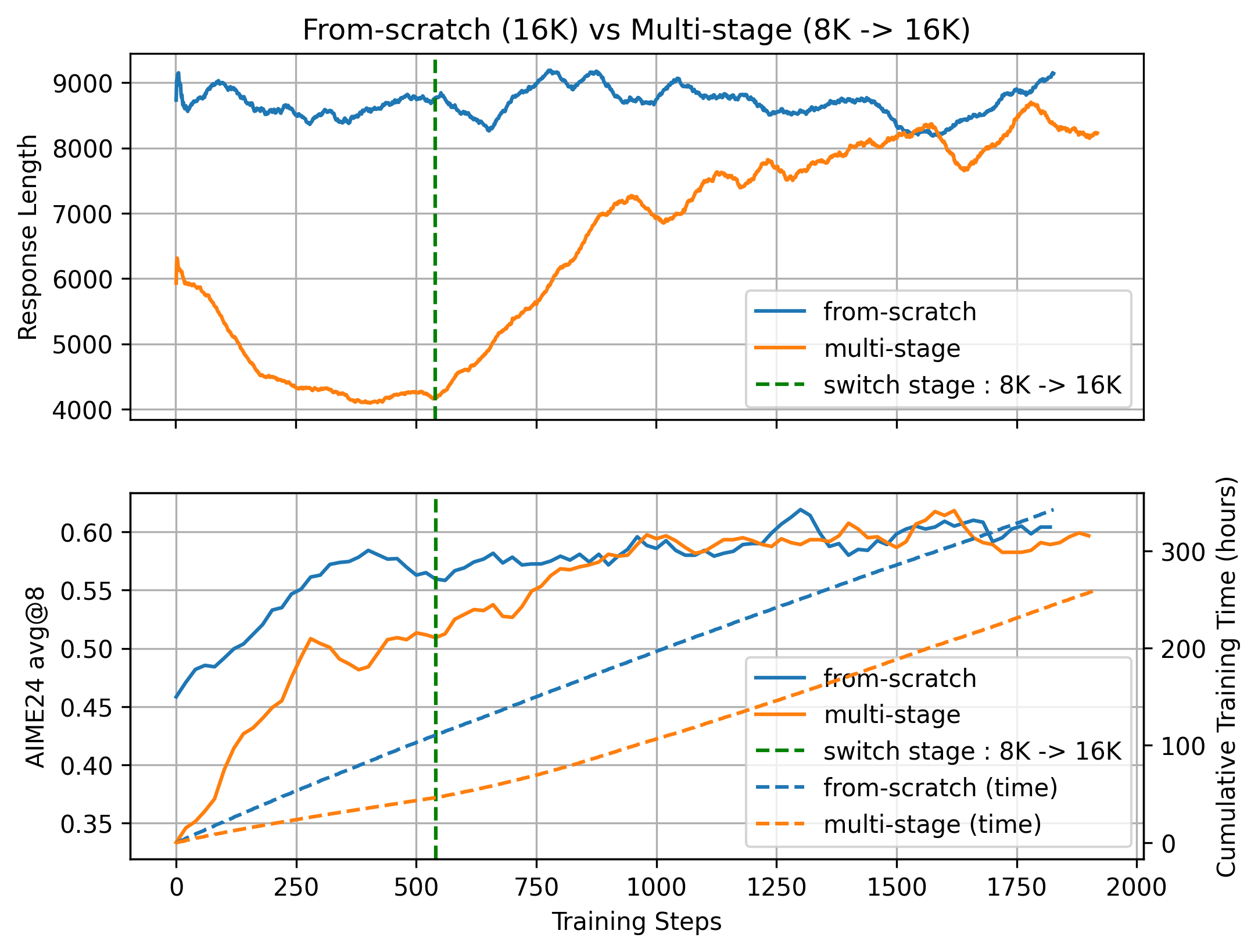}
    \label{fig:multi_stage_vs_from_scratch}
    }
    \hfill
    \subfigure[]{\includegraphics[width=0.5\textwidth, height=5.6cm]{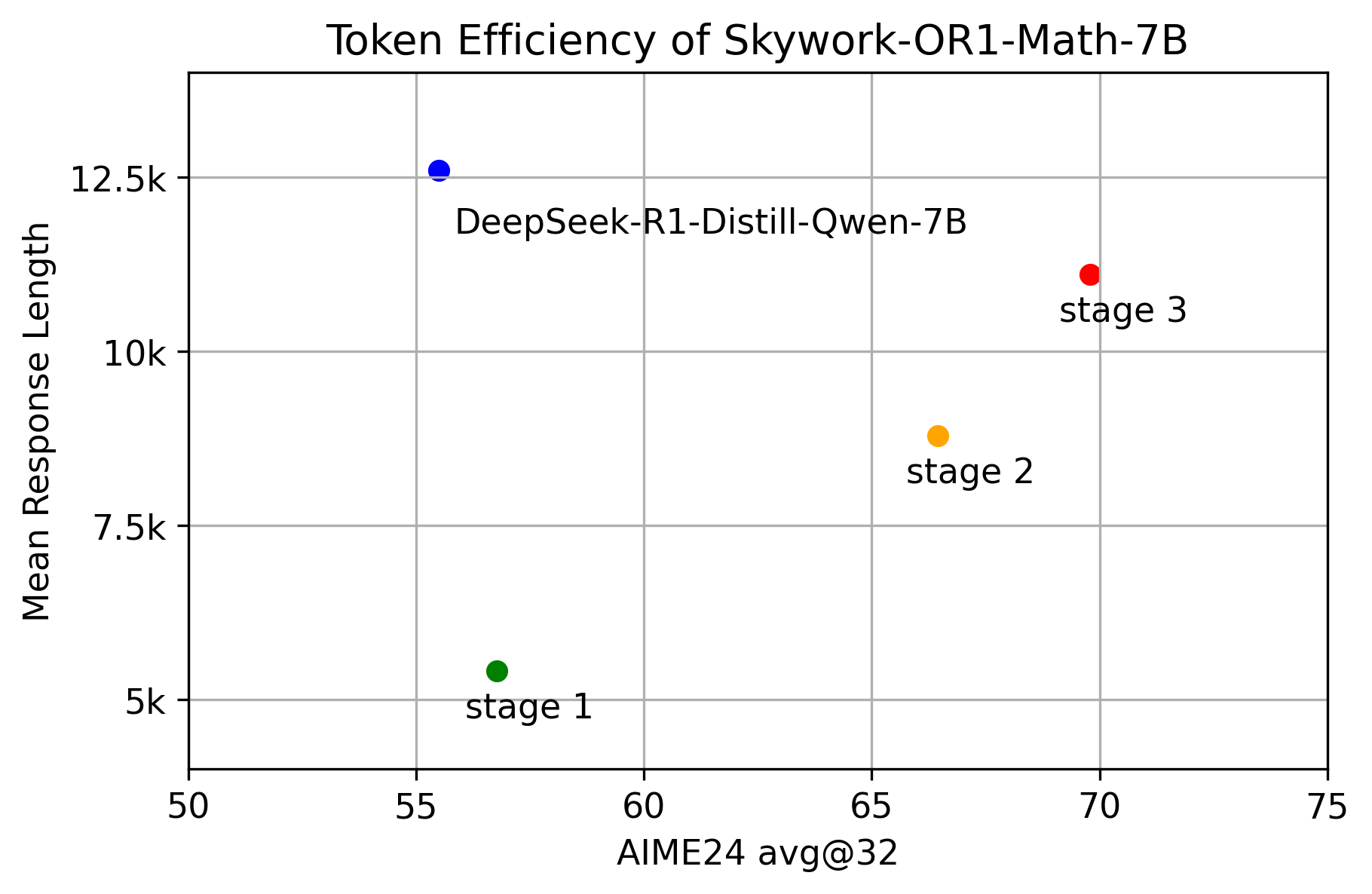}
    \label{fig:token_efficiency}}
    \vspace{-3mm}
    \caption{
    \textbf{Left:} Comparison of \textbf{From-Scratch} vs. \textbf{Multi-Stage} training. \textbf{Top left:} Response length during RL training. \textbf{Bottom left:} AIME24 avg@8 performance at temperature 1 (left y-axis) and cumulative training hours (right y-axis). Multi-stage training achieves the same final accuracy with significantly fewer training hours due to a smaller context length in the early stages. 
    \textbf{Right:} AIME24 avg@32 vs. response length for Skywork-OR1-Math-7B and DeepSeek-R1-Distill-Qwen-7B with 32K context length. The \emph{Stage I} checkpoint of Skywork-OR1-Math-7B reaches comparable performance to DeepSeek-R1-Distill-Qwen-7B with notably better token efficiency; further performance gains are seen in \emph{Stages II}\&\emph{III}.
    }
    \vspace{-4mm}
\end{figure}

One of the major challenges in optimizing long Chain-of-Thought (CoT) models with RL is managing excessively long outputs, which can lead to slow convergence and high training variance. Inspired by DeepScaleR \cite{deepscaler2025}, we incorporated multi-stage training in all our released models to improve training efficiency. Specifically, we used a shorter context length $T$ in the initial stages. Once the model's performance converged, we increased $T$ in the subsequent stage. This approach led to significant performance improvements on benchmarks while also enhancing training efficiency.

\textbf{Same Improvement, Higher Efficiency.} To demonstrate the effectiveness of multi-stage training, we conducted two experiments based on DeepSeek-R1-Distill-Qwen-7B with different schedules for $T$:

% \newtcolorbox{mybox}[2][]{colbacktitle=red!10!white, colback=blue!10!white,coltitle=red!70!black, title={#2},fonttitle=\bfseries,#1}
\stepcounter{ablation}
\begin{mybox}{Ablation Experiments \arabic{ablation}: From-Scratch vs. Multi-Stage}
\begin{enumerate}[leftmargin=15pt,rightmargin=5pt,itemsep=5pt]
\item From-Scratch: We started with $T=16\text{K}$ at step 0 and kept it fixed during training.
\item Multi-Stage: We started with $T=8\text{K}$ at step 0. At a later step (i.e., step 540), we switched to \emph{Stage II} and increased $T$ to $16K$. 
\end{enumerate}
\vspace{5pt}
The other hyper-parameters were kept same for both experiments and are reported in Table \ref{table:shared_parameter_in_ablation_1}. The results are presented in Figure \ref{fig:multi_stage_vs_from_scratch} and Figure \ref{fig:token_efficiency}. 
\end{mybox} 
\begin{table}[!htbp]
\centering
\begin{tabular}{@{\extracolsep{4pt}}ccccccc}
\toprule   
Batch Size & Mini-batch Size & Group Size  & Entropy Control & KL Loss \\
\midrule
64 &  32 & 16 & target-entropy 0.2  & No
\\
\bottomrule
\end{tabular}
\caption{Shared hyperparameters in Ablation Experiments 1 based on Deepseek-R1-Distill-Qwen-7B.} 
\label{table:shared_parameter_in_ablation_1}
\end{table}

Figure~\ref{fig:multi_stage_vs_from_scratch} illustrates how AIME24 accuracy, generated response length, and cumulative training hours evolve with the number of training steps in Ablation Experiments 2. As shown, the AIME24 accuracy in both experiments converges to approximately 60 when the number of training steps is sufficiently large. However, in the multi-stage experiment, the context length in \emph{Stage I} (i.e., 8\text{K}) is only half that used in the from-scratch experiment (i.e., 16\text{K}). As a result, the average response length in the multi-stage experiment is significantly shorter during \emph{Stage I} and the initial steps of \emph{Stage II}, leading to more efficient training due to reduced inference and computational costs (approximately 100 training hours are saved over 1000 training steps). After transitioning to \emph{Stage II}, both the response length and AIME24 accuracy begin to increase immediately. Within roughly 500 training steps in \emph{Stage II}, the accuracy of the multi-stage experiment reaches the same level as that of the from-scratch experiment.

% \begin{figure}[!htbp]
%     \centering
%     \includegraphics[width=0.75\linewidth]{figures/from_scratch_vs_multi_stage_combined.png}
%     \caption{\textbf{from-scratch} vs \textbf{multi-stage}. \textbf{Top}: The response length of generated samples during RL training. \textbf{Bottom}: The AIME24 avg@8 performance at temperature 1 (left y-axis) and the cumulative training hours (right y-axis) during RL training. Multi-stage training finally achieves the same AIME24 accuracy but at a significant less training hours due the small context length in stage 1.}
%     \label{fig:multi_stage_vs_from_scratch}
% \end{figure}

% \begin{figure}[!htbp]
%     \centering
%     \includegraphics[width=0.75\linewidth]{figures/token_efficiency_of_math_7b_full_stages.png}
%     \caption{The avg@32 performance vs. response length on AIME24 for Skywork-OR1-Math-7B and DeepSeek-R1-Distill-Qwen-7B. The Stage 1 checkpoint of Skywork-OR1-Math-7B achieves comparable performance on AIME24 to DeepSeek-R1-Distill-Qwen-7B with significantly improved token efficiency. Performance on AIME24 is further enhanced in Stages 2/3 as response length increases.}
%     \label{fig:token_efficiency}
% \end{figure}

\textbf{Improving Token Efficiency While Preserving Scaling Potential.} Truncated responses are labeled as negative samples in RL training because they lack final answers. A potential concern with multi-stage training is that using short context windows may bias the model toward generating shorter responses, potentially limiting its exploratory capacity and reducing its ability to solve complex problems. \textbf{Our findings demonstrate that multi-stage training not only improves token efficiency in the initial stage but also preserves scaling ability.} In Figure~\ref{fig:token_efficiency}, we observe that training with an 8\text{K} context length in \emph{Stage I} maintains comparable AIME24 accuracy under a 32\text{K} context length while significantly improving token efficiency (reducing the average response length from approximately 12.5\text{K} to 5.4\text{K} tokens). In \emph{Stages II and III}, Skywork-OR1-Math-7B steadily increases response length while concurrently improving performance.

\subsubsection{Advantage Mask for Truncated Responses}
\label{sec:adv_mask}
In practice, responses are sampled within a fixed context length $T$. When response lengths exceed $T$, the outcomes cannot be derived, and accuracy rewards are set to 0, resulting in negative advantages for these truncated responses, which may introduce bias. To mitigate this issue, we investigated several advantage mask strategies aimed at reducing the influence of truncated responses. However, our findings show that assigning negative advantages to truncated samples not only improves token efficiency but also preserves the model's scaling ability in later stages. As a result, we did not apply any mask strategies in our final training pipeline.

\begin{figure}[!htbp]
    \centering
    \includegraphics[width=0.65\linewidth]{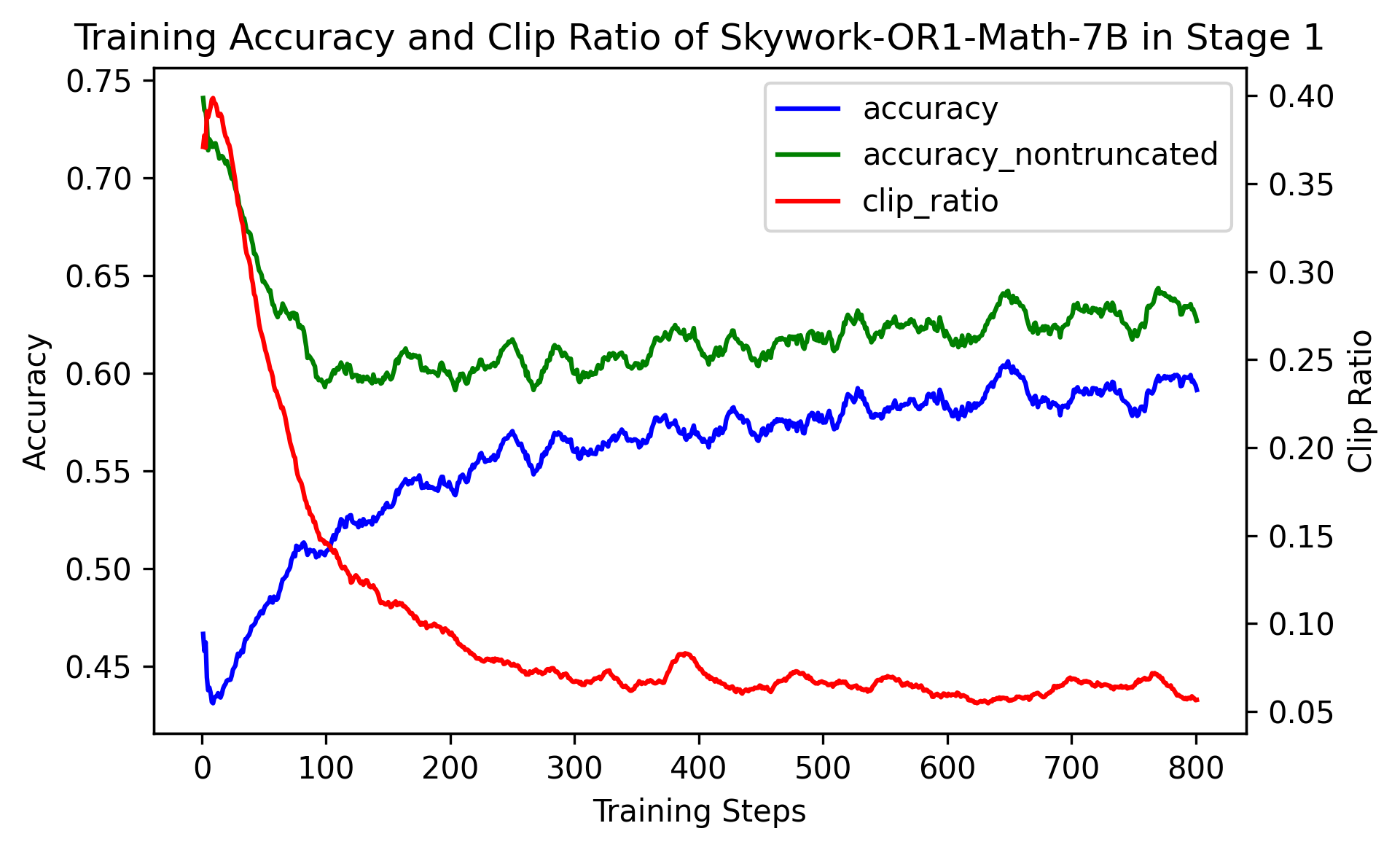}
    \vspace{-1em}
    \caption{Training accuracy and clip ratio during RL training of Skywork-OR1-Math-7B in \emph{Stage I}. \textcolor{blue}{accuracy}: Mean accuracy reward on training batch. \textcolor{ForestGreen}{accuracy\_nontruncated}: Mean accuracy of non-truncated samples. \textcolor{red}{clip\_ratio}: Ratio of truncated responses.}
    \label{fig:adv_mask_stage1}
\end{figure}

\textbf{Two Optimization Directions in Short Context Length.} In our \emph{Stage I} training of Skywork-OR1-Math-7B, we set the context length to $T=8$K, and approximately 40\% of responses were truncated at the initial steps. Although overall training accuracy continued to increase during RL training, we observed that the accuracy of non-truncated samples initially declined sharply within the first 100 training steps before showing a slight upward trend. See Figure~\ref{fig:adv_mask_stage1} for details. A truncated response typically receives an accuracy reward of 0 because the final answer is missing due to truncation, even if it would be correct if fully generated. Therefore, reducing the number of truncated responses improves achievable accuracy. Figure~\ref{fig:adv_mask_stage1} shows that the initial increase in training accuracy (steps 0-100) is primarily due to a sharp decrease in the clip ratio. After step 100, the algorithm begins to improve accuracy for non-truncated responses as well.

% \begin{figure}[!htbp]
%     \centering
%     \includegraphics[width=0.75\linewidth]{figures/adv_mask_stage1.png}
%     \caption{The training accuracy and clip ratio during the RL training of Skywork-OR1-Math-7B in stage 1. Line \textcolor{blue}{accuracy} refers to the mean accuracy reward on training batch. Line \textcolor{ForestGreen}{accuracy\underline{ }nontruncated} is the mean accuracy reward of the non-truncated samples in training batch. Line \textcolor{red}{clip\underline{ }ratio} is the ratio of how many responses are truncated. }
%     \label{fig:adv_mask_stage1}
% \end{figure}

\textbf{A Brief Explanation from a Theoretical Perspective.} We now use mathematical language to clarify this phenomenon further in a formal way. Recall the objective of RL training in \eqref{RL_obj},
\begin{align*}
\pi ^*\in \underset{\pi}{\mathrm{argmax}}\,\,\{\mathcal{J}(\pi):=\mathbb{E} _{x\sim \mathcal{D}}\mathbb{E} _{y\sim \pi \left( \cdot |x \right)}\left[ r\left( x,y \right) \right]\}, 
\end{align*}
where $x$ is the prompt, $\mathcal{D}$ is the distribution of prompts, $y$ is the response sampled from actor $\pi$, $r(x,y)\in\{0,1\}$ is the binary accuracy reward. Note that the response $y$ is  sampled under the context length $T$ . For these truncated responses whose lengths are greater than $T$, i.e. $|y|>T$, the accuracy reward is $r(x,y)=0$ since the outcome can not be derived from the response. Based on this observation, one can easily shows that the objective function $\mathcal{J}(\pi)$ satisfies 
\begin{align*}\mathcal{J} (\pi )&=\mathbb{E} _{x\sim \mathcal{D}}\mathbb{E} _{y\sim \pi \left( \cdot |x \right)}\left[ r\left( x,y \right) \right] 
\\
&=\mathbb{E} _{x\sim \mathcal{D}}\mathbb{E} _{y\sim \pi \left( \cdot |x \right)}\left[ r\left( x,y \right) \mathbb{I} \left\{ \left|y\right|\le T \right\} \right] 
\\
\,\,      &=\mathbb{E} _{x\sim \mathcal{D}}\left[ p_{\mathrm{non}-\mathrm{trunc}}^{\pi}\left( x \right) \mathbb{E} _{y\sim \pi \left( \cdot |x \right)}\left[ \frac{\mathbb{I} \left\{ \left|y\right|\le T \right\}}{p_{\mathrm{non}-\mathrm{trunc}}^{\pi}\left( x \right)}r\left( x,y \right) \right] \right] 
\\
\,\,      &=\mathbb{E} _{x\sim \mathcal{D}}\left[ p_{\mathrm{non}-\mathrm{trunc}}^{\pi}\left( x \right) \mathbb{E} _{y\sim \hat{\pi}_{T}\left( \cdot |x \right)}\left[ r\left( x,y \right) \right] \right] 
\\
\,\,      &=\mathbb{E} _{x\sim \mathcal{D}}\left[ p_{\mathrm{non}-\mathrm{trunc}}^{\pi}\left( x \right) \bar{r}_{\mathrm{non}-\mathrm{trunc}}^{\pi}\left( x \right) \right],
\end{align*}
where $p^\pi_{\mathrm{non-trunc}}\left( x \right) :=\mathbb{P} _{y\sim \pi \left( \cdot |x \right)}\left( \left|y\right|\le T \right)$  is the probability that a response $y$ is not truncated by the limit of context length $T$ (we assume $p^\pi_{\mathrm{non-trunc}}\left( x \right)>0$ for simplicity), $\bar{r}_{\mathrm{non-trunc}}^{\pi}\left( x \right) :=\mathbb{E} _{y\sim \hat{\pi}_{T}\left( \cdot |x \right)}\left[ r\left( x,y \right) \right]$  is the accuracy of the non-truncated responses output by policy $\pi$ and $
\hat{\pi}_{T}\left( y|x \right) :=\frac{\pi \left( y|x \right)}{p_{\mathrm{non}-\mathrm{trunc}}^{\pi}\left( x \right)}\mathbb{I} \left\{ \left| y \right|\le T \right\}.$ This implies that the accuracy on training distribution, i.e. $\mathcal{J}(\pi)$, can be increased by:
\begin{itemize}
    \item increasing $p_{\mathrm{non-trunc}}^{\pi}\left( x \right)$, which means the number of the responses that receive accuracy reward of 0 erroneously decreases.
    \item increasing $r_{\mathrm{non-trunc}}^{\pi}\left( x \right)$, which means the response quality within the context length will be improved.
\end{itemize}

\begin{figure}[t]
    \centering
    \subfigure[]{\includegraphics[width=0.32\textwidth]{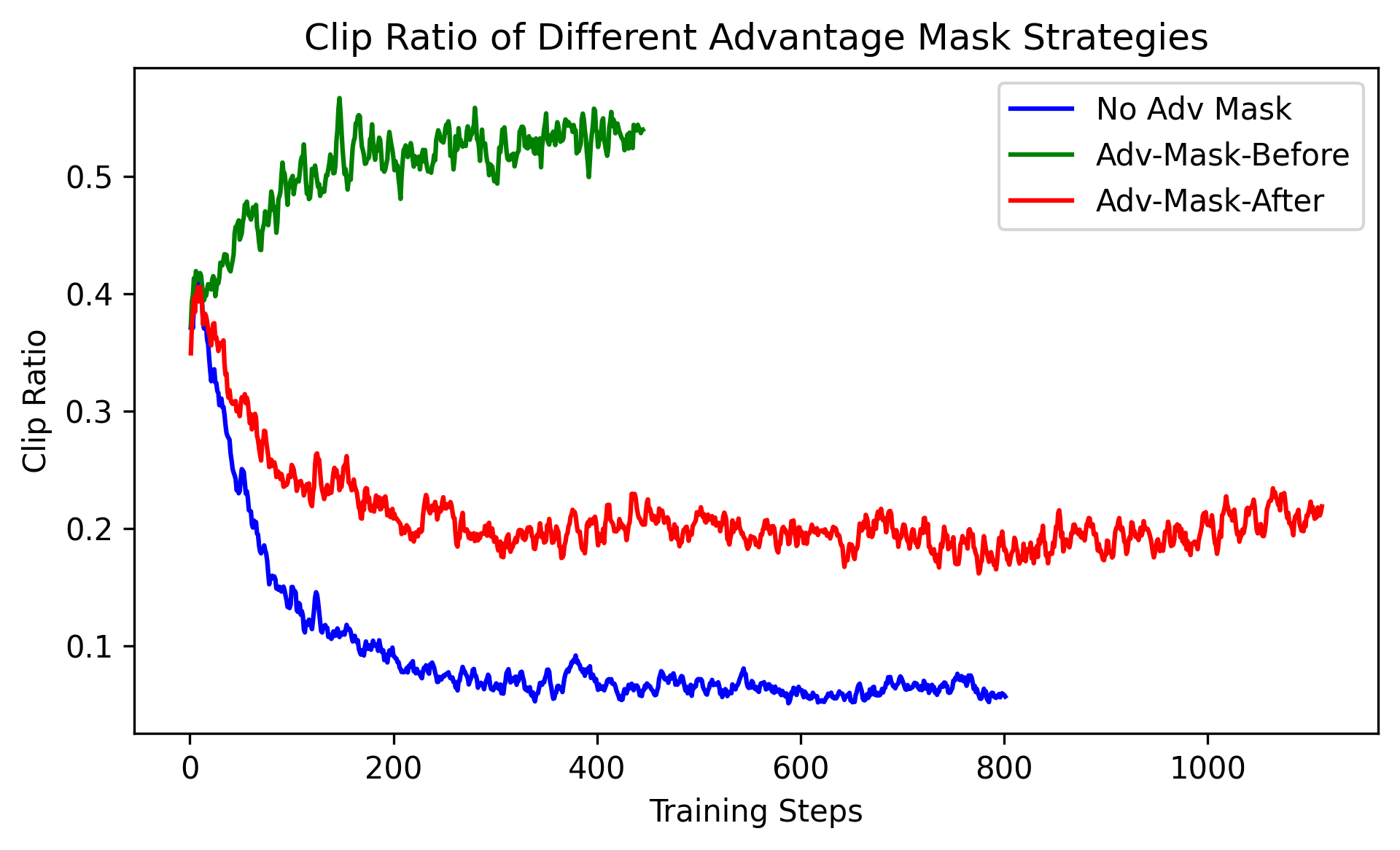} \label{fig:adv_mask_compare_clip_ratio}}
    \subfigure[]{\includegraphics[width=0.32\textwidth]{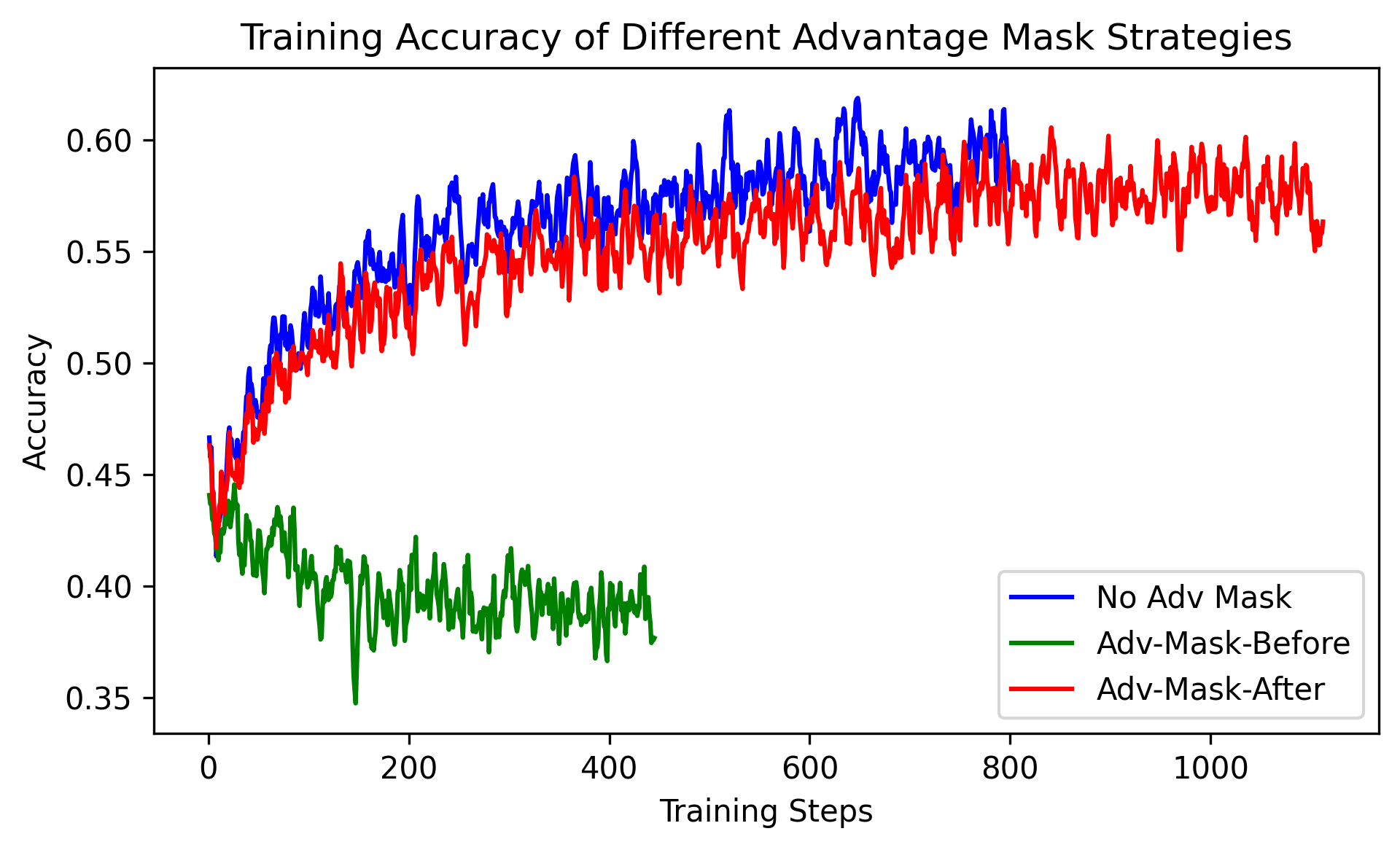}}
    \subfigure[]{\includegraphics[width=0.32\textwidth]{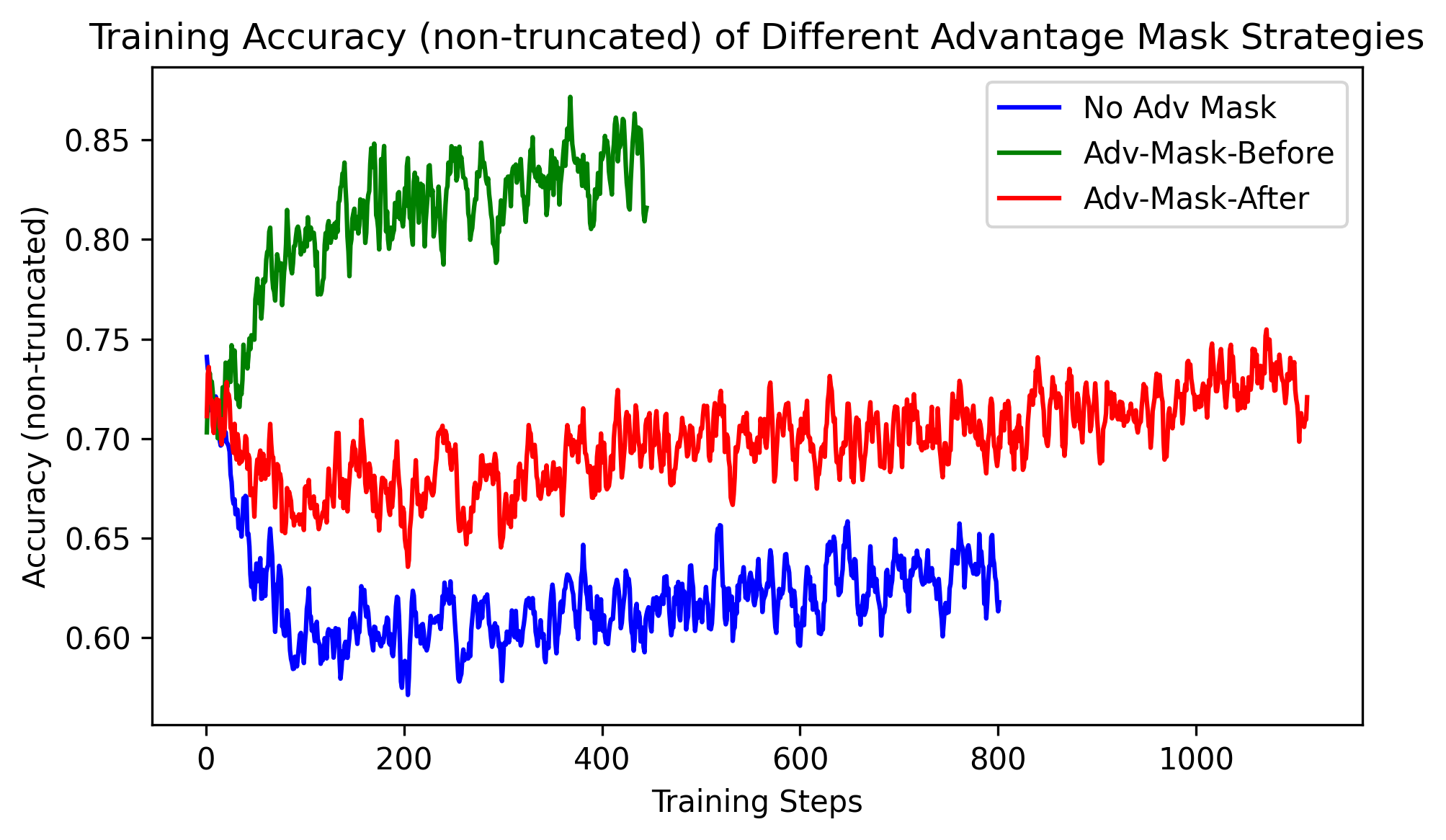}
    \label{fig:adv_mask_compare_acc}
    }
    \vspace{-1em}
    \caption{\textbf{Left:} The clip ratio of generated responses during reinforcement learning training was analyzed after applying various advantage mask strategies in Ablation Experiments 2. Using an advantage mask mitigates the decay in response length. The clip ratio even increased after applying Adv-Mask-Before.
    \textbf{Middle:}
    Training accuracy of responses influenced by different advantage mask strategies in Ablation Experiment 2 shows distinct patterns. After applying the Adv-Mask-Before, training accuracy decreases. In contrast, it continues to increase when using the Adv-Mask-After or No-Adv-Mask strategies.
    \textbf{Right:} 
    Training accuracy of non-truncated responses induced by different advantage mask strategies in Ablation Experiment 2 showed distinct outcomes. After applying the Adv-Mask-Before strategy, the training accuracy of non-truncated responses continued to rise. In contrast, both the Adv-Mask-After and No-Adv-Mask strategies resulted in a sharp decrease during the early steps. }
    \label{fig:adv_mask}
\end{figure}

\textbf{Advantage Mask for Truncated Responses.} To encourage the algorithm to focus on optimizing accuracy within the context length -- i.e., increasing $r_{\mathrm{non-trunc}}^{\pi}\left( x \right)$ -- rather than merely shortening responses to avoid erroneously receiving a zero accuracy reward -- i.e., increasing $p_{\mathrm{non-trunc}}^{\pi}\left( x \right)$ -- we explored various advantage mask strategies. These strategies were designed to mitigate the impact of noisy training signals introduced by truncated samples. We conducted ablation experiments using DeepSeek-R1-Distill-Qwen-7B in \emph{Stage I} to evaluate the effects of different advantage mask strategies.

\begin{figure}
    \centering
    \includegraphics[width=0.65\linewidth]{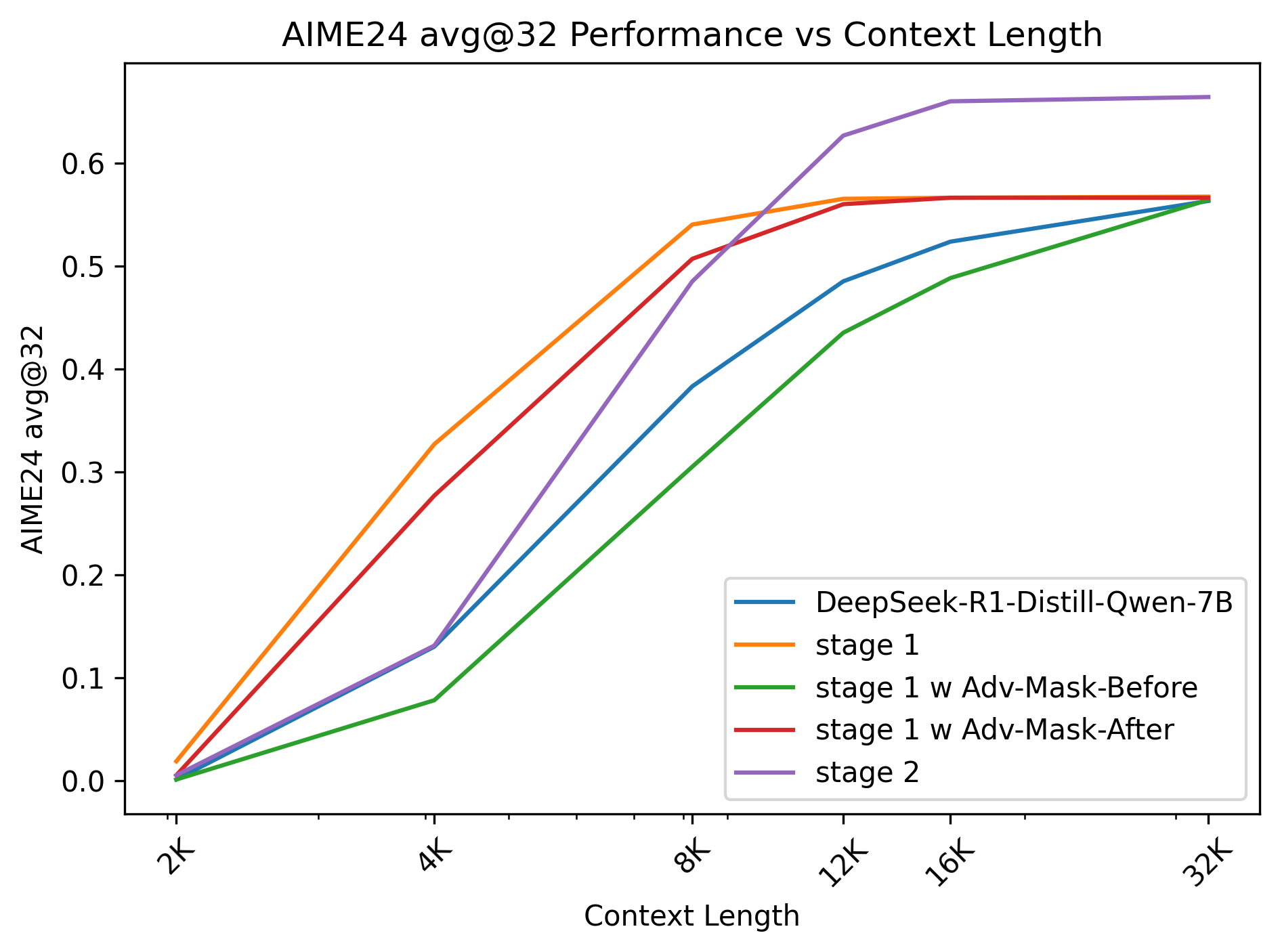}
    \vspace{-1em}
    \caption{AIME24 avg@32 performance vs. context length for different advantage mask strategies in Ablation Experiments \arabic{ablation}. All strategies achieve the same accuracy at the 32K context length. The accuracy was further improved after the training of \emph{Stage II} even though the noisy training signals from truncated responses were introduced in \emph{Stage I}.}
    \label{fig:adv_mask_compare_scaling_law}
\end{figure}

\stepcounter{ablation}
\begin{mybox}{Ablation Experiments \arabic{ablation}: Different Advantage Mask Strategies}
\begin{enumerate}[leftmargin=15pt,rightmargin=5pt,itemsep=5pt]
\item No-Adv-Mask:  We do not employ any advantage mask strategy.
\item Adv-Mask-Before: The truncated responses are \textbf{not involved} in the group advantage calculation for non-truncated responses, and the advantage of these truncated responses are set to 0 (thus not contributing to the policy loss):
\begin{align*}
\forall t : A_{ij}^t=\begin{cases}
	\frac{r\left( x_i,y_{ij} \right) -\mathrm{mean}\left( \hat{\mathbb{R}}_i \right)}{\mathrm{std}\left( \hat{\mathbb{R}}_i \right)}&		\left| y \right|\le T\\
	0&		\left| y \right|>T\\
\end{cases}
\end{align*}
Here $\hat{\mathbb{R}}_i$ is the accuracy rewards group of \textbf{non-truncated} responses of prompt $x_i$.
\item Adv-Mask-After: The truncated responses are \textbf{still involved} in the group advantage calculation for non-truncated responses, and the advantage of these truncated responses are set to 0 (thus not contributing to the policy loss):
\begin{align*}
\forall t : A_{ij}^t=\begin{cases}
	\frac{r\left( x_i,y_{ij} \right) -\mathrm{mean}\left( \mathbb{R}_i \right)}{\mathrm{std}\left( \mathbb{R}_i \right)}&		\left| y \right|\le T\\
	0&		\left| y \right|>T\\
\end{cases}
\end{align*}
Here $\mathbb{R}_i$ is the accuracy rewards group of \textbf{all} responses of prompt $x_i$.
\end{enumerate}
\vspace{5pt}
The other hyperparameters remain the same for both experiments and are reported in Table \ref{table:shared_parameter_in_ablation_2}. The results can be found in Figure \ref{fig:adv_mask_compare_clip_ratio}, Figure \ref{fig:adv_mask_compare_acc} and Figure \ref{fig:adv_mask_compare_scaling_law}. 
\end{mybox}

\begin{table}[!htbp]
\centering
\begin{tabular}{@{\extracolsep{4pt}}ccccccc}
\toprule   
Batch Size & Mini-batch Size & Group Size & Context Length $T$  & Entropy Control & KL Loss \\
\midrule
 256 &  128 & 16 & \emph{Stage I} $8K$  & target-entropy 0.2  & No
\\
\bottomrule
\end{tabular}
\caption{Shared hyperparameters in Ablation Experiments 2 based on Deepseek-R1-Distill-Qwen-7B.} 
\label{table:shared_parameter_in_ablation_2}
\end{table}

Figure~\ref{fig:adv_mask} shows the clip ratio, overall accuracy, and accuracy on non-truncated responses in Ablation Experiments 2. We observe that although the response quality within the context length (i.e., the accuracy of non-truncated responses) increases as expected after applying the Adv-Mask-Before strategy, the overall training accuracy continues to decline, and the clip ratio increases steadily. \textbf{This appears to be a form of reward hacking from our perspective.} More importantly, as shown later in Figure~\ref{fig:adv_mask_compare_scaling_law}, the accuracy of the Adv-Mask-Before strategy under large context lengths -- where responses are typically not truncated (e.g., 32K) -- shows no improvement. This may be attributed to the smaller effective training batch size caused by the increased clip ratio under the Adv-Mask-Before strategy. The behavior of Adv-Mask-After serves as an intermediate point between Adv-Mask-Before and No-Adv-Mask.

\textbf{Advantage Mask Does Not Exhibit Better Performance Given a Larger Inference Budget.} Although Ablation Experiments 2 demonstrate that $\bar{r}_{\mathrm{untrunc}}^{\pi}\left( x \right)$ is optimized under short context lengths when applying advantage masks, we find that accuracy does not improve when the context length is large enough to avoid truncation (i.e., 32K). We compare the test-time scaling behavior on AIME24 for models trained with different advantage mask strategies (see Figure~\ref{fig:adv_mask_compare_scaling_law}). The results show that applying an advantage mask does not improve test-time scaling behavior in \emph{Stage I}, and accuracy at 32K remains unchanged-even though $\bar{r}_{\mathrm{non-trunc}}^{\pi}\left( x \right)$ is optimized during training. In contrast, RL training without an advantage mask in Stage I not only maintains accuracy at large context lengths but also significantly improves token efficiency. Moreover, the shorter response lengths learned in Stage I do not hinder the simultaneous improvements in both response length and accuracy observed in Stage II. Based on these findings, we did not apply any advantage mask to address noisy training signals from truncated samples in our final training recipe.

\subsubsection{High-temperature Sampling}
\label{sec:temperature}

\begin{figure}[h]
    \centering
    \includegraphics[width=0.75\linewidth]{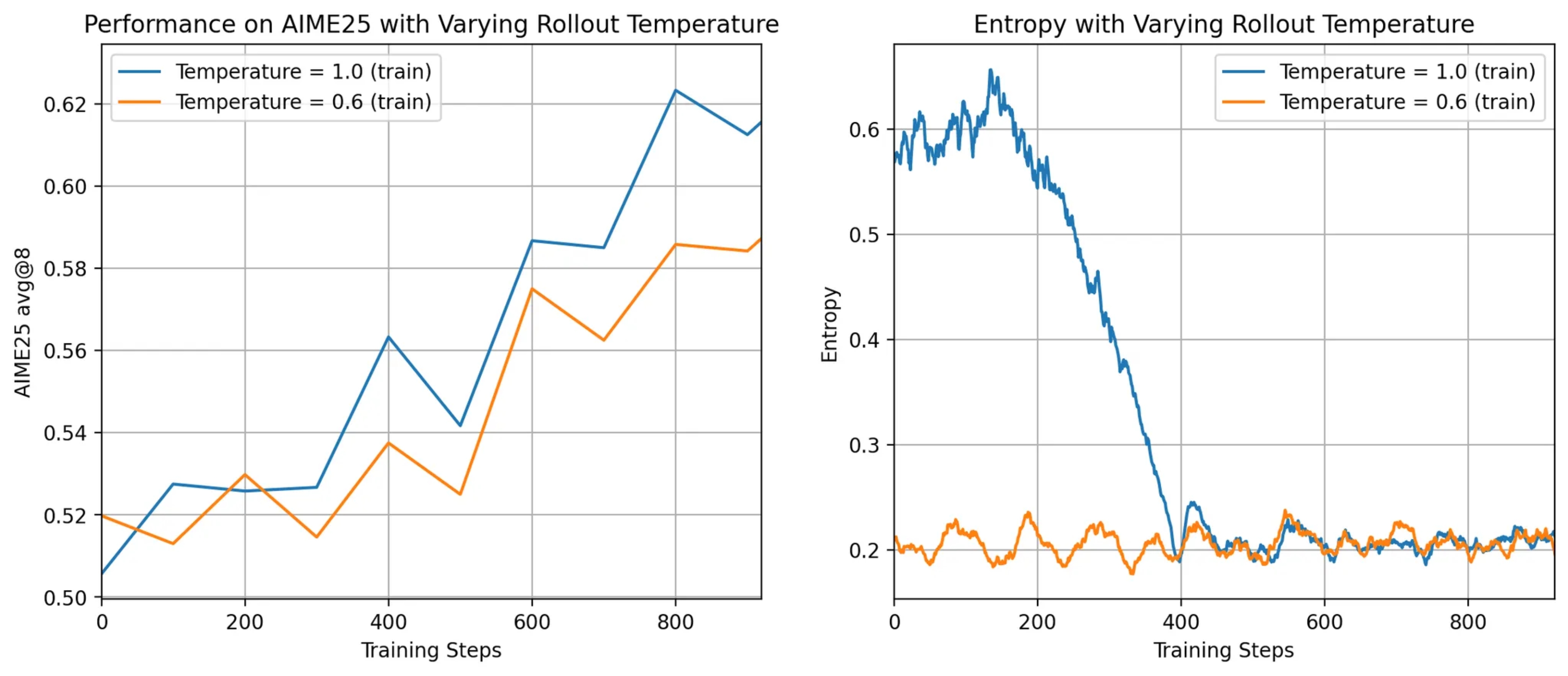}
    \vspace{-1em}
    \caption{AIME25 avg@8 performance and entropy versus the number of training steps in Ablation Experiments 3. Training with a temperature of 0.6 starts with the lowest entropy and learns more slowly than at a temperature of 1.0. Note that the entropy in the right plot remains around 0.2 because adaptive entropy control is enabled. This experiment was conducted on an earlier version of the 32B variant using only math data. Note also that in the left plot, the two temperatures indicate the rollout temperatures used during training. The scores of AIME25 were obtained by evaluating both models at a temperature of 0.6 to ensure a fair comparison.}
    \label{fig:impact_of_tau}
\end{figure}

The group-wise nature of GRPO implies that the sampling procedure for responses directly affects the quality and diversity of each group, which in turn influences learning. Prior work suggests that higher temperatures generally lead to slightly worse performance due to increased randomness. If the temperature is set too high, it may increase the likelihood of sampling groups containing only incorrect responses, thereby reducing training efficiency due to the absence of advantageous signals. On the other hand, using a low temperature reduces group diversity, resulting in solutions that are highly similar or potentially all correct. Therefore, selecting an appropriate temperature is critical to ensure sufficient in-group solution diversity. We conducted ablation experiments on the choice of sampling temperature $\tau$, and the results are presented in Figure~\ref{fig:impact_of_tau}.

\stepcounter{ablation}
\begin{mybox}{Ablation Experiments \arabic{ablation}: Different Online Sampling Temperatures $\tau$}
We compared two different sampling temperatures in online RL training: 
\begin{enumerate}[leftmargin=15pt,rightmargin=5pt,itemsep=5pt]
\vspace{5pt}
\item High Temperature: We set the temperature hyperparameter $\tau=1.0$.
\item Low Temperature: We set the temperature hyperparameter $\tau=0.6$.
\end{enumerate}
\vspace{5pt}
The other hyperparameters were kept the same for both experiments and are reported in Table \ref{table:shared_parameter_in_ablation_3}. The results can be found in Figure \ref{fig:impact_of_tau}. 
\end{mybox} 
\begin{table}[h]
\centering
\begin{tabular}{@{\extracolsep{4pt}}ccccccc}
\toprule   
Batch Size & Mini-batch Size & Group Size & Context Length $T$  & Entropy Control & KL Loss \\
\midrule
 64 & 32 & 16 & \emph{Stage I} 16K & target entropy 0.2  & 0
\\
\bottomrule
\end{tabular}
\caption{Shared hyperparameters in Ablation Experiments \arabic{ablation}}
\label{table:shared_parameter_in_ablation_3}
\end{table}

In our experiments, we identified an additional entropy-related phenomenon: when a low temperature is used (e.g., 0.6), the model either begins with extremely low entropy or its entropy quickly collapses to near zero within approximately 100 steps. This behavior initially slows learning progress and ultimately leads to stagnation. We hypothesize that with a less diverse group of solutions -- despite containing both correct and incorrect responses -- the policy update becomes overly focused on a narrow subset of tokens. This results in a large probability mass being assigned to specific tokens that frequently appear in the sampled responses. When we increased the rollout temperature to 1.0, the model’s initial entropy rose to a more desirable range. Although entropy still eventually converges, the higher temperature substantially enhances the learning signal in the early stages and preserves greater potential for continued training, as shown in the figure above.

\subsubsection{Adaptive Entropy Control}
\label{sec:adaptive_entropy}

Building on the findings from Section~\ref{sec:entropy} -- which suggest that while preventing premature entropy collapse via entropy regularization is beneficial, selecting an appropriate entropy loss coefficient is challenging -- we introduce \textbf{Adaptive Entropy Control}, a method that adaptively adjusts the entropy loss coefficient based on the target and current entropy. Specifically, we introduce two additional hyperparameters: \textbf{tgt-ent} (the desired target entropy) and $\Delta$ (the adjustment step size for the entropy loss coefficient). We initialize the adaptive coefficient with $c_0 = 0$. At each training step $k$, let $e$ denote the current entropy of the actor (estimated from the rollout buffer). If $e$ is less than \textbf{tgt-ent}, we increase $c_k$ by $\Delta$ (i.e., $c_{k+1} = c_k + \Delta$). If $e$ exceeds \textbf{tgt-ent}, we decrease $c_k$ by $\Delta$. To alleviate instability caused by unnecessary entropy loss, we activate the entropy loss only when $e \le$ \textbf{tgt-ent}, i.e., $\alpha_k = c_k \cdot \mathbb{I}\{e \le \text{\textbf{tgt-ent}}\}$, ensuring that the current entropy remains lower-bounded by the target entropy.

\begin{figure}[!htbp]
    \centering
    \includegraphics[width=0.95\linewidth]{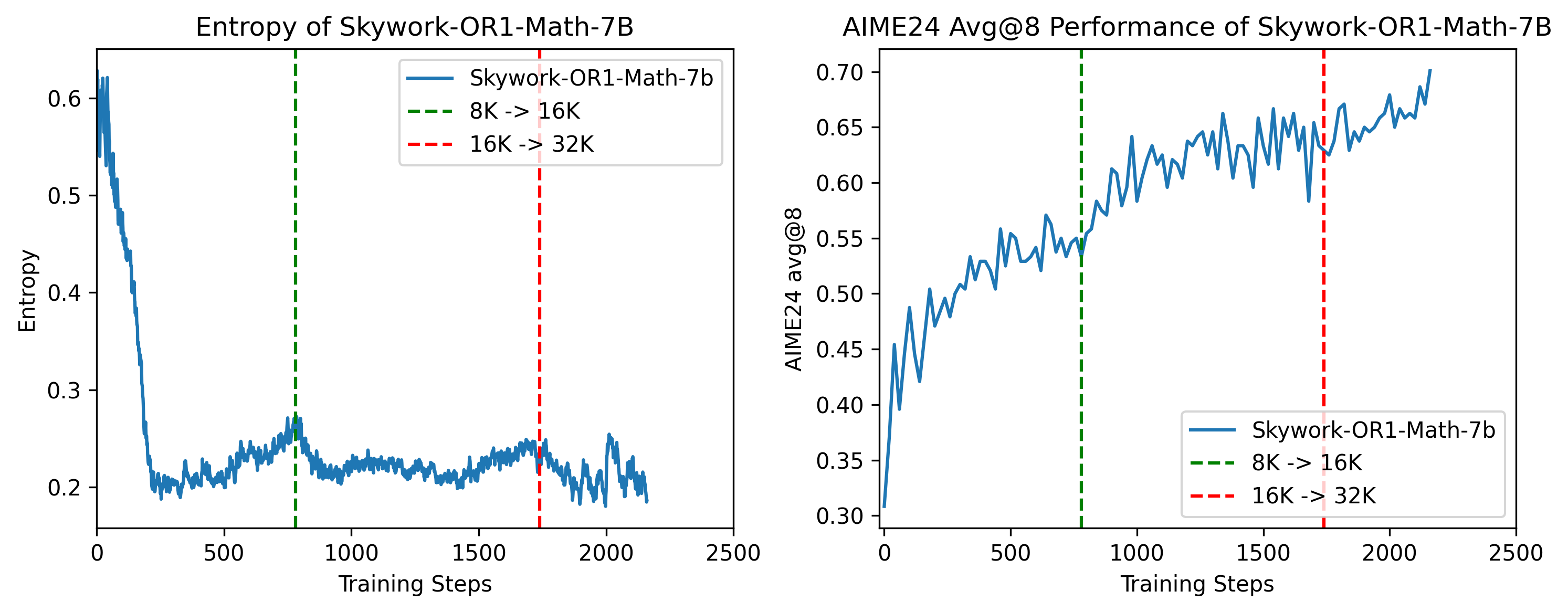}
    \vspace{-1em}
    \caption{Entropy of generated responses (\textbf{left}) and avg@8 performance on AIME24 (\textbf{right})  of Skywork-OR1-Math-7B across all stages. We use adaptive entropy control with \textbf{tgt-ent}=0.2 and $\Delta=0.005$ . Under adaptive entropy control, the entropy of Skywork-OR1-Math-7B is generally lower-bounded by the target entropy 0.2 and the performance on AIME24 has been steadily improving.}
    \label{fig:entropy_of_skywork-R1-math-7b}
\end{figure}

By leveraging adaptive entropy control, we maintain the model's entropy at a reasonable level throughout training and effectively prevent premature collapse. Figure~\ref{fig:entropy_of_skywork-R1-math-7b} illustrates the entropy trajectory of Skywork-OR1-Math-7B across all training stages. In our experiments, we set \textbf{tgt-ent} = 0.2 and $\Delta$ = 0.005. To further validate the effectiveness of adaptive entropy control, we conducted an ablation study detailed in Section~\ref{sec:entropy_control}.
\begin{align}
\alpha_k = c_k \cdot \mathbb{I}\{ e_k \leq \textbf{\text{tgt-ent}} \}, 
c_{k+1} =
\begin{cases}
c_k + \Delta, & \text{if } e_k < \textbf{\text{tgt-ent}} \\
c_k - \Delta, & \text{if } e_k > \textbf{\text{tgt-ent}}
\end{cases}, 
c_0 = 0
\end{align}

% \begin{mybox}{Ablation Experiments 4.2: Adaptive Entropy Control vs Fixed Entropy Coefficient}
% \todo{todo}
% \end{mybox} 

\subsubsection{No KL Loss}
\label{sec:kl}

To investigate the impact of the KL loss, we conducted the following ablation experiments.

\begin{figure}[!htbp]
    \centering
    \subfigure[]{\includegraphics[width=0.47\textwidth]{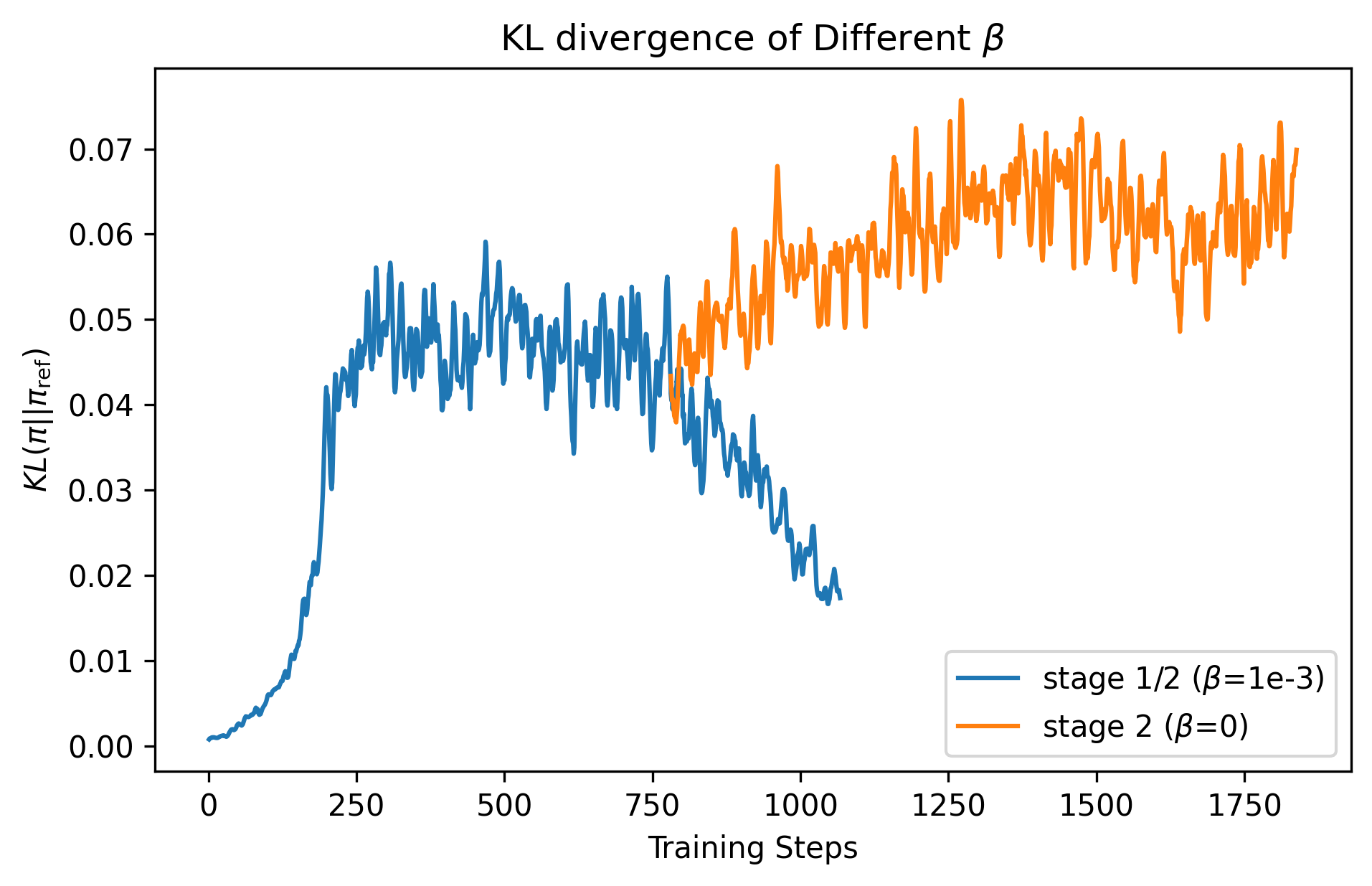}
    \label{fig:kl_loss_comparison}
    }
    \hfill
    \subfigure[]{\includegraphics[width=0.45\textwidth]{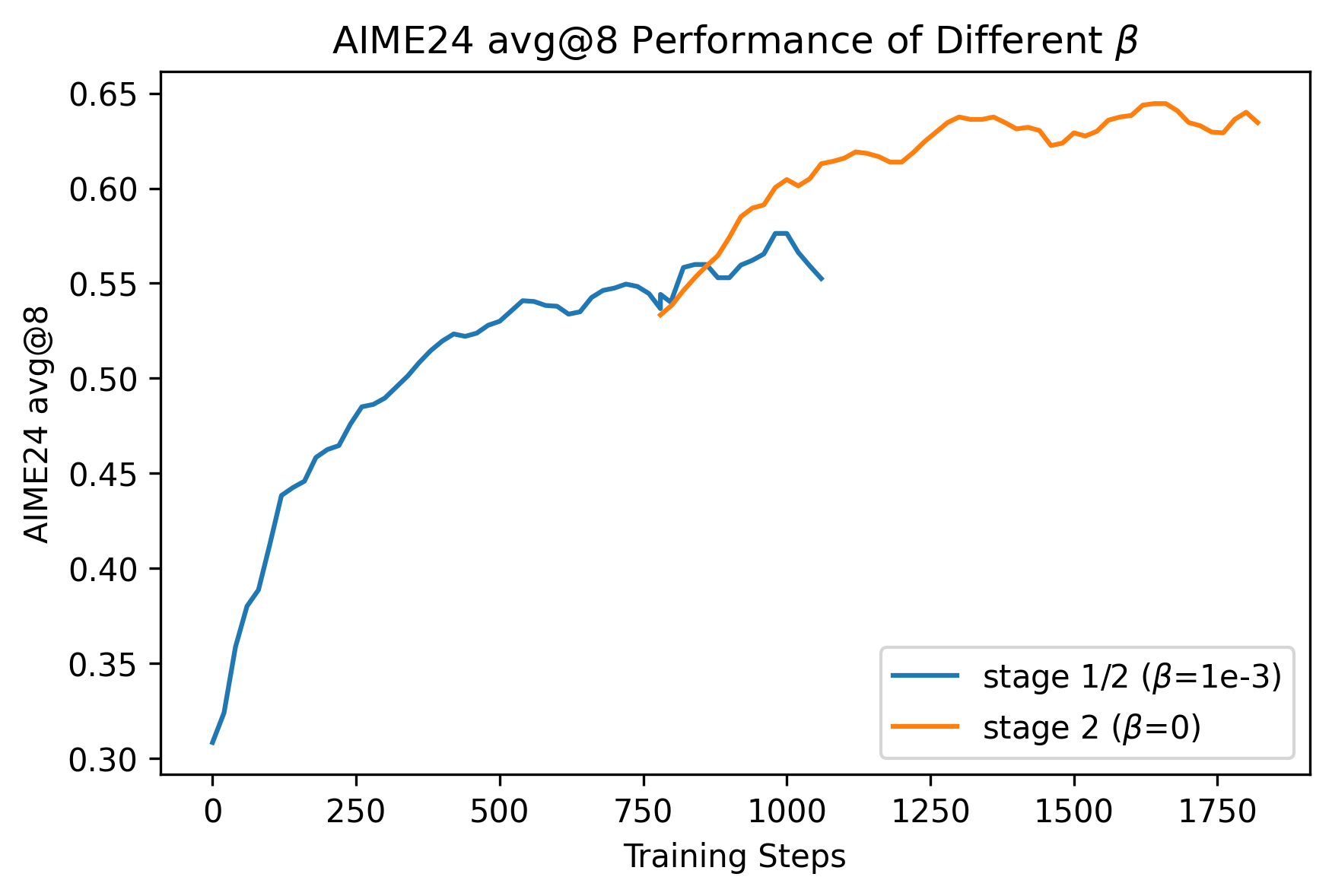}
    \label{fig:kl_loss_aime24_comparison}
    }
    \vspace{-1em}
    \caption{Results of Ablation Experiments \arabic{ablation}. \textbf{Left:} KL divergence between the actor model and the reference model during RL training with different KL loss coefficient $\beta$ in Ablation Experiments 5. Setting $\beta=1\text{e-}3$ pulls the actor model back towards the reference model strongly in stage 2.
    \textbf{Right:} The AIME24 avg@8 performance at temperature 1 during RL training of different $\beta$ in Ablation Experiments 5.
    }
\end{figure}

% \newtcolorbox{mybox}[2][]{colbacktitle=red!10!white, colback=blue!10!white,coltitle=red!70!black, title={#2},fonttitle=\bfseries,#1}
\stepcounter{ablation}
\begin{mybox}{Ablation Experiments \arabic{ablation}: KL Loss vs. No KL Loss}
We consider token-level k3 loss in our ablation and the KL-regularized policy loss we employed is: 
\begin{align*}
\mathcal{L} _{\beta}\left( \theta \right) =\mathcal{L} \left( \theta \right) +\frac{\beta}{T_k}\sum_{i\in \mathcal{T} _k}{\sum_{j=1}^M{\sum_{t=0}^{\left| y_{ij} \right|-1}{\left( \frac{\pi _{\mathrm{ref}}\left( a^{t}_{ij}|s^{t}_{ij} \right)}{\pi _{\theta}\left( a^{t}_{ij}|s^{t}_{ij} \right)}-\log \frac{\pi _{\mathrm{ref}}\left( a^{t}_{ij}|s^{t}_{ij} \right)}{\pi _{\theta}\left( a^{t}_{ij}|s^{t}_{ij} \right)}-1 \right)}}},
\end{align*}
where $\mathcal{L}(\theta)$ is the original policy loss defined in \eqref{policy_loss}, $\beta$ is the KL coefficient. We first run a stage 1 experiment with $\beta$=1\text{e-}3 based on DeepSeek-R1-Distill-Qwen-7B (reference policy). Then in stage 2, we conducted ablations based on the stage 1 checkpoint, comparing $\beta=1\text{e-}3$ with $\beta=0$. The other hyper-parameters are reported in Table \ref{table:shared_parameter_in_ablation_5_1}. The results can be found in Figure \ref{fig:kl_loss_comparison} and Figure \ref{fig:kl_loss_aime24_comparison}.
\end{mybox} 

\begin{table}[!htbp]
\centering
\begin{tabular}{@{\extracolsep{4pt}}ccccccc}
\toprule   
Batch Size & Mini-batch Size & Group Size & Context Length $T$ & Entropy Control \\
\midrule
256 &  128 & 16 & \emph{Stage II} 16K  & target entropy 0.2
\\
\bottomrule
\end{tabular}
\caption{Shared hyperparameters in Ablation Experiments \arabic{ablation} based on stage1 checkpoint} 
\label{table:shared_parameter_in_ablation_5_1}
\end{table}

% \begin{table}[!htbp]
% \label{table:shared_parameter_in_ablation_5_1}
% \centering
% \begin{tabular}{@{\extracolsep{4pt}}ccccccc}
% \toprule   
%  batch size & mini batch size  & group size & $T$& learning rate  & entropy control
% \\
% \midrule
%  256 &  128 & 16 & stage2 16K & 1e-6  & target entropy 0.2
% \\
% \bottomrule
% \end{tabular}
% \caption{Shared hyperparameters in Ablation Experiments 5 based on stage1 checkpoint} 
% \end{table}

We observe that, in Stage 2, the KL loss strongly pulls the actor model’s policy back toward the reference model, causing the KL divergence to rapidly decrease toward zero (see Figure~\ref{fig:kl_loss_comparison}). As a result, performance on AIME24 fails to improve significantly once the actor’s policy becomes too similar to the reference policy (see Figure~\ref{fig:kl_loss_aime24_comparison}). Based on this observation, we set $\beta = 0$ for all training stages of our released models.

% \begin{figure}[!htbp]
%     \centering
%     \includegraphics[width=0.7\linewidth]{figures/kl_loss_comparison.png}
%     \caption{The KL divergence between the actor model and reference model during the RL training with different KL loss coefficient $\beta$ in Ablation Experiments 5. Setting $\beta$=1e-3 pulls the actor model back to the reference model hardly at stage 2.}
%     \label{fig:kl_loss_comparison}
% \end{figure}

% \begin{figure}[!htbp]
%     \centering
%     \includegraphics[width=0.7\linewidth]{figures/kl_loss_aime24_comparison.png}
%     \caption{The AIME24 avg@8 performance at temperature 1 during the RL training of different $\beta$ in Ablation Experiments 5}
%     \label{fig:kl_loss_aime24_comparison}
% \end{figure}

\newpage
\section{Empirical Studies on Mitigating Policy Entropy Collapse}
\label{sec:entropy}

\begin{figure}[!htbp]
    \centering
    \includegraphics[width=0.9\linewidth]{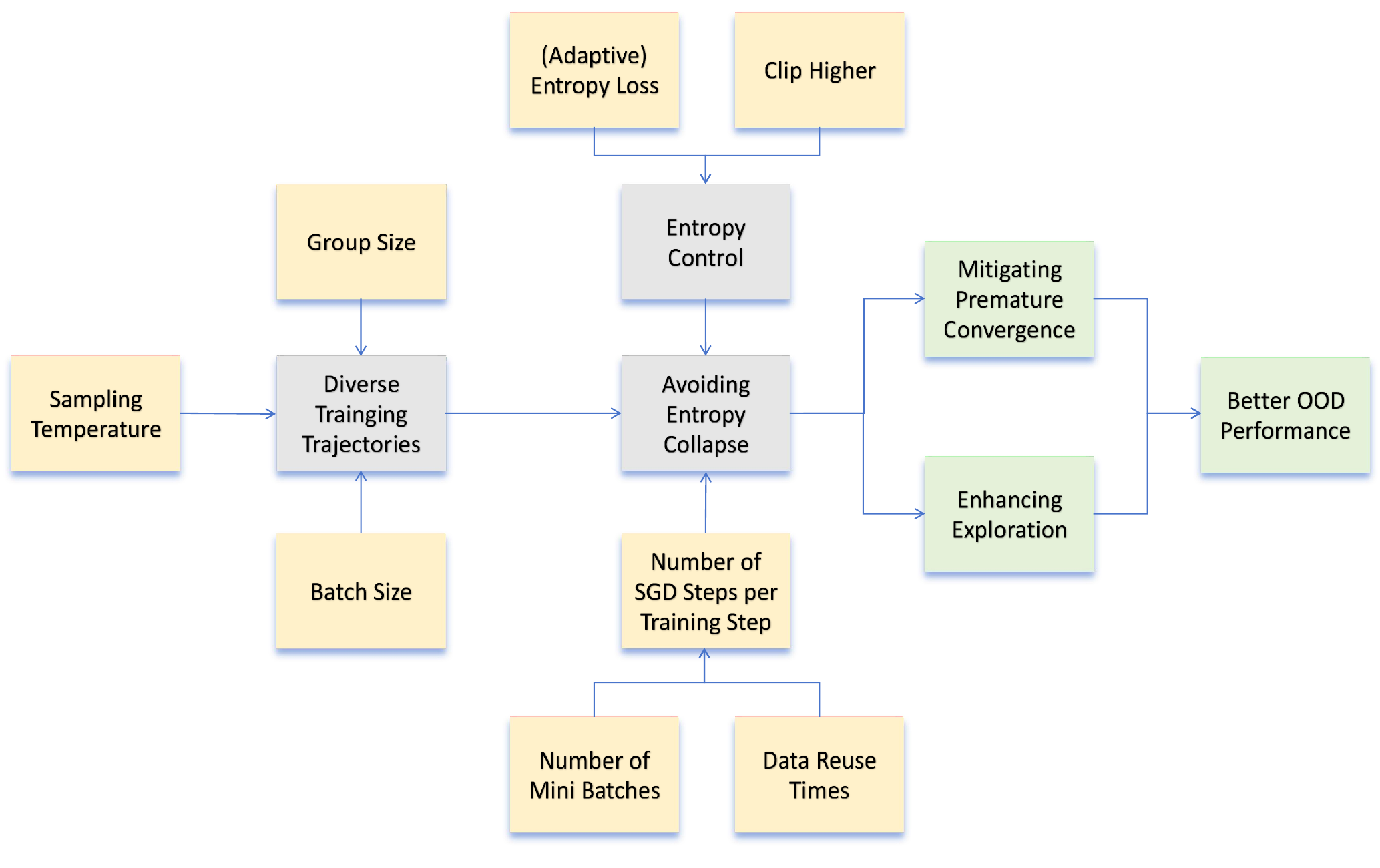}
    \caption{Overview of our empirical studies on mitigating policy entropy collapse. \textbf{Gray‌ and green blocks:} The potential benefits and possible approaches to enhance the model's exploration capability and mitigate entropy collapse. \textbf{Yellow blocks:} The experimental variables in our empirical studies on keeping the model's exploration capability and maintaining high plasticity.}
    \label{fig:entropy_motivation}

\end{figure}

Exploration and exploitation represent one of the most fundamental dilemmas in RL training \cite{sutton2018reinforcement}, particularly in on-policy algorithms. In brief, \textbf{achieving better performance requires sufficient exploration.} However, if the agent's policy prematurely converges to a specific solution, that policy may be suboptimal, and such convergence hinders the exploration of diverse trajectories. 
An important metric for monitoring the convergence of RL algorithms is policy entropy. In general, when a model's policy entropy converges to a very small value (e.g., near zero), the policy stabilizes. At this point, the model's generation behavior becomes resistant to updates from training data, leading to reduced learning efficiency and diminished output diversity. To expose the model to more effective training signals and improve its out-of-distribution (OOD) performance, it is therefore critical to prevent premature entropy collapse in practice.
This section investigates which hyperparameters and components of the policy update process help prevent entropy collapse and, in turn, improve OOD generalization. The overall framework of our empirical study on alleviating policy entropy collapse is illustrated in Figure~\ref{fig:entropy_motivation}.
Initially, we hypothesize that the following two sources may influence the model's entropy and convergence behavior:

\begin{itemize}
    \item \textbf{Rollout diversity.} If the rollout data contain a greater diversity of correct responses, this prevents the model from overfitting to a single correct trajectory. We examine how sampling-related hyperparameters -- such as sampling temperature, rollout batch size, and group size -- affect the model’s policy entropy during RL training.
    
    \item \textbf{Policy update.} We also investigate how different components of the policy update influence entropy. In this section, we focus primarily on the number of stochastic gradient descent (SGD) steps per training step and the use of additional entropy control methods (e.g., entropy loss).
\end{itemize}

After conducting exhaustive ablation experiments, we present our main results below.

\begin{tcolorbox}[colback=SeaGreen!10!CornflowerBlue!10,colframe=RoyalPurple!55!Aquamarine!100!,title=\textbf{Empirical Results of Our Entropy Collapse Study}]
\begin{enumerate}[leftmargin=5pt,
rightmargin=5pt,itemsep=5pt]
    
    \item Faster entropy collapse generally leads to worse test performance. In Section~\ref{sec:entropy_motivation} and Section~\ref{sec:entropy_control}, we show that appropriate entropy control, which prevents premature policy convergence, can yield improved test performance. 

    \item Increasing rollout diversity by enlarging the batch size and group size has only a minor effect on entropy dynamics, whereas using a higher sampling temperature significantly impacts initial entropy. See Section~\ref{sec:impact_of_rollout_diversity} for details.
    
    \item Increasing the number of SGD steps per training step -- whether by using more mini-batches or increasing data reuse -- significantly accelerates entropy collapse and generally results in degraded test performance due to the introduction of off-policy data. See Section~\ref{sec:sgd_steps} for more information.

    \item Our ablation experiments in Section~\ref{sec:entropy_control} show that the entropy loss is highly sensitive to both the training data and the loss coefficient. By either adaptively adjusting the entropy loss coefficient or appropriately applying the clip-higher trick \cite{yu2025dapo}, entropy dynamics can be stabilized and lower-bounded, leading to improved test performance.
\end{enumerate}
\end{tcolorbox}

\subsection{Ablation Setup} 
\label{sec:ablation_setup}
All ablation experiments presented in Section \ref{sec:entropy} are conducted using the training pipeline described in Section \ref{sec:policy_update}. We start from the following baseline experiment based on DeepSeek-R1-Distill-Qwen-7B with its hyperparameters reported in Table \ref{table:baseline_experiments}, the key symbols used are defined as follows:
\begin{itemize}
    \item $D_R$ is the rollout batch size (the number of prompts used to generate responses in one training step). 
    \item $D_T$ is the mini-batch size (the number of prompts corresponding to the responses used per policy update step).
    \item $N_\text{reuse}$ is the number of times the rollout buffer is traversed.
    \item $gs$ is the group size (the number of responses generated for each prompt).
    \item $T$ is the context length. 
    \item $\tau$ is the sampling temperature.
\end{itemize}
\begin{table}[!htbp]
\centering
\begin{tabular}{@{\extracolsep{4pt}}cccccccccccc}
\toprule   
$D_R$ & $D_T$ & $N_\text{reuse}$ & $gs$ & $T$ &  $\tau$& Learning Rate & Entropy Control & KL loss
\\
\midrule
 64 & 64 & 1 & 16 &  16K & 1.0 & 1e-6  & No & No
\\
\bottomrule
\end{tabular}
\caption{Hyperparameters of our baseline experiment in the ablation study presented in Section \ref{sec:entropy}.} 
\label{table:baseline_experiments}
\end{table}
Unless otherwise specified, the default training configurations for all ablation experiments in this section are aligned with those of the baseline experiment presented above. We use AIME24, AIME25, and LiveCodeBench \cite{jain2024livecodebench} (2024.08–2025.02) as evaluation sets. The test performance reported in our ablation study is computed as the empirical mean of avg@8 performance on AIME24/25 and pass@1 performance on LiveCodeBench. Notably, the baseline experiment achieves 69.2\% avg@8 on AIME24, 53.3\% avg@8 on AIME25, and 50.5\% pass@1 on LiveCodeBench after 2,700 training steps using 32 H800 GPUs. These results, which closely approximate the performance of our final Skywork-OR1-7B release, establish a strong baseline for analyzing key factors that affect test performance and contribute to entropy collapse.

\subsection{Premature Entropy Collapse Generally Manifests as Worse Performance}
\label{sec:entropy_motivation}
\begin{figure}[t]
    \centering
    \includegraphics[width=0.85\linewidth]{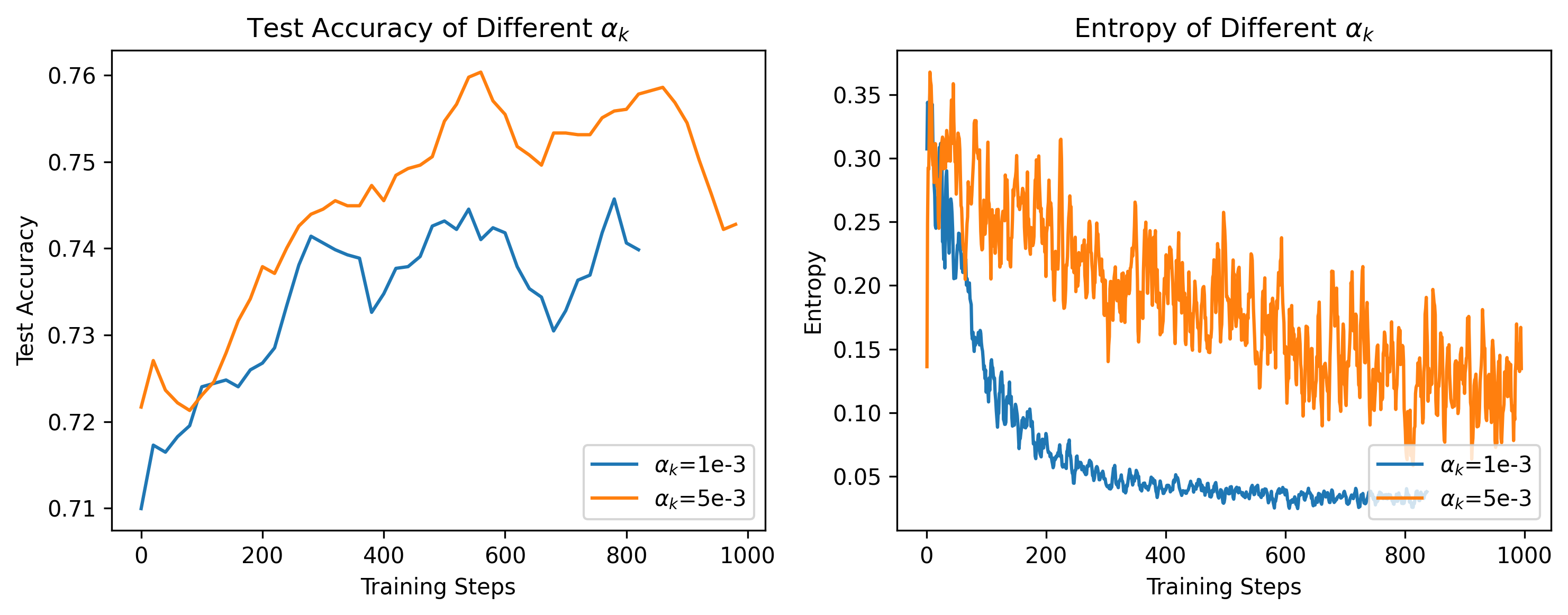}
    \vspace{-1em}
    \caption{Preliminary experiments on mitigating entropy collapse by introducing entropy loss. We tested two different coefficients $\alpha_k = $ 1e-3 and 5e-3, and found that the entropy loss with the higher coefficient $\alpha_k$, i.e., 5e-3, more effectively prevents entropy collapse and achieves higher test performance. \textbf{Left}: Accuracy curves on test benchmarks during  RL training. \textbf{Right}: Entropy of generated responses during  RL training.}
    \label{fig:ent_motivation}
\end{figure}

As previously noted, entropy dynamics during RL training reflect the degree of policy convergence. When the actor converges to a specific policy and enters a low-entropy state, both learning efficiency and rollout diversity tend to decline. In our preliminary experiments, we observed that the entropy of the actor model often decreased rapidly during training. To mitigate premature entropy collapse, we introduced an entropy loss term, hypothesizing that it would allow the actor to converge toward a better policy. Our results confirmed this hypothesis: test performance improved with the addition of entropy loss.
Figure~\ref{fig:ent_motivation} presents the accuracy curves on test benchmarks and the entropy of generated responses from two preliminary experiments using different values of the entropy loss coefficient $\alpha_k$ (1e-3 vs. 5e-3). The results show that using a higher coefficient (i.e., 5e-3) more effectively prevents entropy collapse and leads to better generalization performance. Furthermore, our ablation experiments in Section~\ref{sec:sgd_steps} reinforce this finding, showing that RL training accompanied by premature entropy collapse generally results in worse test performance.
These observations motivate our integration of entropy control mechanisms into the training pipeline, as well as our systematic investigation into how hyperparameters and other RL components influence entropy dynamics.

\subsection{The Impact of Rollout-Diversity-Related Hyperparameters}
\label{sec:impact_of_rollout_diversity}
\begin{figure}[!ht]
    \centering
    \includegraphics[width=0.95\linewidth]{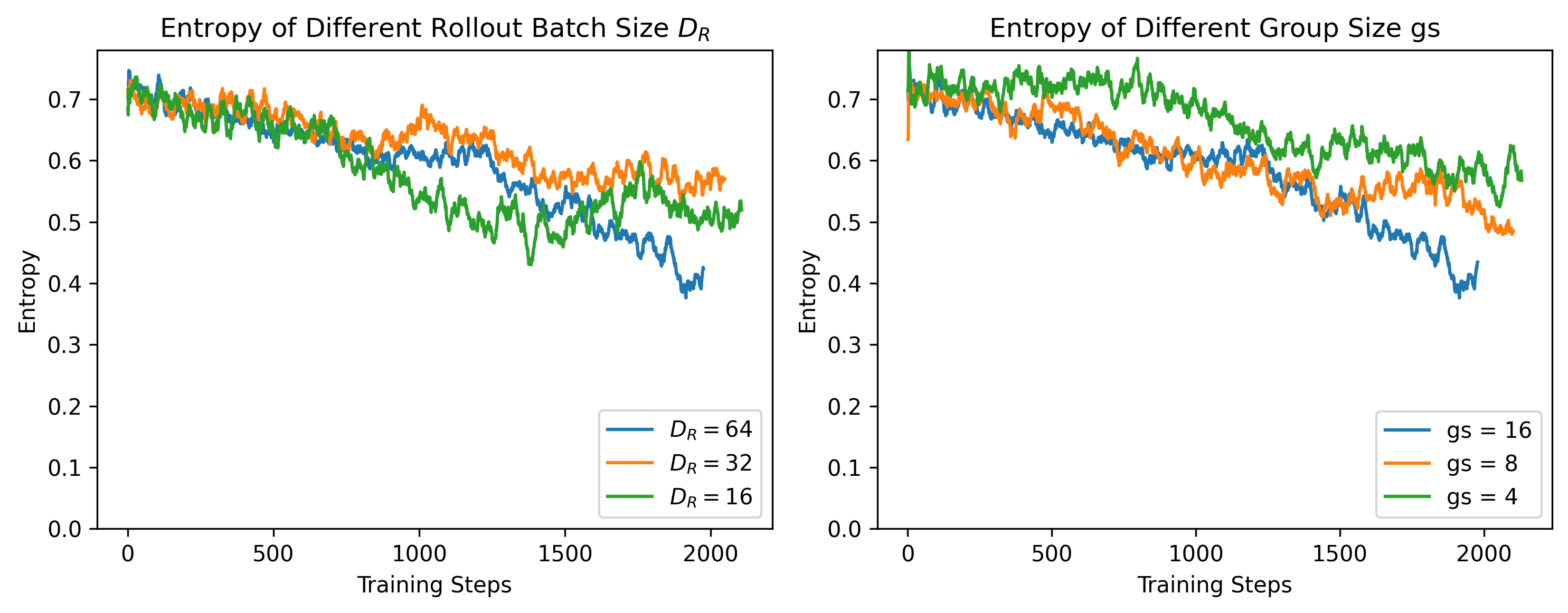}
    \vspace{-1em}
    \caption{Entropy of generated responses during on-policy updates with different rollout batch sizes $D_R$ (\textbf{left}) and group size $\text{gs}$ (\textbf{right}). All the experiments exhibit similar entropy dynamics.}
    \label{fig:entropy_of_bsz_and_gs}
\end{figure}

We investigated how the rollout batch size $D_R$, group size $gs$, and sampling temperature $\tau$ influence entropy dynamics. Note that increasing the rollout batch size $D_R$ and group size $gs$ during the rollout stage results in a larger rollout budget, which typically requires greater computational resources to accelerate training. Therefore, we provide a detailed discussion of the impact of $D_R$ and $gs$ in Section~\ref{sec:training_time_scaling}, which focuses on training-time computational resource allocation for improved test performance. Here, we present only the experimental results related to policy entropy.
Specifically, we conducted ablation experiments using rollout batch sizes $D_R = 16, 32, 64$ and group sizes $gs = 4, 8, 16$, based on the baseline experiment described in Section~\ref{sec:ablation_setup} and analyzed in Section~\ref{sec:training_time_scaling}. Our results (Figure~\ref{fig:entropy_of_bsz_and_gs}) indicate no significant differences in entropy dynamics across these on-policy configurations. Notably, none of these experiments exhibited entropy collapse.
Regarding the sampling temperature $\tau$, we found that using a properly chosen but relatively high temperature led to lower test accuracy during the initial training steps, but ultimately resulted in greater performance improvements. For further details, please refer to Section~\ref{sec:temperature}.

\subsection{The Impact of Off-policy Update by Increasing $N_\text{SGD}$}
\label{sec:sgd_steps}
\begin{figure}[!htbp]
    \centering
    \includegraphics[width=1\linewidth]{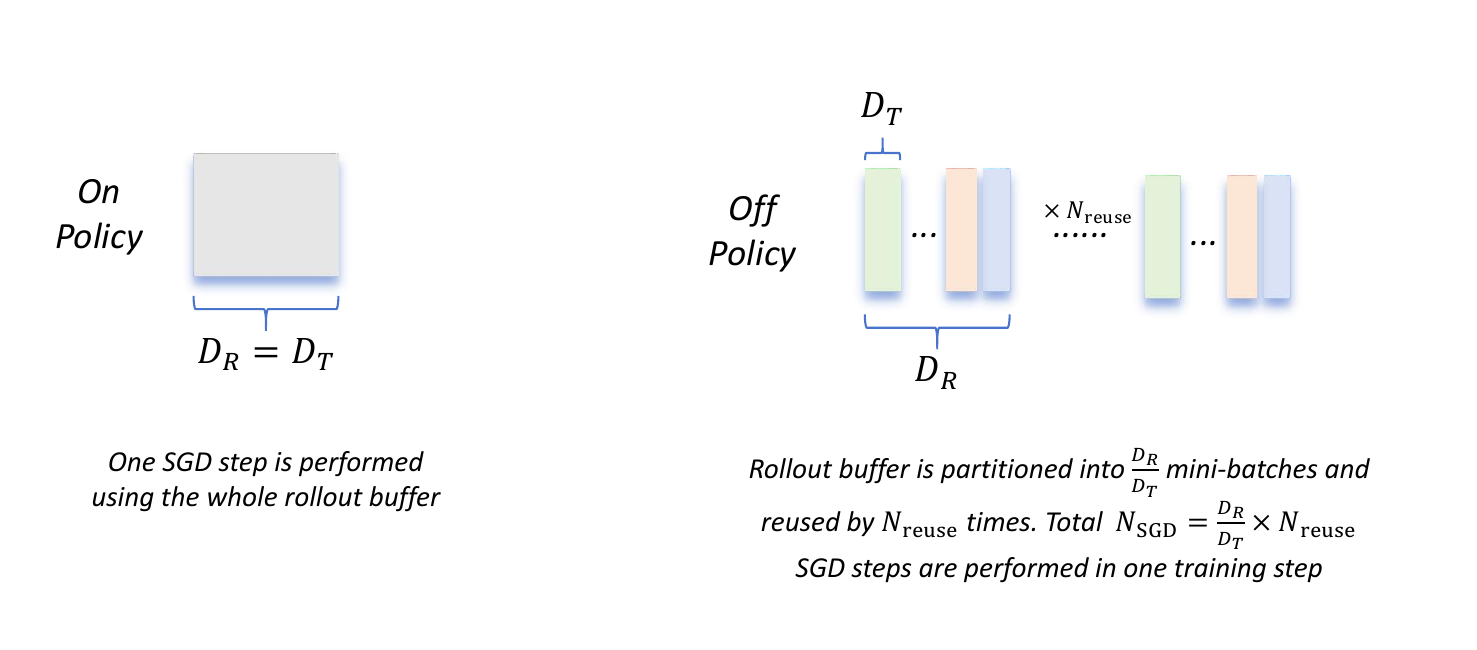}
    \caption{Illustration of on-policy vs. off-policy update in PPO-style policy loss. On-policy update applies a single SGD step to the entire rollout batch, whereas off-policy update implements multiple SGD steps through rollout batch decomposition and reuse. The rollout batch is partitioned into $\frac{D_R}{D_T}$ mini-batches, with each mini-batch undergoing an independent SGD step. Then, one can iterate over the rollout batch $N_\text{reuse}$ times. Thus, the total number of SGD steps performed on one rollout batch is $\frac{D_R}{D_T}\times N_\text{reuse}$.}
    \label{fig:number_of_sgds}
\end{figure}
Note that the policy loss \eqref{policy_loss} in MAGIC is PPO-style, which naturally allows for performing multiple SGD steps through rollout batch decomposition and reuse (as illustrated in Figure \ref{fig:number_of_sgds}). Recalling the definitions of $D_R$,$D_T$ and $N_\text{reuse}$ from Section \ref{sec:ablation_setup}, it is clear that the number of SGD steps performed in one training step, i.e. $N_\text{SGD}$, satisfies
\begin{align}
    N_\text{SGD}=\frac{D_R}{D_T}\cdot N_{\text{reuse}}.
\label{eq:sgd_times_eq}
\end{align}
When $D_R=D_T$ and $N_{\text{reuse}}=1$, the policy update is purely on-policy since $N_\text{SGD}=1$. In contrast, when $D_T<D_R$ or $N_{\text{reuse}}\ge 2$, $N_\text{SGD} \ge 2$ and the off-policy data is introduced into the policy update. In this section, we investigate how $N_\text{SGD}$ affects the entropy dynamics and the test performance improvement. 

\paragraph{More SGD Steps, Faster Convergence with Worse Test Performance.} We conducted the following ablation experiments on different $N_\text{SGD}$ values by decreasing $D_{T}$ or increasing $N_\text{reuse}$ given fixed $D_R$. 

\begin{figure}[!htbp]
    \centering
    \includegraphics[width=0.95\linewidth]{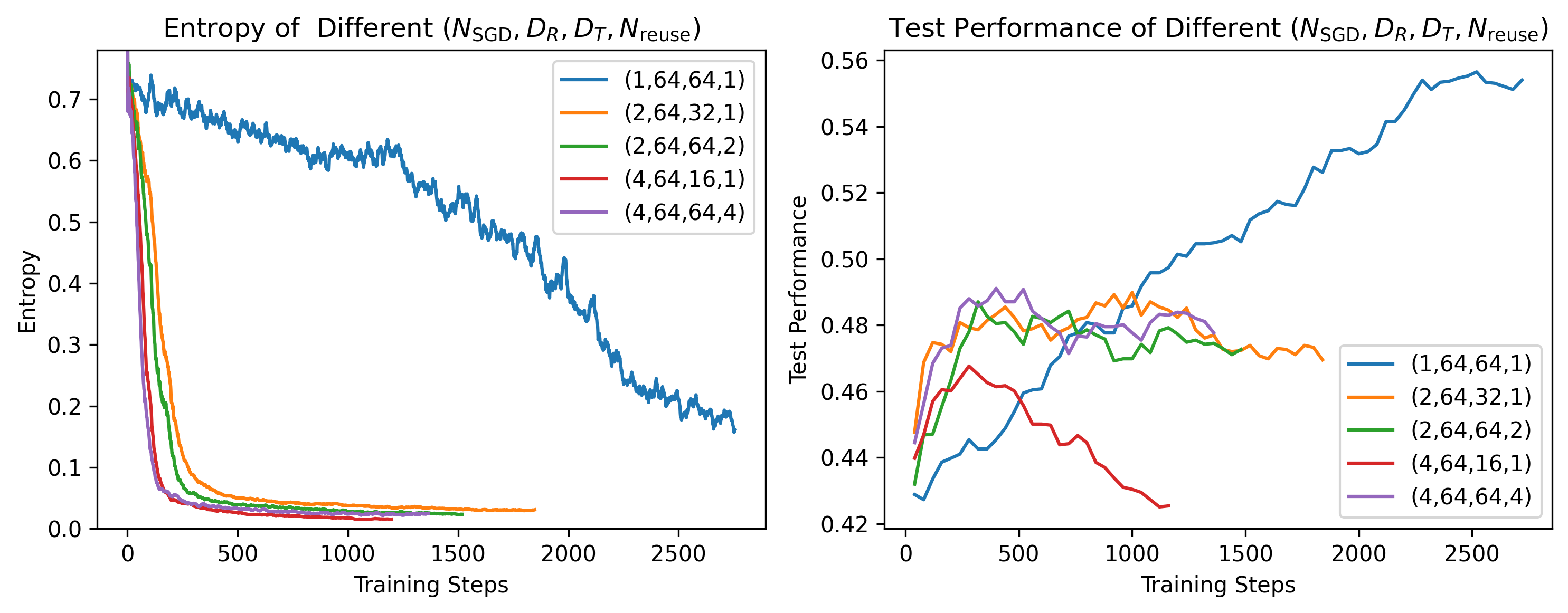}
    \vspace{-1em}
    \caption{Results of Ablation Experiments \arabic{ablation}. Off-policy training with increased $N_\text{SGD}$ by either decreasing $D_T$ or increasing $N_\text{reuse}$ accelerates entropy collapse and exhibits worse test performance. \textbf{Left}: Entropy of generated responses during RL training. \textbf{Right}: Test performance during RL training.}
    \label{fig:nsgd_compare_1}
\end{figure}

\stepcounter{ablation}
\begin{mybox}{Ablation Experiments \arabic{ablation}: The Impact of Different Numbers of SGD Steps $N_\text{SGD}$}
\label{ablation_sgd}
Consider the quadruple $(N_\text{SGD},D_R,D_T,N_\text{reuse})$. 
We started from the baseline experiment (1,64,64,1) presented in Section \ref{sec:ablation_setup} and adjusted either $D_T$ or $N_\text{reuse}$ to increase $N_\text{SGD}$. The experiments are listed below:
\begin{enumerate}[leftmargin=15pt,rightmargin=5pt,itemsep=5pt]
\vspace{5pt}
\item $N_\text{SGD}=1$: The baseline experiment with the quadruple (1,64,64,1).
\item $N_\text{SGD}=2$: We ran two experiments with the quadruples (2,64,32,1) and (2,64,64,2).
\item $N_\text{SGD}=4$: We ran two experiments with the quadruples (4,64,16,1) and (4,64,64,4).
\vspace{5pt}
\end{enumerate}
The experimental results can be found in Figure \ref{fig:nsgd_compare_1}.
\end{mybox} 

As shown in Figure~\ref{fig:nsgd_compare_1}, experiments with $N_\text{SGD} \in \{2, 4\}$ exhibit faster policy convergence, with entropy decaying to very small values within a few training steps. As a result, test performance fails to improve consistently once the model enters a low-entropy state. In contrast, using an on-policy update with the configuration $(1,64,64,1)$ significantly alleviates this issue, leading to a gradual decline in entropy and a steady, albeit slower, improvement in test performance. Ultimately, the on-policy update with configuration $(1,64,64,1)$ achieves superior test performance when the number of training steps is sufficiently large.

\paragraph{Off-Policy Data Harms Test Performance.} We now investigate which factor in off-policy updates is more likely to contribute to degraded test performance. We identify the following two potential contributors that may influence the gradient direction in each SGD step: (1) the mini-batch size $D_T$, and (2) the use of off-policy data.
In the data reuse experiments with $N_\text{reuse} \in \{2, 4\}$, since $D_T$ is held constant and matches the value used in the on-policy setting, we attribute the degraded test performance to the use of off-policy data introduced through rollout batch reuse. In experiments that involve more mini-batches (i.e., $D_T \in \{16, 32\}$), the performance drop compared to the on-policy update may be due to both the smaller mini-batch size -- leading to greater gradient variance -- and the presence of off-policy data.
To better understand which factor contributes more significantly, we conducted the following ablation experiments.

\begin{figure}[!htbp]
    \centering
    \includegraphics[width=0.95\linewidth]{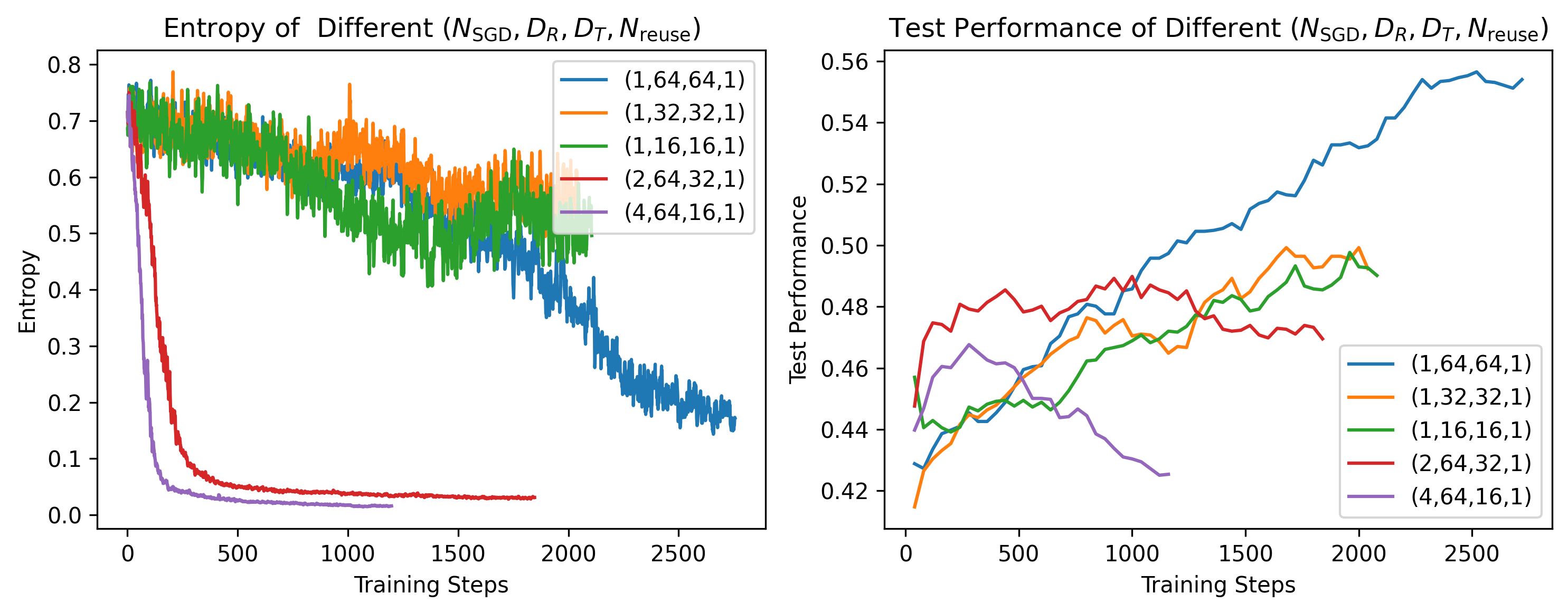}
    \vspace{-1em}
    \caption{Results of Ablation Experiments \arabic{ablation}. On-policy experiments, i.e. $N_\text{SGD}=1$, do not exhibit premature entropy collapse and finally outperform the off-policy counterparts with the same $D_T$ when training step is sufficiently large. \textbf{Left}: Entropy of generated responses during RL training. \textbf{Right}: Test performance at temperature 1 during RL training.}
    \label{fig:on_policy_vs_off_policy}
\end{figure}

\stepcounter{ablation}
\begin{mybox}{Ablation Experiments \arabic{ablation}: On-policy vs. Off-policy with the Same SGD Data Size $D_T$}
Consider the quadruple $(N_\text{SGD},D_R,D_T,N_\text{reuse})$.
\begin{enumerate}[leftmargin=15pt,rightmargin=5pt,itemsep=5pt]
\vspace{5pt}
\item Off-policy update: We considered two off-policy experiments in Ablation Experiments 6 with the quadruples (2,64,32,1) and (4,64,16,1), which have smaller $D_T$ compared to the baseline (1,64,64,1).

\item On-policy update: We ran two experiments, configured with the quadruples (1,32,32,1) and (1,16,16,1) respectively as the on-policy counterparts to the off-policy update. These were based on the baseline configuration from Section \ref{sec:ablation_setup}.
\end{enumerate}
\vspace{5pt}
The experimental results are reported in Figure \ref{fig:on_policy_vs_off_policy}.
\end{mybox} 

The experimental results shown in Figure~\ref{fig:on_policy_vs_off_policy} indicate that the on-policy update with a smaller $D_T$ -- relative to the baseline experiment -- still yields steady improvements in test performance, and premature entropy collapse does not occur. Ultimately, the on-policy update outperforms the off-policy update with the same $D_T$ when the number of training steps is sufficiently large. Based on these observations, we hypothesize that the degraded test performance in the off-policy update is primarily caused by the introduction of off-policy data in each SGD step.

\begin{figure}[h]
    \centering
    \vspace{-1em}
    \includegraphics[width=0.65\linewidth]{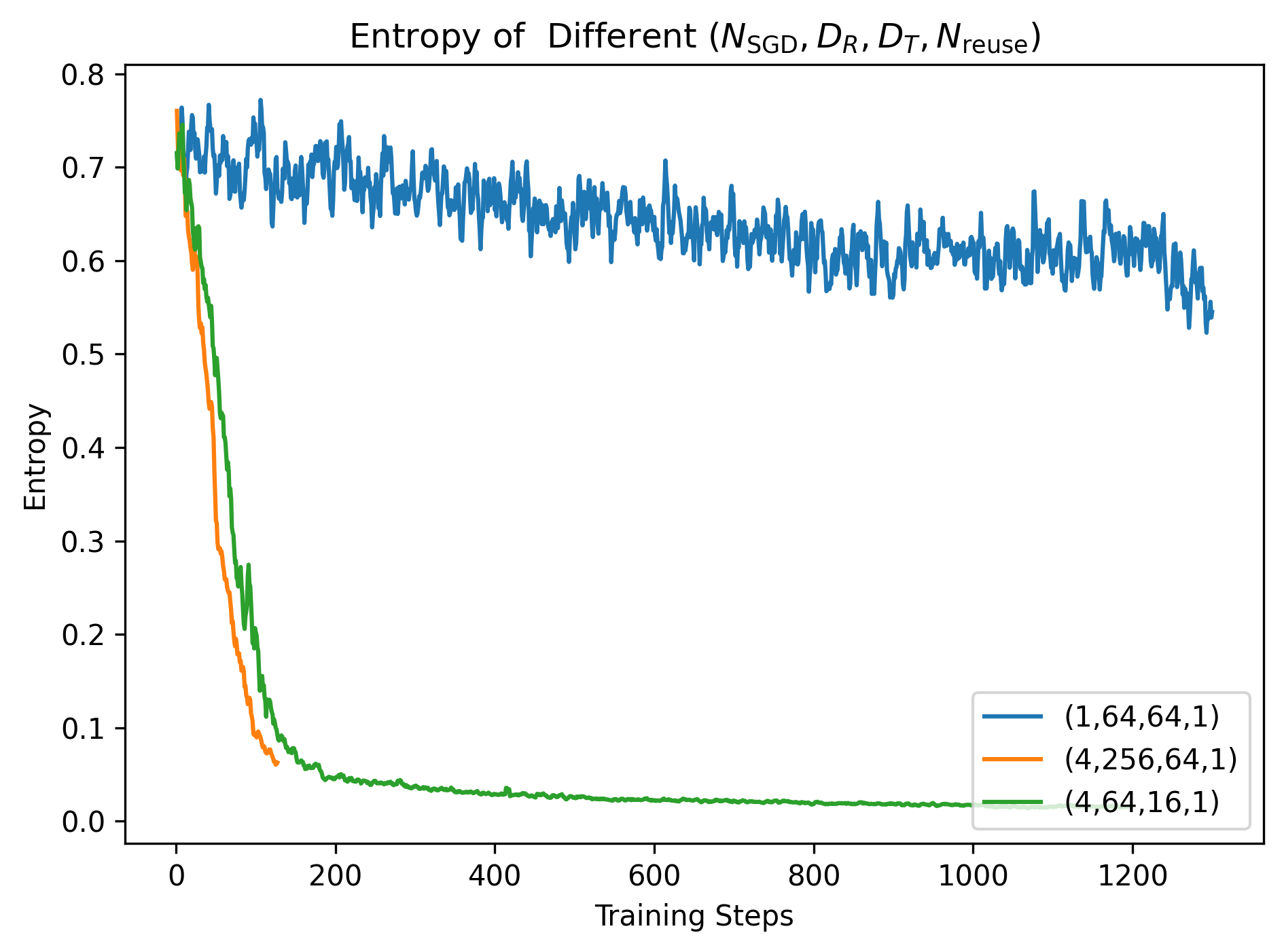}
    \caption{Keeping $\frac{D_R}{D_T}=4$ and $N_\text{reuse}=1$, off-policy training with a larger $D_R$, i.e., $D_R=256$, does not prevent the premature entropy collapse. Both off-policy experiments, i.e. $N_\text{SGD}=4$, exhibit faster entropy convergence compared with the on-policy experiment with $N_\text{SGD}=1$.} 
    \label{fig:nsgd_compare_3}
\end{figure}

\textbf{Can a Large $D_R$ in Off-Policy Updates Prevent Premature Entropy Collapse?} Consider the off-policy experiment in Ablation Experiments 6 with the quadruple $(N_\text{SGD},D_R,D_T,N_\text{reuse})= (4,64,16,1)$. We attempted to increase the rollout batch size $D_R$ from 64 to 256 while keeping $N_\text{SGD}=4$ fixed (i.e., resulting in the configuration $(N_\text{SGD},D_R,D_T,N_\text{reuse})=(4,256,64,1)$),
with the expectation that this would introduce more diverse samples and prevent convergence on single trajectory. However, our results in Figure \ref{fig:nsgd_compare_3} indicates that even with a larger $D_R$, premature entropy collapse not only still occurs but may even do so more rapidly.

\subsection{Preventing Premature Entropy Collapse}
\label{sec:entropy_control}
As previously discussed, premature entropy collapse is often associated with degraded test performance. It is therefore reasonable to expect that proper entropy control can lead to improved outcomes. As shown earlier, increasing $N_\text{SGD}$ and introducing off-policy data accelerate entropy convergence. However, there are an increasing number of scenarios where the use of off-policy data is unavoidable -- for example, in asynchronous training frameworks. Thus, it is also important to study entropy control mechanisms under off-policy settings.
We begin by examining entropy regularization, a straightforward approach that attempts to prevent entropy collapse by directly adding an entropy loss term. Our preliminary experiments, presented in Section~\ref{sec:entropy_motivation}, show that applying entropy regularization with an appropriately chosen coefficient can mitigate entropy collapse and improve test performance. However, we later observed that the effectiveness of entropy regularization is highly sensitive to both the choice of coefficient and the characteristics of the training data, making it difficult to select an optimal coefficient in advance. This motivates a dynamic adjustment of the entropy loss coefficient.
In addition, we consider the clip-higher trick proposed in \cite{yu2025dapo} as another means of entropy control. In the following, we present our detailed findings.

\begin{figure}[!h]
    \centering
    \includegraphics[width=0.85\linewidth]{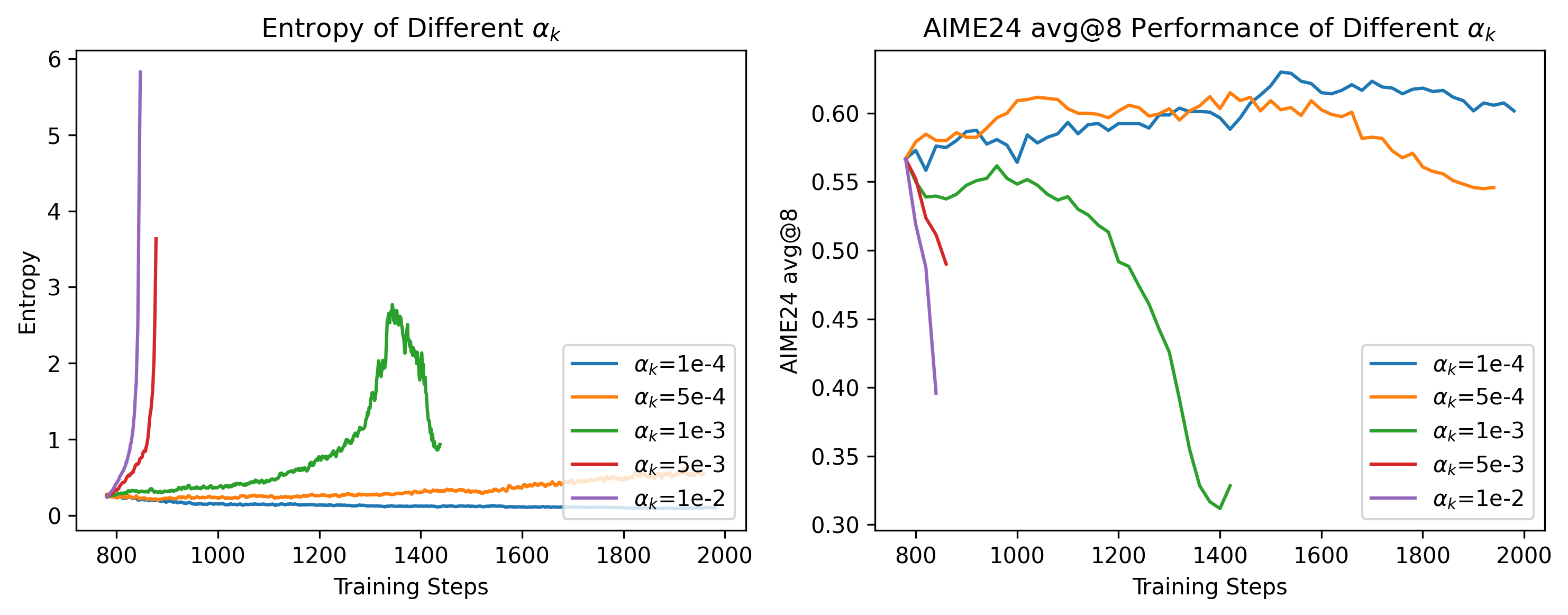}
    \vspace{-1em}
    \caption{The results of Ablation Experiments \arabic{ablation}. \textbf{Left:} The entropy of generated responses during RL training. \textbf{Right:} The AIME24 avg@8 performance at temperature 1 during RL training.}
    \label{fig:different_ent_coeff_comparison}
\end{figure}

\paragraph{Entropy Loss Is Sensitive to the Coefficient $\alpha_k$.} To demonstrate the sensitivity of entropy loss to the choice of $\alpha_k$, we conduct the following ablation study. 
\stepcounter{ablation}
\begin{mybox}{Ablation Experiments \arabic{ablation}: Entropy Loss with Different Coefficients $\alpha_k$}
We conducted ablation studies on a wide range of constant coefficients $\alpha_k$ based on Skywork-OR1-Math-7B-stage1 (\textbf{not} the baseline experiment in Section \ref{sec:ablation_setup} ). We select $\alpha_k$=1e-4, 5e-4, 1e-3, 5e-3, 1e-2. The other hyperparameters are reported in Table \ref{table:shared_parameter_in_ablation_4_1}. The results are presented in Figure \ref{fig:different_ent_coeff_comparison}. 
\end{mybox} 

\begin{table}[!htbp]
\centering
\begin{tabular}{@{\extracolsep{4pt}}ccccccc}
\toprule   
 Batch Size & Mini-batch Size & Group Size & $T$ &  Temperature $\tau$ & KL Loss
\\

\midrule
 64 &  32 & 16 & \emph{Stage II} 16K & 1.0  & No
\\
\bottomrule
\end{tabular}
\caption{Shared Hyperparameters in Ablation Experiments \arabic{ablation} Based on Skywork-OR1-Math-7B-stage1} 
\label{table:shared_parameter_in_ablation_4_1}
\end{table}

From the results in Figure \ref{fig:different_ent_coeff_comparison}, we find that:
\begin{itemize}
    \item For $\alpha_k$ = 5e-4, 1e-3, 5e-3, and 1e-2, the entropy eventually rises sharply, leading to model collapse. The larger the $\alpha_k$, the more rapidly the entropy increases.
    \item For $\alpha_k$ = 1e-4, while entropy does not exhibit a continuous rise, it still collapses, persistently decreasing toward zero.
\end{itemize}

\begin{figure}[t]
    \centering
    \includegraphics[width=0.55\linewidth]{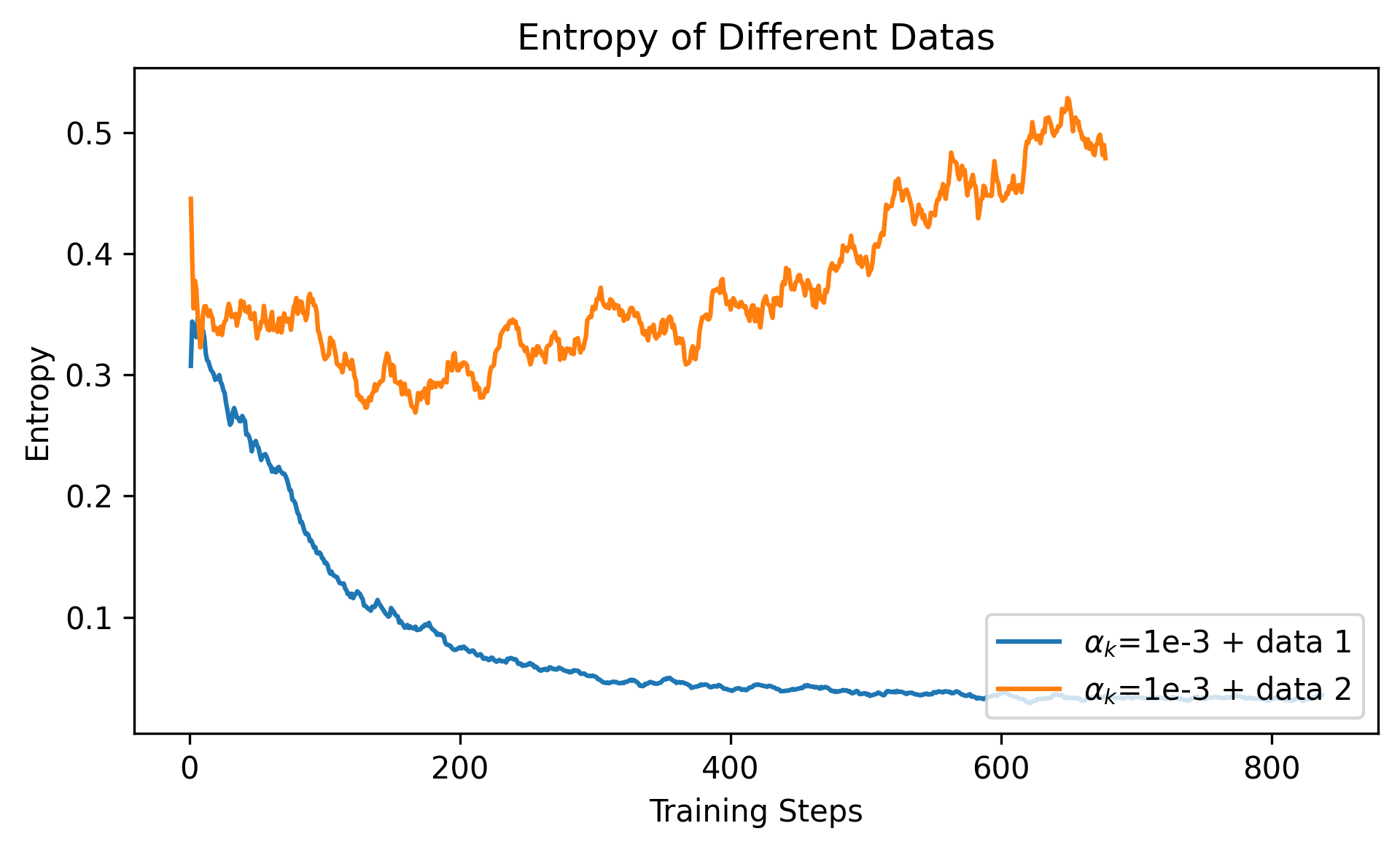}
    \vspace{-1em}
    \caption{Preliminary experiments investigating how training data affects the entropy during RL training. Both experiments used the same hyperparameter configurations with $\alpha_k$=1e-3 but differed in the training data. Both datasets are in math domain. simply switching the dataset resulted in dramatically different entropy evolution patterns }
    \label{fig:ent_sensitive_to_data}
\end{figure}

\paragraph{Entropy Loss Is Sensitive to Training Data.} From our two preliminary experiments, we observe that the entropy loss is highly sensitive to variations in training data. We conducted two experiments under identical configurations, both using an entropy loss coefficient of 1e-3. The only difference between the two setups was the training dataset used (both datasets belong to the math domain). The results, shown in Figure~\ref{fig:ent_sensitive_to_data}, reveal a striking difference in entropy dynamics: while the original dataset exhibited a steady decline in entropy throughout training, the new dataset resulted in a consistent upward trend in entropy. This finding highlights the data-dependent nature of tuning the entropy loss coefficient.

\paragraph{Adjusting the Coefficient of Entropy Loss Adaptively.} Based on our findings regarding the sensitivity of entropy loss, we propose a method called \textbf{adaptive entropy control} (see Section~\ref{sec:adaptive_entropy} for details), which dynamically adjusts the entropy loss coefficient during training. As shown in Figure~\ref{fig:entropy_of_skywork-R1-math-7b}, the entropy of Skywork-OR1-Math-7B remains lower-bounded by the target entropy throughout the RL training process. To further validate the effectiveness of adaptive entropy control, we conduct the following ablation experiments.

\begin{figure}[!htbp]
    \centering
    \includegraphics[width=0.95\linewidth]{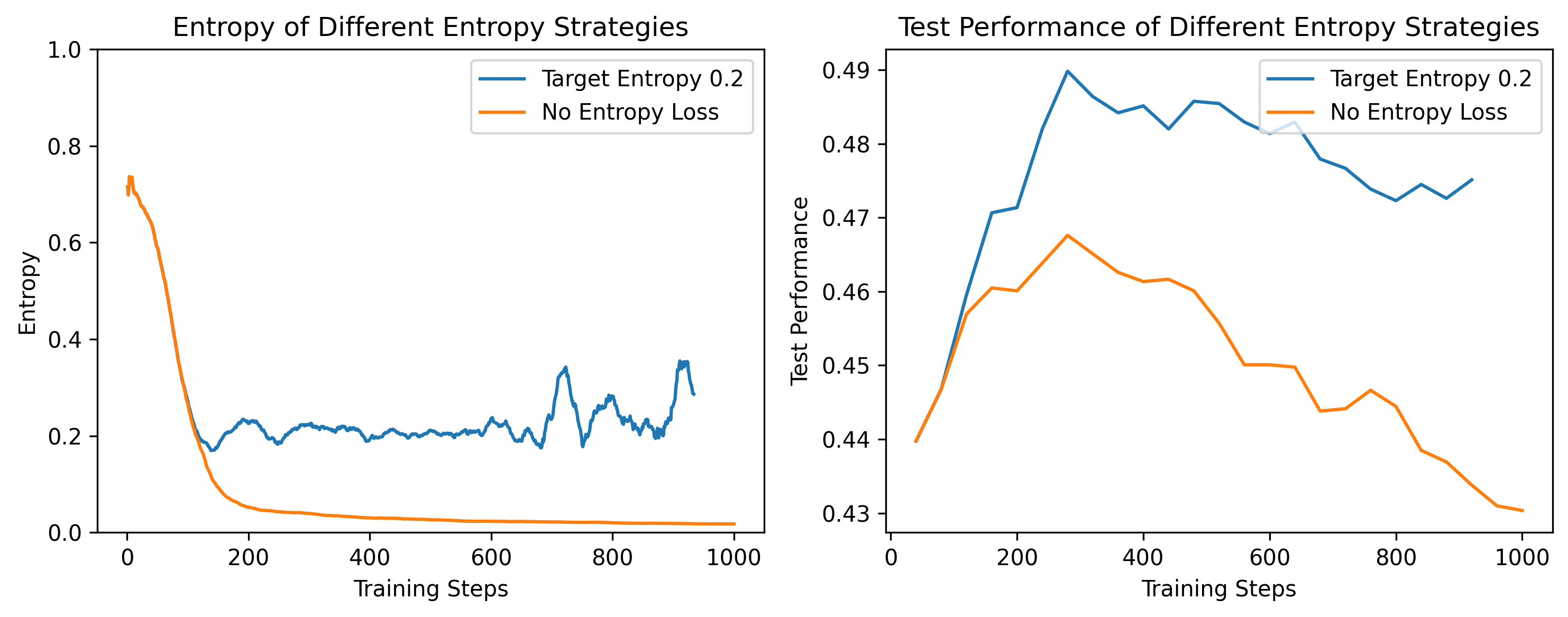}
    \vspace{-1em}
    \caption{The results of Ablation Experiments \arabic{ablation}. Applying adaptive entropy control prevents the entropy collapse, leading to a better test performance. \textbf{Left:} Entropy of generated responses during RL training. \textbf{Right:} Test performance during RL training.}
    \label{fig:target_entropy}
\end{figure}

\stepcounter{ablation}
\begin{mybox}{Ablation Experiments \arabic{ablation}: Effectiveness of Adaptive Entropy Control}
Consider the off-policy experiment in Ablation Experiments 6 with $(N_\text{SGD},D_R,D_T,N_\text{reuse})=(4,64,16,1) $, which exhibits fast entropy collapse and bad test performance. Note that there is no entropy loss in this experiment. We ran an experiment based on its configuration with adaptive entropy control (using a target entropy of 0.2) enabled. We report the results in Figure \ref{fig:target_entropy}.
\end{mybox} 
As previously analyzed, increasing $N_\text{SGD}$ accelerates policy convergence and leads to degraded test performance. As shown in Figure~\ref{fig:target_entropy}, applying adaptive entropy control successfully prevents entropy collapse and results in higher test performance. However, it is worth noting that, although the coefficient is adjusted adaptively, entropy remains unstable when $N_\text{SGD}$ is large. We speculate that this is due to the entropy loss being computed over the entire vocabulary, which may increase the probability of many unintended tokens.
Therefore, we do not recommend using adaptive entropy control in scenarios where $N_\text{SGD}$ is large. Nonetheless, we find that when $N_\text{SGD} = 1$ or $2$, entropy dynamics remain acceptably stable under adaptive entropy control. Based on these findings, we adopt adaptive entropy control in the training of our Skywork-OR1 models.

\paragraph{The Impact of the Clip-Higher Trick.} We tested a popular trick called clip-higher \cite{yu2025dapo} used in PPO-style policy loss to prevent the entropy collapse when $N_\text{SGD} > 1$. We conduct the following ablation experiments.

\begin{figure}[ht]
    \centering
    \includegraphics[width=0.95\linewidth]{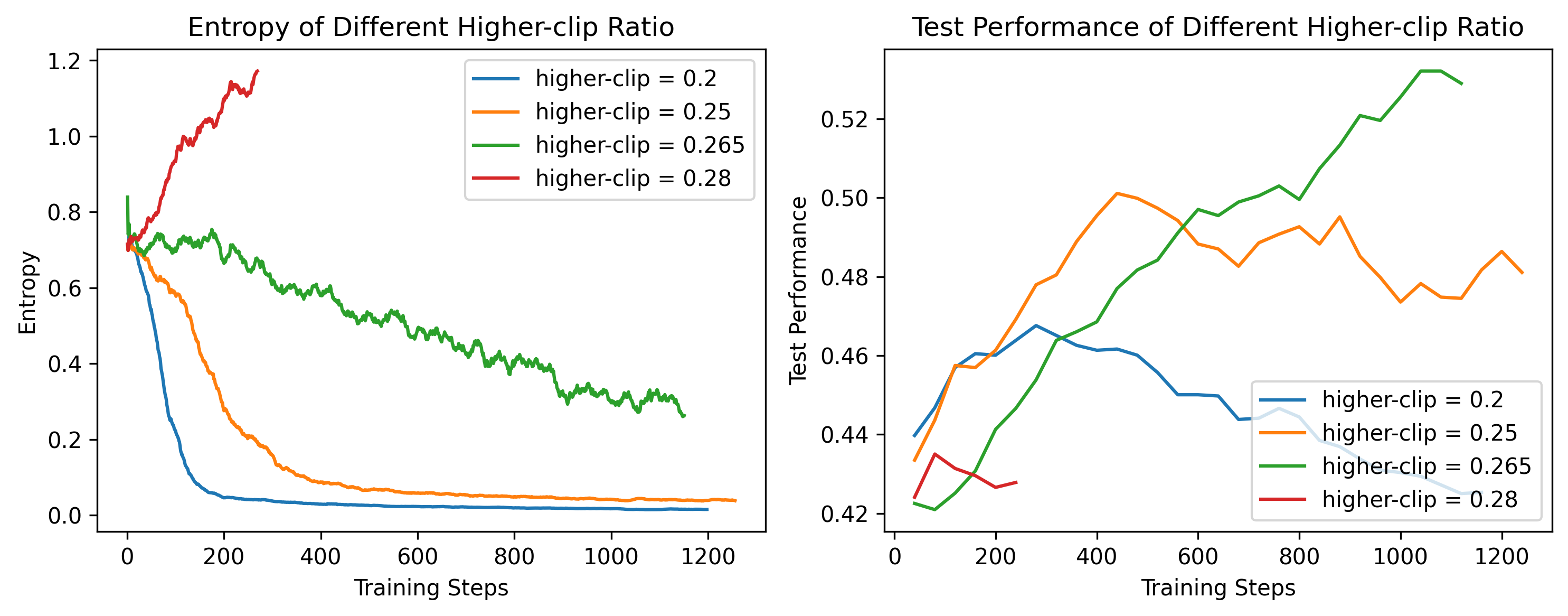}
    \vspace{-1em}
    \caption{The results of Ablation Experiments \arabic{ablation}. Increasing the higher-clip ratio to an adequate value (e.g., 0.25 and 0.265) yields slower convergence and better test performance. However, we find that when the higher-clip ratio is set to 0.28 as recommended in \cite{yu2025dapo}, then entropy rises sharply and test performance is not improved. \textbf{Left:} Entropy of generated responses during RL training. \textbf{Right:} Test performance during RL training.}
    \label{fig:clip_higher}
\end{figure}

\stepcounter{ablation}
\begin{mybox}{Ablation Experiments \arabic{ablation}: The Impact of Different Higher-clip Ratios}
Consider the off-policy experiment in Ablation Experiments 6 with the quadruple $(N_\text{SGD},D_R,D_T,N_\text{reuse})=(4,64,16,1) $, which exhibits fast entropy collapse and poor test performance. Note that the clip ratio $\epsilon=0.2$ was applied in this experiment. We raised the higher-clip ratio from $0.20$ to $0.25, 0.265$, and $0.28$ while keeping the lower-clip ratio fixed at $0.2$. We report the results in Figure \ref{fig:clip_higher}.
\end{mybox} 

Our results, shown in Figure~\ref{fig:clip_higher}, indicate that using a properly chosen higher-clip ratio -- e.g., 0.25 or 0.265 -- can prevent premature entropy collapse and lead to better test performance. However, it is worth noting that when the higher-clip ratio is set to 0.28, as suggested in \cite{yu2025dapo}, entropy increases sharply, resulting in poor test performance. This suggests that the optimal higher-clip ratio is task-dependent.

\section{Empirical Studies on Training Resource Allocation}
\label{sec:training_time_scaling}

During the RL training process, our goal is to select hyperparameters that make training both efficient and effective. This objective gives rise to two practical questions:

\begin{itemize}
    \item Given fixed computational resources, how can we improve training efficiency? 
    \item Given additional computational resources, how should we allocate them to achieve better test performance or improved training efficiency?
\end{itemize}

In this section, we address these questions in the context of long CoT scenarios, using results from exhaustive ablation experiments as supporting evidence. The training process of online RL algorithms can generally be divided into two distinct phases: \textbf{data rollout} and \textbf{policy update} (which includes both forward and backward passes). Let $t_R$, $t_T$, and $t_O$ denote the time spent on rollout, policy update, and other operations (e.g., reward computation, experience generation), respectively. The total time consumption under a synchronous training framework is:
\begin{align*}
t_{\mathrm{total}} = t_R + t_T + t_O.
\end{align*}
Given a fixed context length, the rollout time $t_R$ is primarily influenced by the rollout batch size $D_R$ and the group size ($gs$). As analyzed in Section~\ref{sec:sgd_steps}, the policy update time $t_T$ depends on the number of SGD steps $N_\text{SGD}$, which is determined by the number of mini-batches $\frac{D_R}{D_T}$ and the data reuse factor $N_\text{reuse}$. In the following subsections, we investigate how these factors impact both training efficiency and final performance.

\subsection{Improving Training Efficiency with Fixed Computational Resources}
\label{sec:sgd_steps_efficiency}
In this section, we aim to answer the first question: Given fixed computational resources, how can training efficiency be improved?

\begin{figure}[!htbp]
    \centering
    \includegraphics[width=1\linewidth]{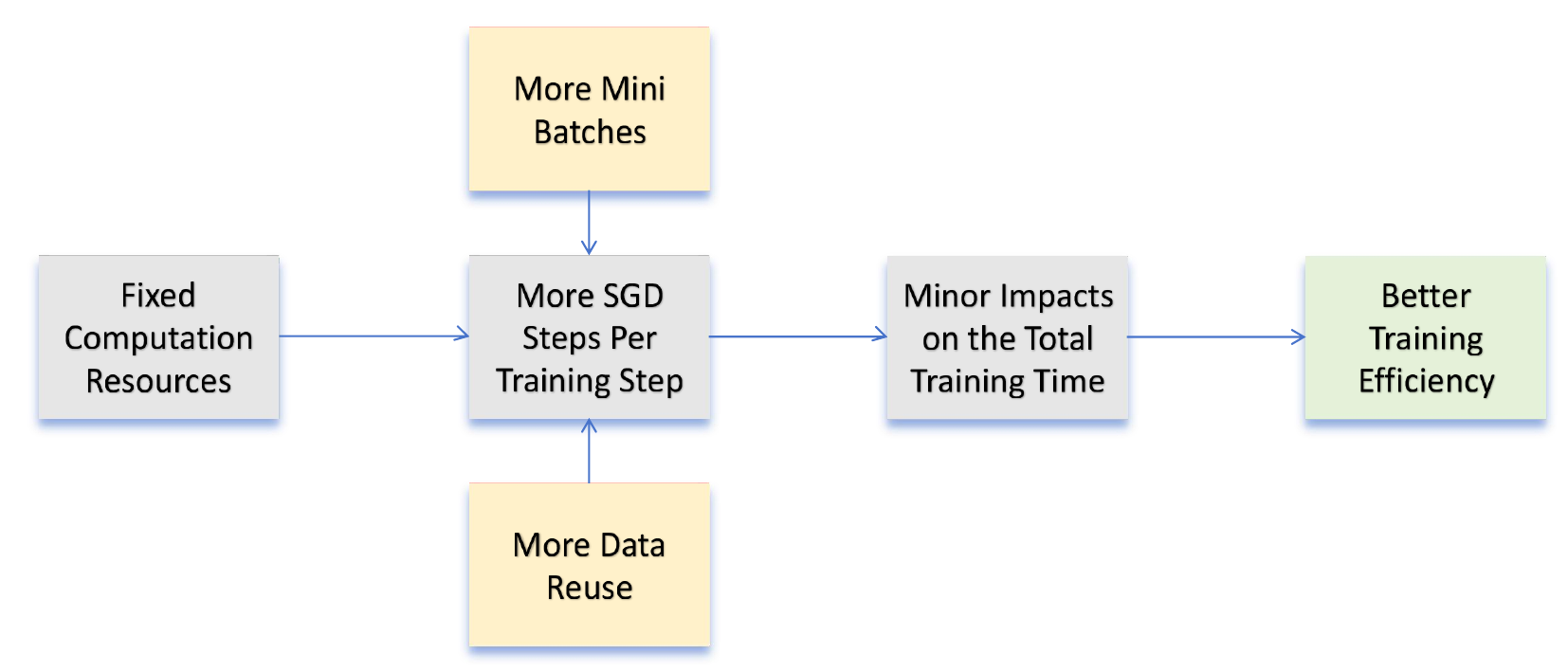}
    \caption{Overview of empirical studies on improving training efficiency given fixed computational resources. \textbf{Grey‌ blocks:} Potential approaches to enhance training efficiency and their underlying principles. \textbf{Yellow blocks:} Experimental variables in the empirical studies}
    \label{fig:fixed_resources}
\end{figure}

\paragraph{Rollout Time $t_R$ Dominates the Total Training Time $t_\text{total}$.} A fundamental observation regarding long CoT models (e.g. Deepseek-R1-Distill model series) is that the total training time is primarily determined by the rollout time.
Table \ref{table:time_usage_of_skywork_OR1_32b} presents the values of
$t_{\text{total}}$, $t_R$, $t_T$ and $t_O$ of Skywork-OR1-32B over 1000 training steps. Clearly, $t_R$ dominates $t_\text{total}$. 
\begin{table}[!htbp]
\centering
\begin{tabular}{@{\extracolsep{4pt}}ccccccc}
\toprule   
\textbf{Time Usage} & \makecell[cc]{total \\ $t_\text{total}$} & \makecell[cc]{rollout \\ $t_R$} & \makecell[cc]{policy update \\ $t_T$} & \makecell[cc]{others \\ $t_O$} & $t_R/t_\text{total}$ & $t_T/t_\text{total}$

\\
\midrule
\textbf{Hours} & 309 &  223 & 27  & 59 & 72.1\% & 8.7\%
\\
\bottomrule
\end{tabular}
\caption{Analysis of training time usage of Skywork-OR1-32B for 1000 training steps.} 
\label{table:time_usage_of_skywork_OR1_32b}
\end{table}

Since the primary bottleneck for $t_\text{total}$ in long CoT training is $t_\text{R}$, it is reasonable to expect that appropriately increasing the number of SGD steps per training step, i.e., $N_\text{SGD}$, will have minimal impact on $t_\text{total}$ while improving training efficiency. Therefore, in the following, we investigate the impact of the number of mini-batches ($\frac{D_R}{D_T}$) and the data reuse times ($N_\text{reuse}$) on both the total training time $t_\text{total}$ and test performance. The overall idea of our study is illustrated in Figure~\ref{fig:fixed_resources}.

\paragraph{More SGD Steps, More Training Efficiency but Worse Performance.} 
We have already examined the impact of increasing $N_\text{SGD}$ on entropy dynamics, as discussed in Ablation Experiments 6 (Section~\ref{sec:sgd_steps}). Consider the configuration tuple $(N_\text{SGD}, D_R, D_T, N_\text{reuse})$. We report the detailed time usage for the configurations (1, 64, 64, 1), (2, 64, 32, 1), and (4, 64, 16, 1) in Table~\ref{table:time_usage_of_hyper_ent}. It is evident that increasing $N_\text{SGD}$ leads to a higher $t_T$. However, the impact on the overall training time $t_\text{total}$ remains minor, provided that $D_R$ is fixed. Thus, the configurations with $N_\text{SGD} \in \{2, 4\}$ perform multiple SGD steps within comparable training time, improving training efficiency. That said, the experimental results in Section~\ref{sec:sgd_steps} show that accelerating training via rollout batch decomposition or data reuse leads to faster entropy collapse and poorer test performance. Therefore, we do not recommend increasing $N_\text{SGD}$ solely for the purpose of improving training efficiency -- unless appropriate mechanisms are in place to mitigate entropy collapse, particularly those caused by off-policy updates -- as doing so may result in degraded generalization performance.

\begin{table}[!htbp]
\centering
\begin{tabular}{@{\extracolsep{4pt}}ccccccc}
\toprule   
\makecell[cc]{Experiment \\ $(N_\text{SGD},D_R,D_T,N_\text{reuse})$} & \makecell[cc]{total \\ $t_\text{total}$} & \makecell[cc]{rollout \\ $t_R$} & \makecell[cc]{policy update \\ $t_T$} & \makecell[cc]{others \\ $t_O$} & $t_R/t_\text{total}$ & $t_T/t_\text{total}$
\\
\midrule
\textbf{(1,64,64,1)} & 116 &  90 & 8  & 18 & 77.6\% & 6.9\%
\\
\textbf{(2,64,32,1)} & 114 &  87 & 10  & 17 & 76.3\% & 8.7\%
\\
\textbf{(4,64,16,1)} & 118 &  90 & 12  & 16 & 76.3\% & 10.2\%
\\
\bottomrule
\end{tabular}
\caption{Detailed time usage for three experiments from Ablation Experiments 6 over 1000 training steps. All the experiments utilized the same training resources (i.e., 32 H800 GPUs).} 
\label{table:time_usage_of_hyper_ent}
\end{table}

\subsection{Improving Test Performance with More Computational Resources}
\begin{figure}[!htbp]
    \centering
    \includegraphics[width=0.95\linewidth]{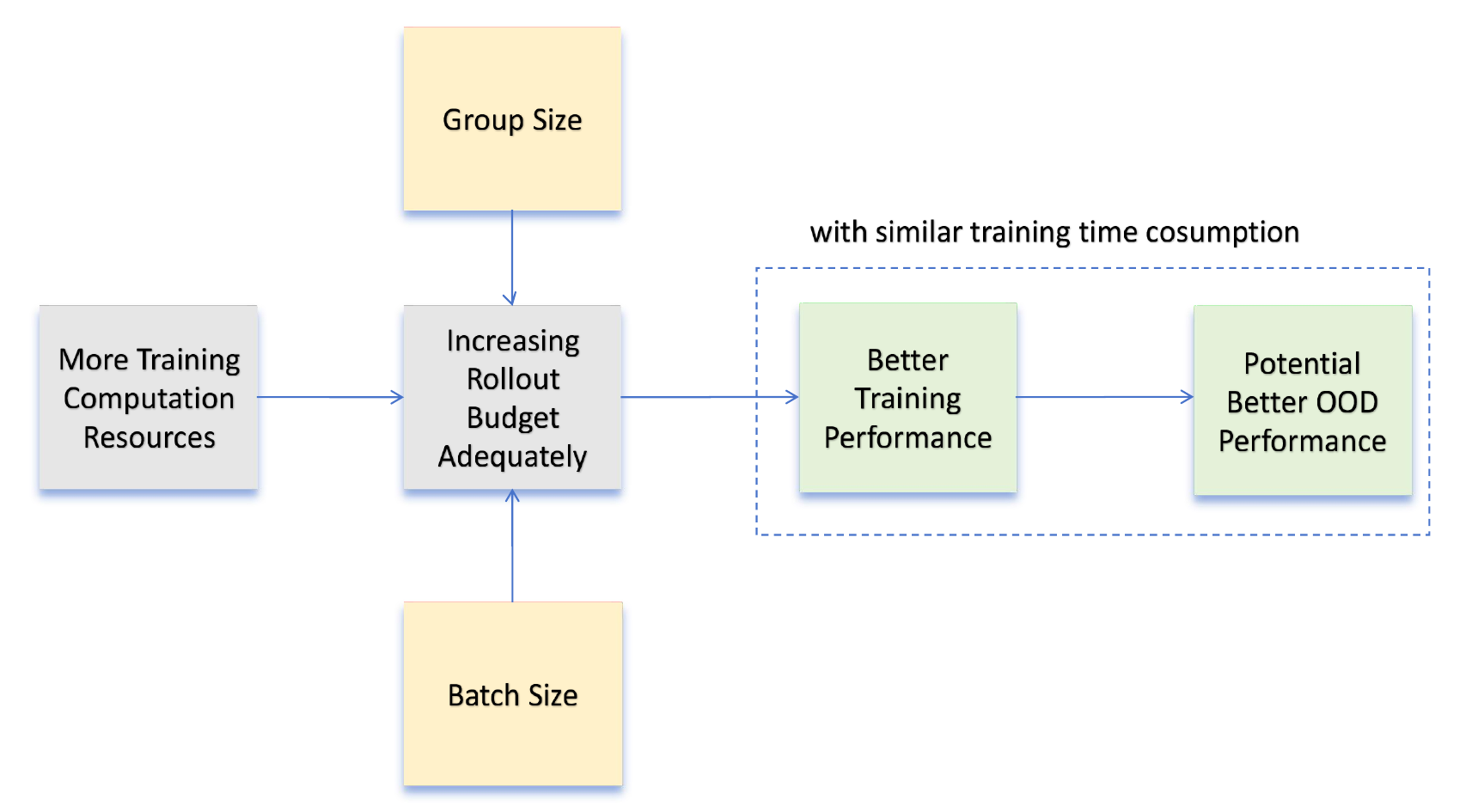}
    \caption{Overview of empirical studies on the effect of an increased rollout budget when more training resources are available. \textbf{Grey and green blocks}: The motivation of the empirical studies. \textbf{Yellow blocks}: The experimental variables in the empirical studies.}
    \label{fig:more_computation_resources}
\end{figure}

In this section, we address the second question: given more computational resources, how should training resources be allocated to achieve higher test performance or better training efficiency? Regarding training efficiency, two approaches may be considered. On the one hand, increasing the number of SGD steps -- previously discussed -- may seem promising. However, experimental findings do not support the effectiveness of this approach (see Section~\ref{sec:sgd_steps_efficiency}). On the other hand, under a fixed rollout budget (i.e., the number of samples to be rolled out), one might expect a significant reduction in rollout time $t_R$ as training resources are scaled up. In practice, however, this expectation is not fully realized.
 Table \ref{table:rollout_time_of_different_resources}
\begin{table}[!htbp]
\centering
\begin{tabular}{@{\extracolsep{4pt}}ccccccc}
\toprule   
\textbf{The number of H800} & 32  & 64 & 128 & 256
\\
\midrule
\textbf{Rollout time $t_R$ (reduction)} & 375 & 270 (-105) & 225 (-45) & 205 (-20)
\\
\bottomrule
\end{tabular}
\caption{Rollout time $t_R$ (seconds) for generating 1024 responses in one training step. The data shows that as computational resources increase, the incremental reduction in $t_R$ diminishes.} 
\label{table:rollout_time_of_different_resources}
\end{table}
shows the rollout time $t_R$ for 1024 samples under varying training resources. Notably, as training resources increase, the reduction in $t_R$ diminishes. This is because $t_R$ is primarily determined by the batch size and the time required to generate the longest response. Once sufficient resources are available, further scaling does not significantly reduce the processing time dominated by the generation of the longest sample. Therefore, when additional training resources are available, a more effective strategy is to increase the rollout budget appropriately, such that the rollout time $t_R$ remains roughly constant or increases only marginally. By leveraging a larger rollout buffer, more accurate gradient estimates can be obtained, which may improve training efficiency and enhance test performance. In the following, we focus on how the rollout budget -- determined by the rollout batch size and group size -- affects RL performance. The overall idea of these studies are illustrated in Figure \ref{fig:more_computation_resources}

\paragraph{Larger Batch Size, Better Test Performance.}
To investigate how the rollout batch size $D_R$ affects the training dynamics, we conducted the following ablation experiments. 

\begin{figure}[ht]
    \centering
    \includegraphics[width=0.95\linewidth]{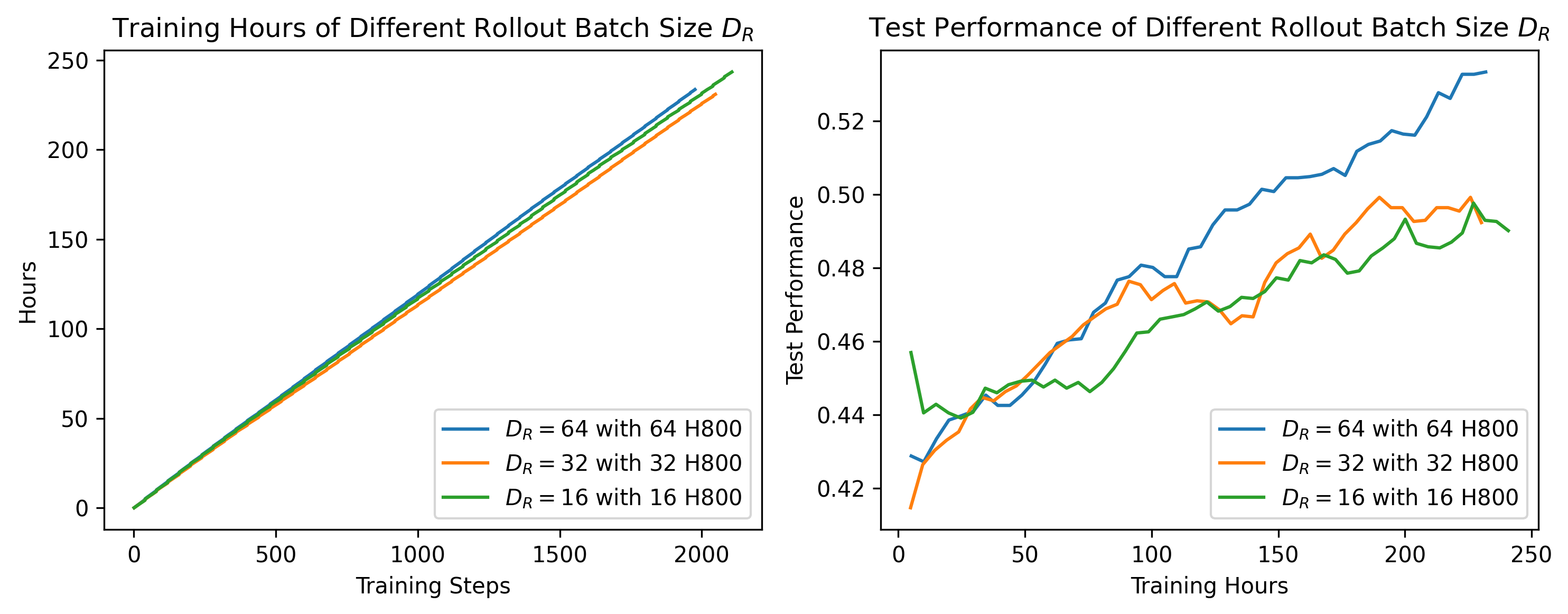}
    \vspace{-1em}
    \caption{Results of Ablation Experiments \arabic{ablation}. Given more training resources, increasing the rollout budget by increasing $D_R$  achieves better test performance with similar total training hours.}
    \label{fig:batch_size_time_compare}
\end{figure}

\stepcounter{ablation}
\begin{mybox}{Ablation Experiments \arabic{ablation}: The Impact of Rollout Batch Size $D_R$}
Consider the  quadruple $(N_\text{SGD},D_R,D_T,N_\text{reuse})$.
We consider the baseline experiment with the quadruple (1,64,64,1) in Section \ref{sec:ablation_setup} and two on-policy experiments in Ablation Experiments 7 with the quadruples (1,32,32,1) and (1,16,16,1) respectively. These three experiments were conducted using 64,32 and 16 H800 respectively. We present the experimental results in Figure \ref{fig:batch_size_time_compare}.
\end{mybox} 
The results in Figure \ref{fig:batch_size_time_compare} indicate that increasing the rollout batch size $D_R$ in accordance with available training resources can lead to better test performance with similar training time consumption.

\begin{figure}[!htbp]
    \centering
    \includegraphics[width=0.95\linewidth]{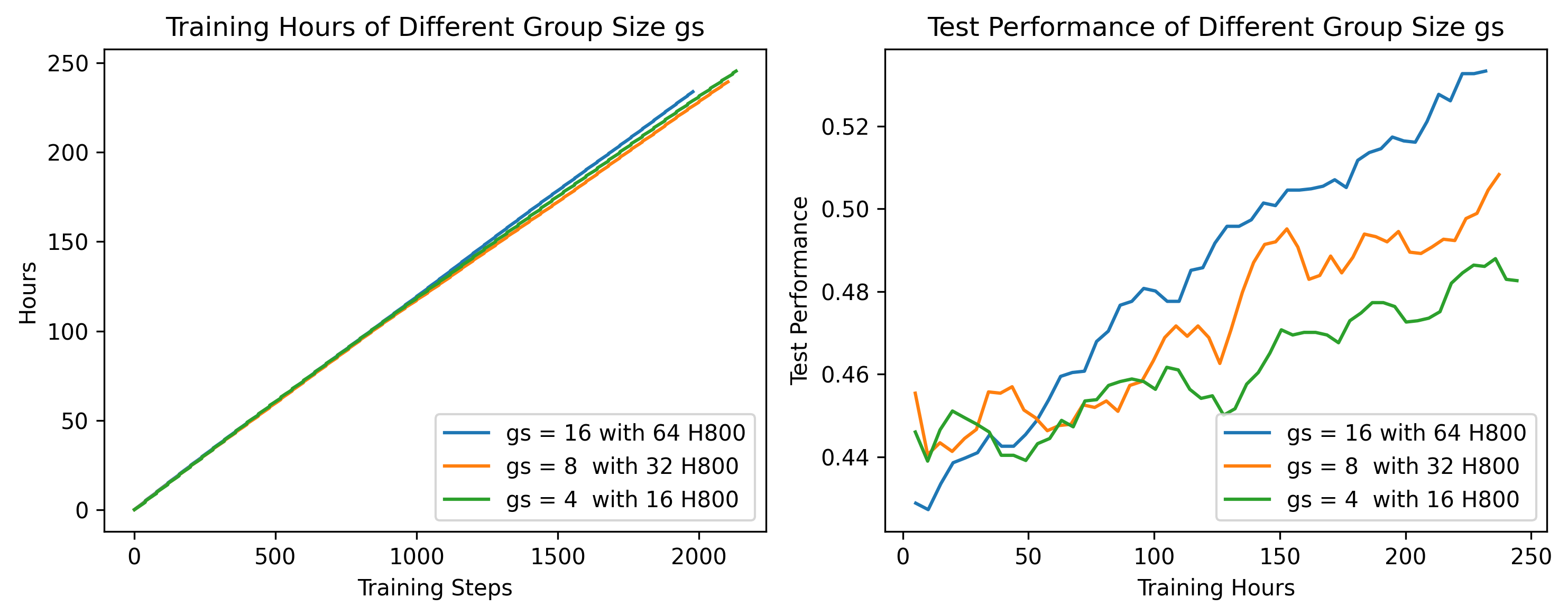}
    \vspace{-1em}
    \caption{Results of Ablation Experiments \arabic{ablation}. Given more training resources, increasing rollout budget by increasing the group size can achieve better test performance with similar total training hours.}
    \label{fig:group_size_time_comparison}
\end{figure}

\paragraph{Larger Group Size, Better Test Performance.}  
To investigate how the group size affects the training dynamics, we conducted the following ablation experiments. 
\stepcounter{ablation}
\begin{mybox}{Ablation Experiments \arabic{ablation}: The Impact of Group Size ($gs$)}
Consider the baseline experiment with group size 16 in Section \ref{sec:ablation_setup}. We ran two additional on-policy experiments with $\text{gs}=8,4$ respectively. These three experiments were conducted using 64,32 and 16 H800 respectively. The experimental results are presented in Figure \ref{fig:group_size_time_comparison}.
\end{mybox} 
It can be observed from Figure \ref{fig:group_size_time_comparison}, given more training resources, increasing rollout budget by increasing the group size can lead to a better test performance with similar total training hours.

\section{Dataset Preparation}
\label{sec:dataset}

In this section, we introduce the processing pipeline for our RL training data.

\subsection{Data Source Selection and Preprocessing}
For the math domain, we primarily focus on NuminaMath-1.5 \cite{numina_math_datasets}, a comprehensive dataset containing 896K math problems drawn from widely used sources and advanced mathematical topics. Although the dataset is sufficiently large, its quality requires careful examination prior to use.

For the code domain, we find that data source options are more limited, and the overall difficulty of available datasets is generally low relative to the capabilities of current models. In our pilot studies, we experimented with several popular datasets -- including CODE-RL \cite{coderl}, TACO \cite{taco}, and the Eurus-RL collection \cite{eurusrl} -- in their original mixtures, but obtained unsatisfactory results.

\paragraph{Selection Criteria} To select and curate high-quality data for RL, we adhere to the following general criteria for both data domains:
\begin{enumerate}
    \item \textbf{Verifiable}: We exclude problems that cannot be verified, such as proof-based problems and code problems lacking test cases.
    \item \textbf{Correct}: We filter out math problems with invalid or incorrect answers, as well as code problems without comprehensive test cases.
    \item \textbf{Challenging}: We pre-filter problems for which all N generations from the base model are either entirely correct or entirely incorrect.
\end{enumerate}
Following these criteria, we incorporate challenging problems from NuminaMath-1.5 and other sources to enhance problem difficulty and diversity in our data mixture: 1) NuminaMath-1.5 subsets: amc\underline{ }aime, olympiads, olympiads\underline{ }ref, aops\underline{ }forum, cn\underline{ }contest, inequalities, and number\underline{ }theory. 2) DeepScaleR. 3) STILL-3-Preview-RL-Data. 4) Omni-MATH. 5) AIME problems prior to 2024. For the code data mixture, we primarily consider problems from the following two sources, which offer sufficiently challenging coding questions: 1) LeetCode problems~\cite{xia2025leetcodedataset}. 2) TACO~\cite{li2023taco}.

\paragraph{Preprocessing Pipeline} For both math and coding problems, we first perform in-dataset deduplication to eliminate redundancy. For all collected math problems:
\begin{itemize}
    \item We use Math-Verify~\cite{mathverify2025} to re-extract answers from the provided textual solutions and retain only those problems where the extracted answer matches the corresponding answer in the dataset.
    \item We remove all instances that contain external URLs or potential figures in the problem statement.
    \item We then perform cross-dataset deduplication to eliminate potentially duplicated problems from similar sources and decontaminate against AIME24 and AIME25 problems, following DeepScaleR’s deduplication scheme.
\end{itemize}
This process yields approximately \textbf{105K math problems.} For coding problems, we apply a more rigorous filtering process as follows:
\begin{itemize}
    \item We discard samples with empty, incomplete, or corrupted original unit test cases.
    \item We programmatically verify all test cases using the provided original solutions. A sample is marked as valid only if the solution passes all corresponding test cases perfectly.
    \item We conduct extensive deduplication based on embedding similarity across the collected coding problems, as many share the same problem with only slight variations in instructions.
\end{itemize}
This results in a total of \textbf{13.7K coding questions (2.7K from LeetCode and 11K from TACO)} in the final dataset.

\subsection{Model-Aware Difficulty Estimation}
Due to the zero-advantage in GRPO when all sampled responses are either entirely correct or entirely incorrect within a group, we conduct an initial offline difficulty estimation for each problem relative to the models being trained. Specifically, \textbf{for each problem, we perform N=16 rollouts for math problems and N=8 for coding questions using a temperature of 1.0 and a maximum token length of 32K}, and use the percentage of correct solutions as a proxy for problem difficulty with respect to a given model. After verifying the correctness of the sampled solutions, we exclude problems with 0/N (all incorrect) or N/N (all correct) rollouts. We report the percentage statistics of discarded and retained math/code problems for both the 7B and 32B models as follows:

\begin{table}[!htbp]
\centering
\begin{tabular}{@{\extracolsep{4pt}}c|cccccc}
\toprule   
 \multirow{2}{*}{} &  $\frac{0}{N}$ Correct  &  $\frac{N}{N}$ Correct & Remaining 
 \\ 
 &  (math/code)  & (math/code) & (math/code)
 
\\
\midrule
Deepseek-R1-Distill-Qwen-7B &  21.4\% / 28\% &  32.4\% / 24\% & 46.2\% / 48\%
\\
\midrule
Deepseek-R1-Distill-Qwen-32B  &  20.7\% / 17.1\% &  42.0\% / 45.4\% & 37.3\% / 37.6\% 
\\
\bottomrule
\end{tabular}
\label{table:shared_parameter_in_ablation_5}
\end{table}

\subsection{Quality Assessment via Human and LLM-as-a-Judge}
During the data processing stage, we identified that many problems in the math portion were either incomplete or poorly formatted. Consequently, we conducted an additional round of strict human-LLM-combined inspection to ensure data quality. We sampled a few hundred questions from the remaining pool and asked human evaluators to assess whether each problem met the following criteria:
\begin{enumerate}
    \item \textbf{Clear Wording}: Is the problem stated in a way that is easy to understand?
    \item \textbf{Complete Information}: Does the problem provide all necessary details?
    \item \textbf{Good Formatting}: Are the numbers, symbols, and equations clear and appropriately formatted?
    \item \textbf{No Distractions}: Is the problem free of irrelevant information?
\end{enumerate}
We provide below examples of original problem statements that human evaluators identified as problematic:
\begin{tcolorbox}
\textbf{Incomplete Problems:}
\vspace{5pt}
\begin{itemize}[leftmargin=5pt,rightmargin=5pt,itemsep=5pt]
    \item \textit{6. Five spherical surfaces can divide space int  $\qquad$ parts.} (NuminaMath-1.5, Olympiads)
    \item \textit{Which of the following numbers is equal to 33 million?} (STILL-3-Preview-RL-Data)
    \item \textit{Which number is greater than 0.7}  (STILL-3-Preview-RL-Data)
    \item \textit{Example 27 Find $\sigma_{2}(28)=$ ?}  (NuminaMath-1.5, Number Theory)
\end{itemize}
\end{tcolorbox}
\begin{tcolorbox}
\textbf{Irrelevant Information:} 
\vspace{5pt}
\begin{itemize}
    \item \textit{250. $y=\ln \left(x^{3}-1\right)$.$\textbackslash$n$\textbackslash$n 250. $y=\ln \left(x^{3}-1\right)$.$\textbackslash$n$\textbackslash$n The above text has been translated into English, retaining the original text's line breaks and format. However, since the original text is a mathematical expression, the translation is identical to the original as mathematical expressions are generally universal and do not change between languages.} (NuminaMath-1.5, Olympiads)
    \item \textit{1. (12 points) The figure is composed of 5 identical squares. The number of triangles that can be formed using the 12 points in the figure as vertices is.10. (12 points) The figure is composed of 5 identical squares. The number of triangles that can be formed using the 12 points in the figure as vertices is.}  (NuminaMath-1.5, Olympiads)
\end{itemize}
\end{tcolorbox}

Interestingly, these problems passed the difficulty estimation procedure (i.e., a model can produce a correct answer even when the problem is invalid or incomplete). This indicates that the models answered these problems correctly at least once during the 16 rollouts, suggesting they may have been trained on similar examples or that the answers were trivially guessable.

To efficiently curate the entire dataset, we employed Llama-3.3-70B-Instruct and Qwen2.5-72B-Instruct to automatically filter out low-quality problems. Each model was prompted to evaluate a given math problem based on clarity, completeness, formatting, and relevance, and to identify reasons a problem might be considered low quality, ultimately providing a binary rating. This process mimics human assessment while being significantly more efficient. For each problem and each LLM judge, we collected 16 evaluations, resulting in a total of 32 votes per problem. We retained problems that received at least 9 valid votes and removed approximately 1K-2K math questions in total.

\section{Math \& Code Verifiers}
\label{sec:verifiers}

\subsection{Math Verifiers}
During the initial stage of all experiments on math reasoning, we conducted several preliminary analyses of the rule-based math verifiers available at the time. These verifiers included:
\begin{itemize}
    \item The original MATH verifier (verl version)
    \item PRIME verifier
    \item Qwen2.5 verifier
    \item DeepScaleR’s verifier
    \item Math-Verify
\end{itemize}
We first sampled a small set of problems along with their associated solutions and answers, and manually examined the quality of their parsers and verifiers. We found that the Qwen2.5 verifier tends to lose information during the parsing process (e.g., when parsing \verb|\boxed{a^2}} $|, it fails to retain \verb|^2|). We also observed that the PRIME verifier can occasionally stall during execution. As a result, we excluded these two verifiers from further analysis.

We then used rollout data from the difficulty estimation procedure and applied the remaining verifiers to evaluate the generated solutions. We plotted the number of problems at each difficulty level (0--8) in Figure~\ref{fig:reward_models_1}:

\begin{figure}[!htbp]
    \centering
    \includegraphics[width=1\linewidth]{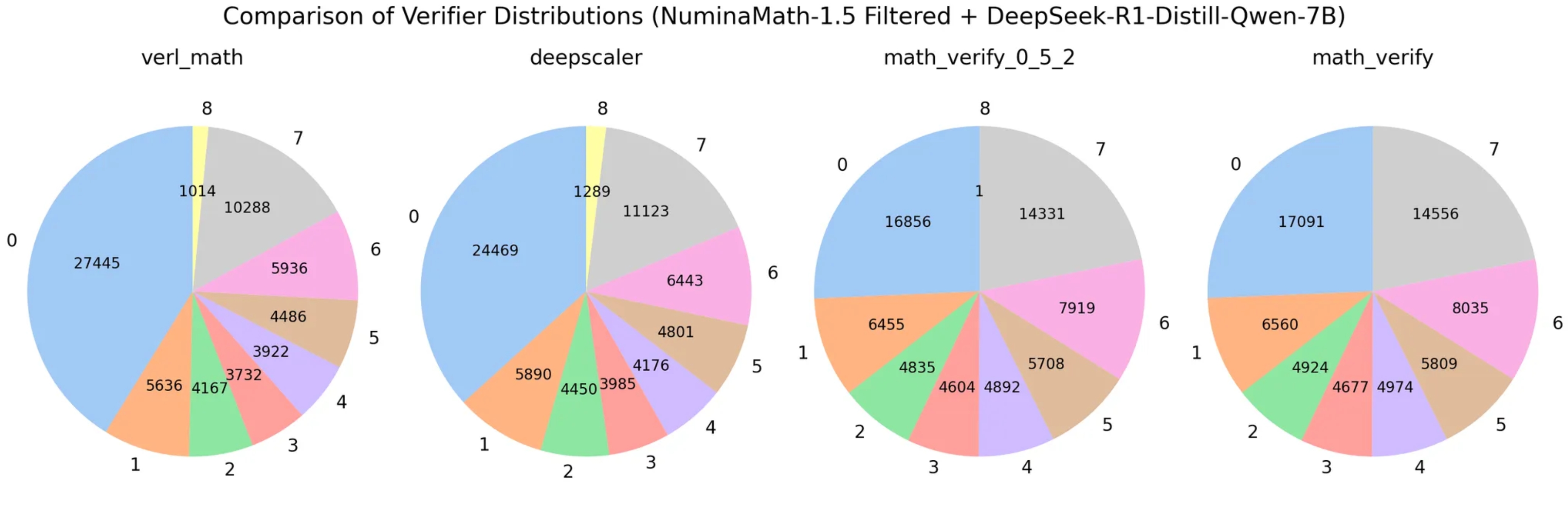}
    \vspace{-1em}
    \caption{Distributions of the number of correct rollouts from DeepSeek-R1-Distill-Qwen-7B, obtained using four different verifiers on a subset of NuminaMath-1.5 problems. The numbers 0--8 indicate difficulty levels. The size of each sector represents the number of problems at a specific difficulty level.}
    \label{fig:reward_models_1}
\end{figure}

Based on a combination of verifier results and human judgments, we observed the following:
\begin{itemize}
    \item Both the original MATH verifier (verl version) and DeepScaleR’s verifier produced higher rates of false positives and false negatives.
    \item For Math-Verify, some implementation details changed as we explored different versions. Therefore, we include both version 0.5.2 and the default version (0.6.0), which we extensively used in model development, noting only trivial differences between them.
\end{itemize}

Note that Math-Verify may still yield incorrect results for solutions with non-standard formatting or mathematical expressions it does not support (e.g., problems with multiple answers).

In our final implementation of the reward function, we verify whether the answer in a text solution is correct using the following steps:

\begin{itemize}
    \item Extract the answer that appears after the reasoning process.
    \item Use Math-Verify’s parser to parse the answer and obtain its string representation.
    \item If the string representation directly matches the gold answer, return True; otherwise, fall back to Math-Verify’s verify function.
    \item Wrap the gold answer in $\text{\\} \text{boxed}\{\}$ and run the verification to obtain the final result.
\end{itemize}

We find that wrapping the gold answer with $\text{\\} \text{boxed}\{\}$ is a crucial step. Parsing the gold answer directly can alter the mathematical expression.

\subsection{Code Sandboxes}
For unit test execution, we constructed a highly efficient and secure local code sandbox based on LiveCodeBench’s implementation, leveraging subprocess processing. This sandbox supports various testing methods, including standard input-output testing, solution function unit testing, and assertion-based tests. To further enhance its security and robustness, we implemented the following measures:
\begin{itemize}
    \item \textbf{Syntax validation:} We first validate submitted solutions using Abstract Syntax Trees (AST). If syntax errors are detected, the sandbox immediately terminates the test and returns False.
    \item \textbf{Memory monitoring:} During training, we identified potential memory leak risks in some generated solutions. To mitigate this, we integrated a memory monitoring mechanism for each test process. If a process exceeds 50GB of memory usage, the sandbox proactively terminates the test and returns False, effectively preventing resource exhaustion.
    \item \textbf{Parallel stability optimization:} Initially, we used asynchronous testing combined with process pools for parallel execution. However, we later discovered that the sandbox could crash under this setup, leading to incorrect test results. To resolve this, we revised our approach to rely solely on multiprocessing, ensuring stable and efficient parallel execution.
\end{itemize}

Additionally, we conducted a performance comparison between our sandbox and the PRIME sandbox. The results demonstrate the superior effectiveness of our implementation on specific datasets. Notably, the PRIME sandbox occasionally misclassified correct solutions as failures, whereas our sandbox more accurately evaluated solution correctness.

It is also important to note a limitation of our sandbox identified during practical usage: it does not currently handle cases where the same input can yield multiple valid outputs. Such cases are common in real-world code testing scenarios involving non-deterministic or open-ended problems.

\section{Experiments}
\label{sec:experiments}
In this section, we present the experimental results of our three models: Skywork-OR1-Math-7B, Skywork-OR1-7B, and Skywork-OR1-32B. We begin with the details of the training configurations, followed by an analysis of the training results. Finally, we discuss the evaluation outcomes.

\subsection{Training and Evaluation Details}

\paragraph{Training Configurations} Below, we describe the training configurations of our Skywork models. The 7B and 32B models are fine-tuned based on DeepSeek-R1-Distill-Qwen-7B and DeepSeek-R1-Distill-Qwen-32B, respectively. We collect math and code problems from various sources and apply comprehensive preprocessing, difficulty filtering, and quality control. This ensures a problem mixture that is verifiable, valid, and challenging. See Section~\ref{sec:dataset} for details.
Based on this curated mixture, all three models are fine-tuned by optimizing the policy loss~\eqref{policy_loss} with a constant learning rate of 1e-6, clip ratio of 0.2, target entropy of 0.2, sampling temperature of 1.0, and rejection sampling. Notably, we do not apply any KL loss in our training process, as discussed in Section~\ref{sec:kl}. Please refer to Section~\ref{sec:policy_update} for more details on the policy update procedure. All experiments use multi-stage training.
We report the detailed configuration for each training stage in Table~\ref{table:skywork-or1-math-7b-config}, Table~\ref{table:skywork-or1-7b-config}, and Table~\ref{table:skywork-or1-32b-config}. The released checkpoints correspond to step 2160 for Skywork-OR1-Math-7B, step 1320 for Skywork-OR1-7B, and step 1000 for Skywork-OR1-32B.

\begin{table}[!htbp]
\centering
\begin{tabular}{@{\extracolsep{4pt}}ccccccc}
\toprule   
Stage & Steps &  Context Length  $T$ & Batch Size & Mini-batch Size  & Group Size  
\\
\midrule
 1 & 0-740 & 8K & 256 & 128  & 16 
\\
 2 & 740-1740 & 16K & 256 & 128  & 16 
\\
 3 & 1740-2080 & 32K & 256 & 128  & 16 
 \\
 3.5 & 2080-2160 & 32K & 128 & 64  & 64 
\\
\bottomrule
\end{tabular}
\caption{Training configurations of Skywork-OR1-Math-7B.} 
\label{table:skywork-or1-math-7b-config}
\end{table}

\begin{table}[ht]
\centering
\begin{tabular}{@{\extracolsep{4pt}}ccccccc}
\toprule   
Stage & Steps & Context Length $T$ & Batch Size & Mini-batch Size  & Group size  
\\
\midrule
 1 & 0-660 & 16K & 256 & 256  & 16 
\\
 2 & 660-1320 & 32K & 160 & 160  & 32 
\\
\bottomrule
\end{tabular}
\caption{Training configurations of Skywork-OR1-7B.} 
\label{table:skywork-or1-7b-config}
\end{table}

\begin{table}[!htbp]
\centering
\begin{tabular}{@{\extracolsep{4pt}}ccccccc}
\toprule   
Stage & Steps &  Context Length  $T$ & Batch Size & Mini-batch Size  & Group Size  
\\
\midrule
 1 & 0-760 & 16K & 256 & 256  & 16 
\\
 2 & 760-1130 & 24K & 160 & 160  & 32 
\\
\bottomrule
\end{tabular}
\caption{Training configurations of Skywork-OR1-32B.} 
\label{table:skywork-or1-32b-config}
\end{table}

\paragraph{Benchmarks \& Baselines} We evaluate our models on challenging benchmarks. For math capabilities, we assess performance on the American Invitational Mathematics Examination (AIME) 2024 and 2025. For coding capabilities, we use LiveCodeBench~\cite{jain2024livecodebench} (from 2024-08 to 2025-02). We compare against several strong baselines, including DeepSeek-R1~\cite{deepseekai2025deepseekr1incentivizingreasoningcapability}, Qwen3-32B~\cite{qwen3}, QwQ-32B~\cite{qwq32b}, Light-R1-32B~\cite{lightr1proj}, TinyR1-32B-Preview~\cite{tinyr1proj}, and several 7B RL models based on DeepSeek-R1-Distill-Qwen-7B, such as AceReason-Nemotron-7B~\cite{nv_ace_reasoner}, AReaL-boba-RL-7B~\cite{areal2025}, and Light-R1-7B-DS~\cite{lightr1proj}.

\paragraph{Evaluation Setup} We set the maximum generation length to 32,768 tokens for all models. For AIME24/25, we report avg@32 performance; for LiveCodeBench (2024-08 to 2025-02), we report avg@4 performance. Responses are generated using a temperature of 1 and top-p of 1. The avg@$n$ metric is defined as
\begin{align*}
\mathrm{avg}@n = \frac{1}{n}\sum_{i=1}^n{\mathbb{I} \left\{ \left( x, y_i \right) \,\,\text{is correct} \right\}},
\end{align*}
where $x$ is the evaluation question and $y_i$ is the $i$-th response.

\subsection{Evaluation Results of Skywork-OR1 models}

\begin{table}[ht]
\centering
\begin{tabular}{@{\extracolsep{4pt}}lcccccccc}
\toprule   
\textbf{Model}  & \makecell[cc]{\textbf{AIME 24}\\avg@32} & \makecell[cc]{\textbf{AIME 25} \\ avg@32} & \makecell[cc]{\textbf{LiveCodeBench} \\ \textbf{(2024-08 - 2025-02)} \\ avg@4 }
\\
\midrule
\multicolumn{4}{c}{\textbf{7B Models}}
\\
\midrule
\textbf{DeepSeek-R1-Distill-Qwen-7B}  & 55.5    & 39.2 &  37.6 
\\
\textbf{Light-R1-7B-DS	}  & 59.1    & 44.3	 &  39.5 
\\
\textbf{AReaL-boba-RL-7B}  & 61.9   & 48.3	 &  -
\\
\textbf{AceReason-Nemotron-7B } & 69.0 & 53.6 & \textbf{51.8}
\\
\textbf{Skywork-OR1-Math-7B	}  & 69.8   & 52.3  &  43.6
\\
\textbf{Skywork-OR1-7B	}  & \textbf{70.2}   & \textbf{54.6}  &  47.6
\\
\midrule
\multicolumn{4}{c}{\textbf{$\ge$32B Models}}
\\
\midrule
\textbf{DeepSeek-R1-Distill-Qwen-32B}  & 72.9    & 59.0  &  57.2 
\\
\textbf{TinyR1-32B-Preview	}  & 78.1    & 65.3  &  61.6
\\
\textbf{Light-R1-32B}  & 76.6   & 64.6 &  -
\\
\textbf{QwQ-32B}  & 79.5   & 65.3  &  61.6
\\
\textbf{Qwen3-32B	}  & 81.4   & 72.9  &  65.7
\\
\textbf{DeepSeek-R1}  & 79.8   & 70.0  &  \textbf{65.9}
\\
\textbf{Skywork-OR1-32B	}  & \textbf{82.2}   & \textbf{73.3} &  63.0
\\
\bottomrule
\end{tabular}
\caption{Comparison of Skywork-OR1 models and other models on reasoning-related benchmarks.}
\label{table:eval_results}
\end{table}

As shown in Table~\ref{table:eval_results}, Skywork-OR1 models achieve significant improvements over their base SFT models (e.g., the DeepSeek-R1-Distill series). Specifically, Skywork-OR1-32B achieves scores of 82.2 on AIME24, 73.3 on AIME25, and 63.0 on LiveCodeBench, outperforming strong contemporary models such as DeepSeek-R1 and Qwen3-32B on key math benchmarks, setting new SOTA records at the time of release. Skywork-OR1-7B scores 70.2 on AIME24, 54.6 on AIME25, and 47.6 on LiveCodeBench, demonstrating competitive performance relative to similarly sized models across both math and coding tasks. Our earlier released model, Skywork-OR1-Math-7B, also delivers competitive results among models of similar size, scoring 69.8 on AIME24, 52.3 on AIME25, and 43.6 on LiveCodeBench. These SOTA results are especially noteworthy given that they are obtained through fine-tuning the DeepSeek-R1-Distill series -- SFT base models with relatively modest initial performance -- clearly demonstrating the substantial impact of our pipeline.

\section{Conclusion}
In this work, we present Skywork-OR1, an effective and scalable reinforcement learning (RL) implementation for enhancing the reasoning capabilities of long CoT models. Building upon the DeepSeek-R1-Distill model series, our RL approach achieves significant performance improvements on various mathematical and coding benchmarks. The Skywork-OR1-32B model outperforms both DeepSeek-R1 and Qwen3-32B on AIME24 and AIME25, while delivering comparable results on LiveCodeBench. Additionally, the Skywork-OR1-7B and Skywork-OR1-Math-7B models demonstrate competitive reasoning performance among similarly sized models.
Our comprehensive ablation studies validate the effectiveness of the core components in our training pipeline, including data mixture and filtration, multi-stage training without advantage masking, high-temperature sampling, exclusion of KL loss, and adaptive entropy control. We conduct extensive investigations into entropy collapse phenomena, identifying key factors that influence entropy dynamics. Our findings show that preventing premature entropy collapse is critical for achieving optimal test performance, offering valuable insights for future research and development. Furthermore, we explore how different training resource allocations affect both training efficiency and final model performance.

\bibliographystyle{plain}
\bibliography{refs}

\end{document}